\documentclass[3p,twocolumn,sort]{elsarticle}

\journal{Knowledge-Based Systems}

\usepackage[switch,columnwise,modulo,mathlines]{lineno} 
\usepackage{comment} 
\usepackage[dvipsnames]{xcolor} 

\usepackage{enumitem}
\usepackage{url}
\usepackage{float}

\usepackage[nolist,nohyperlinks]{acronym}
\acrodef{AA}{Algorithm Adaptation}
\acrodef{ABA}{ADWIN Bagging}
\acrodef{ABO}{ADWIN Boosting}
\acrodef{AMF}{Aggregated Mondrian Forest}
\acrodef{AMR}{Adaptive Model Rules}
\acrodef{ARF}{Adaptive Random Forest}
\acrodef{ADWIN}{Adaptive Windowing}
\acrodef{BA}{Oza Bagging}
\acrodef{BR}{Binary Relevance}
\acrodef{BO}{Oza Boosting}
\acrodef{BOLE}{Boosting Online Learning Ensemble}
\acrodef{CC}{Classifier Chain}
\acrodef{EFDT}{Extremely Fast Decision Tree}
\acrodef{GOCC}{Online Stacked Ensemble of Classifier Chain of HT}
\acrodef{GORT}{Online Stacked Ensemble of iSOUPT}
\acrodef{HAT}{Hoeffding Adaptive Tree}
\acrodef{HT}{Hoeffding Tree}
\acrodef{IDT}{Incremental Decision Tree}
\acrodef{iSOUPT}{Incremental Structured Output Prediction Tree}
\acrodef{kNN}{\textit{k}-Nearest Neighbors}
\acrodef{LP}{Label Powerset}
\acrodef{LR}{Logistic Regression}
\acrodef{MLBELS}{Multi-Label Broad Ensemble Learning System}
\acrodef{MLC}{Multi-Label Classification}
\acrodef{MLHAT}{Multi-Label Hoeffding Adaptive Tree}
\acrodef{MLHT}{Multi-Label Hoeffding Tree}
\acrodef{MLHTPS}{MLHT of Prune Set}
\acrodef{MT}{Mondrian Tree}
\acrodef{NB}{Gaussian Naïve Bayes}
\acrodef{RBF}{Radial Basis Function}
\acrodef{SOM}{Self-Organized Map}
\acrodef{SRP}{Streaming Random Patches}
\acrodef{SGT}{Stochastic Gradient Tree}
\acrodef{TPE}{Tree-structured Parzen Estimator}
\acrodef{PT}{Problem Transformation}

\usepackage{amsmath, bm, siunitx}
\usepackage[longend, ruled, linesnumbered]{algorithm2e}
\SetAlFnt{\scriptsize}
\DontPrintSemicolon
\SetKwProg{Fn}{Function}{:}{end function}
\SetKwFunction{FleafLearning}{leafLearning}

\makeatletter
\SetKwInput{KwSymbols}{Symbols}
\SetFuncSty{myitalic}
\SetCommentSty{myitalic}
\SetKwComment{Comment}{$\rhd~~$}
\makeatother

\usepackage{tabularray}
\UseTblrLibrary{booktabs}
\usepackage{adjustbox}

\usepackage[font=footnotesize,labelfont=footnotesize]{subfig}

\usepackage{amssymb}
\usepackage{pifont}
\DeclareMathOperator{\erf}{erf}

\begin{document}

\begin{frontmatter}

\title{Hoeffding adaptive trees for multi-label classification on data streams}

\author[1]{Aurora~Esteban}
\ead{aestebant@uco.es}
\author[2]{Alberto~Cano}
\ead{acano@vcu.edu}
\author[1]{Amelia~Zafra\corref{mycorrespondingauthor}}
\ead{azafra@uco.es}
\author[1]{Sebastián~Ventura}
\ead{sventura@uco.es}

\cortext[mycorrespondingauthor]{Corresponding author}

 \affiliation[1]{organization={Dept. of Computer Science and Numerical Analysis, Andalusian Research Institute in Data Science and Computational Intelligence (DaSCI), University of Cordoba},
    city={Cordoba},
    postcode={14071}, 
    country={Spain}}
\affiliation[2]{organization={Dept. of Computer Science, Virginia Commonwealth University},
    city={Richmond},
    postcode={23284-3068}, 
    state={VA},
    country={USA}}

\begin{abstract}
Data stream learning is a very relevant paradigm because of the increasing real-world scenarios generating data at high velocities and in unbounded sequences. Stream learning aims at developing models that can process instances as they arrive, so models constantly adapt to new concepts and the temporal evolution in the stream. In multi-label data stream environments where instances have the peculiarity of belonging simultaneously to more than one class, the problem becomes even more complex and poses unique challenges such as different concept drifts impacting different labels at simultaneous or distinct times, higher class imbalance, or new labels emerging in the stream.
This paper proposes a novel approach to multi-label data stream classification called \acf{MLHAT}. \ac{MLHAT} leverages the Hoeffding adaptive tree to address these challenges by considering possible relations and label co-occurrences in the partitioning process of the decision tree, dynamically adapting the learner in each leaf node of the tree, and implementing a concept drift detector that can quickly detect and replace tree branches that are no longer performing well.
The proposed approach is compared with other 18 online multi-label classifiers on 41 datasets. The results, validated with statistical analysis, show that \ac{MLHAT} outperforms other state-of-the-art approaches in 12 well-known multi-label metrics.
\end{abstract}




\end{frontmatter}


\section{Introduction}

Nowadays, many real applications in cyber-physical scenarios generate high volumes of data, arriving in unbounded sequences, at high velocities, and with limited storage restrictions \cite{Akerblom2023, Krawczyk2017}. In this context, data stream mining emerges as an important paradigm in which the learner does not have access to all data at once, must process each instance rapidly, and must learn in an online fashion. 
Thus, online learning aims at developing models capable of constantly expanding and adapting, in contrast to the classic batch learning scenario, where a model is trained once all data are available \cite{Bahri2021}. Since data streams are evolving, the underlying distribution may eventually change, experiencing concept drift, which can impact the decision boundaries and the performance of the classifier. This change can be abrupt, gradual, or incremental, and it also may be recurring when past concepts reappear \cite{Aguiar2024}. Additional challenges include the high dimensionality of the input space \cite{Aguiar2023}, the imbalance ratio in the class labels \cite{Roseberry2021}, or the presence of missing labels in the training data \cite{Bakhshi2024}, which increase the complexity and processing time of the learner and can lead to ignore or forget minority classes.

In this context of complex data arriving rapidly and continuously in a system, that may have limited storage or computation capabilities, decision trees have great potential due to their good balance between accuracy and low complexity. For forecasting on sequential data, for example, there are interesting applications such as the work of Li et al. \cite{Li2023}, which proposes a method based on XGBoost to predict energy consumption.
For classification, \acp{IDT} are very popular and effective algorithms for data stream classification \cite{Bahri2021}. \acp{IDT} are designed to handle such scenarios by incrementally updating the decision model as new data arrives, that is, they learn in a truly online manner, processing instance by instance and being able to produce predictions at any time. This allows the model to adapt to the data-changing distribution and detect changes in the underlying patterns concepts, making them suitable for time-sensitive applications \cite{Akerblom2023, Krawczyk2017}. Additionally, \acp{IDT} are inherently interpretable models, since humans can analyze the path followed in the classification process. This makes them more reliable models in critical applications. Despite these advantages, \acp{IDT} have been applied mainly in the classical scenario of multi-class data stream classification, but not in other more complex scenarios such as multi-label data stream classification.

\ac{MLC} extends the multi-class paradigm by allowing instances to belong to more than one class simultaneously. Traditional \ac{MLC} methods are not coping well with the increasing needs of large and complex structures \cite{Bahri2021}, as they assume a batch learning environment where all data are available in advance. Thus, solving this problem with online learning is a promising approach in these data-intensive environments. However, online \ac{MLC} presents additional unique challenges, like different concept drifts impacting different labels at simultaneous or distinct times \cite{Roseberry2021,Du2020}, or new labels emerging in the stream \cite{Bakhshi2024}. These particularities, along with the own challenges related to both \ac{MLC} and stream mining, make online \ac{MLC} especially difficult.

One of the most relevant models in \acp{IDT} are \acp{HT}, based on the Hoeffding bound \cite{Domingos2000}, which offers mathematical guarantees of convergence with respect to the equivalent batch-learning decision tree. A \ac{HT} builds the tree incrementally and never revisits decisions already made. In contrast, \ac{HAT} \cite{Bifet2009} is a very popular evolution that incorporates concept drift detection in each split as a mechanism to replace parts of the tree if they become obsolete. In the multi-label scenario, there is only one native adaptation of this paradigm exists, \ac{MLHT} \cite{Read2012}, that adapts classic \ac{HT} by incorporating a multi-label classifier in the leaves. Other proposals \cite{Shi2014, Bykakir2018, Liang2022} include this algorithm in different ensemble approaches to increase performance or include concept drift adaptations. However, there is still room for improvement acting directly on the base model, since a single tree versus an ensemble has multiple advantages in terms of computational load, processing speed, or adaptation to concept drift. Our work aims to fill this gap by providing an accurate model for \ac{MLC} in data streams with additional difficulties such as concept drift and label set imbalance evolution, contributing to a significant advance in the multi-output \acp{IDT} research.

This paper presents a novel approach for multi-label data stream classification based on Hoeffding bound named \ac{MLHAT}. \ac{MLHAT} evolves over \ac{HAT} with three main novelties to deal with \ac{MLC} and its complex decision space, label imbalance, and fast response to concept drift. The main contributions of the work can be summarized as:

\begin{itemize}
	\item The presentation of the novel \ac{MLHAT} algorithm, which makes significant advances in multi-label \ac{IDT} in terms of:
	\begin{itemize}[leftmargin=*]
    	\item Considering relations and labels' co-occurrences in the partitioning process of the decision tree. The Bernoulli distribution is used to model the probabilities.
    	\item Dynamically adapting to the high imbalance between labels in the multi-label stream. The number of instances arriving at each leaf node and the impurity among them are monitored to choose between different multi-label classifiers to perform the prediction at leaves.
    	\item Updating the decision model to the change in the data stream distribution or concept drift. A concept drift detector is inserted into each intermediate node of the tree for building a background branch when its performance starts to decrease. This way, at the moment that concept drift is confirmed, the background branch substitutes the main one in the decision tree.
    	\item Increasing the diversity in the learning process and the accuracy of the results. Two well-known methods in stream learning are adapted to the model. On the one hand, applying bootstrapping in instances to learn and, on the other hand, combining the predictions produced by the main tree and the possible background branches if they are significant enough. 
	\end{itemize}
	
    
    \item An extensive experimental study explores the performance of \ac{MLHAT} compared to the state of the art in tree-based methods for online \ac{MLC}. Specifically, 18 classifiers, 41 datasets, and 12 metrics are included in the experimental study.
    
    \item The source code of \ac{MLHAT} and experiments are publicly available\footnote{\url{https://github.com/aestebant/mlhat}} for the sake of reproducible comparisons in future work. In addition, all the datasets used in the experimentation are public: either they belong to well-known \ac{MLC} benchmarks, or they are available in the repository in the case of datasets generated synthetically to include explicit information about concept drift in the multi-label context.
\end{itemize}

The rest of this work is organized as follows. The next section provides a comprehensive review of related works in \ac{MLC}, starting with the fundamentals and going deeper into previous \ac{IDT} proposals. Section \ref{sec:proposal} introduces the \ac{MLHAT} algorithm. Section \ref{sec:exp} presents the experimental study and the results, which compare our approach with state-of-the-art methods on benchmark datasets, in addition to studying its potential in an ensemble architecture. Finally, Section \ref{sec:conclusions} presents the conclusions, summarizes our research findings, and discusses future research directions.
\section{Related work}\label{sec:rwork}

\subsection{Multi-label data stream classification}

The problem of multi-label data stream classification involves predicting multiple labels or categories for incoming data instances in a streaming fashion. Let $S$ represent the data stream as a potential unbounded sequence of instances $(X_1, Y_1)$, $(X_2, Y_2)$, ... $(X_i, Y_i)$, ... where $X_i$ is the feature vector of the $i$-th data instance, and $Y_i$ is a binary vector of labels for the $i$-th data instance with $n$ binary indicators $(y_1, y_2, ... y_n)$ for the presence or absence of each label of the label space $\mathcal{L}$ in the $i$-th instance.
The goal of multi-label data stream classification is to predict the label vectors $Y_i$ for new, unseen instances of $X_i$ in the stream as they arrive, using a predictive model $M$, which can be a classifier or a combination of them that can handle the multi-label nature of the problem:
\begin{equation}
    M(X_i) = Y_i \quad \forall i \in S
\end{equation}

In the online learning scenario, instances arrive at the model one at a time, so it must be able to learn incrementally, updating its knowledge to the last characteristics seen in the stream, unlike in the traditional batch learning scenario, where models have all the data available from the beginning for training and are built statically. In this context, the so-called concept drift arises, which may affect the decision boundaries. In concept drift, two factors must be considered.
On the one hand, when and how the concept drift appears. Concept drift can occur suddenly, incrementally, gradually, or recurrently \cite{Aguiar2024}. In sudden drift, there is an instant change in the data distribution at a particular time, rendering previous models unreliable. Incremental drift is characterized by a steady progression through multiple concepts, with each shift resulting in a new concept closer to the target distribution, and models adapt incrementally to the drift. Gradual drift involves incoming data alternating between two concepts, with a growing bias toward the new distribution over time, and models can adapt gradually. Finally, recurring drift refers to the reappearance of a previously seen concept, and models can be saved and restored when this occurs.
On the other hand, it must be studied if drift is contained within one concept \cite{Alberghini2022}. Thus, the real concept drift refers to a change that makes the previous knowledge about a class' decision boundary invalid, i.e., new knowledge is required to adjust to the shift. In contrast, virtual concept drift is a change that only alters the distribution of data within a known concept, but not the decision boundary. Distinguishing between these two types of drift prevents unnecessary modifications to the classifier.
Although concept drift is a general problem in stream learning, the additional complexity of the multi-label space increases the potential for label imbalance and label distribution changes.

As in the batch learning scenario, there are two general approaches to deal with \ac{MLC} in data streams \cite{Liu2021b}: (i) transforming multi-label data into problems that can be solved using multi-class classifiers, known as \ac{PT}, or (ii) adapting the algorithms from multi-class context to the multi-label paradigm by changing the decision functions, known as \ac{AA}. Attending to \ac{PT}, two main approaches can be followed: either \ac{LP}, which transforms every combination in the label set into a single-class value to convert the multi-label problem into a multi-class problem; or \ac{BR}, that passes from $d$-dimensional label vector $\mathcal{L}$ to $d$ binary classification learners that model each one a label to combine the independent results into a multi-label output. \ac{CC} is a variation of \ac{BR} that compose a chain where the predictions of the previous learners are fed as extra features to the subsequent classifiers.
\ac{PT} approaches allow an easy and straightforward solution for \ac{MLC}, but they also have some known issues \cite{Moyano2018}, including over-training, worsening of class-imbalance problem and worst-case computational complexity in the case of \ac{LP}; loss of label correlation and increase of computational load in the case of \ac{BR}; or sensitivity to the label order and error propagation for \ac{CC}.

\ac{AA} takes the opposite approach with respect to \ac{PT}, focusing on creating methods that natively support the multi-label output space without transforming it. There are some examples of algorithm adaptations based on \ac{kNN} \cite{Zhang2007, Roseberry2021, Roseberry2023}, on rules \cite{Sousa2016, Sousa2018}, or on neural networks trained following a mini-batch approach \cite{Du2020, Bakhshi2024}. In general, models based on \ac{AA} imply a significant reduction of model complexity and computational load, in addition to being better adapted to the multi-label task, with respect to the \ac{PT} paradigm. In any case, models and theoretical results obtained so far in online \ac{MLC} are very limited, and more effort should be put in this direction \cite{Liu2021b}.

A significant challenge in \ac{MLC} is the inherent class imbalance present in many real-world applications \cite{Tarekegn2021a}. This imbalance manifests as a non-uniform distribution of samples and their respective labels over the data space, becoming increasingly complex as the number of labels grows. The challenge is further exacerbated in the streaming context, where the imbalance often occurs simultaneously with concept drift. In this dynamic environment, not only do labels definitions change, but the imbalance ratio itself becomes fluid, with labels roles potentially switching over time \cite{Aguiar2023}. This renders static solutions ineffective, as streams may oscillate between varying degrees of imbalance and periods of balance among labels. Moreover, imbalanced data streams can present additional difficulties such as small sample sizes, borderline and rare instances, class overlapping, and noisy labels. These factors compound the complexity of developing effective classification algorithms for multi-label data streams. In this context, current approaches to handling imbalanced data streams typically fall into two big categories: data-level approaches that resample the dataset to make it balanced, or algorithm-level approaches that design methods to make classifiers robust to skewed distributions \cite{Aguiar2023}. However, focusing on the multi-label data stream field, the research focuses on the second approach. Specifically in this context, ensembles are very popular, with \ac{BR}, previously discussed, as the most straightforward method, although there are other methods such as GOOWE-ML \cite{Bykakir2018} which utilizes spatial modeling to assign optimal weights to a stacked ensemble. In the case of algorithms based on a single model, they need to incorporate specific mechanisms adapted to their nature.

\subsection{Incremental decision trees for multi-label data streams}

\acp{IDT} are a type of decision tree designed to adapt to changes in the data distribution over time. They are based on adapting their structure as new instances arrive, extending the branches, or deleting them if they are no longer accurate. Thus, they differ from classic decision trees such as ID3, C4.5, or CART, in that they do not rebuild the entire tree from scratch to learn from new data.
In multi-class classification based on \acp{IDT}, the state of the art is based on applying the Hoeffding bound \cite{Bahri2021}. The Hoeffding bound offers a mechanism for guaranteeing that the incremental tree building at any time would be equivalent to that built in a batch learning scenario with a confidence level given by the user.
Thus, Domingos and Hulten proposed the \acf{HT}, also known as the Very Fast Decision Tree, \cite{Domingos2000}, which uses the Hoeffding bound to statistically support the decision of the best possible split with the minimum number of instances seen at any moment. Manapragada et al. propose in \cite{Manapragada2018} the \acf{EFDT}, a faster approach in the splitting process that uses the Hoeffding bound to split a node as soon as the split improves the previous node. Another popular approach is \acf{HAT}, presented by Bifet and Gavaldà \cite{Bifet2009}, that evolves from \ac{HT} to incorporate a concept drift detector based on \ac{ADWIN}. This mechanism monitors the performance of every split to build a parallel tree when drift is detected, then it uses the Hoeffding bound to determine whether to replace the main branch with the alternate sub-tree. 

More recently, other \acp{IDT} beyond the Hoeffding approach have arisen. Online \ac{SGT} \cite{Gouk2019} presented by Gouk et al. as an incremental adaptation of the stochastic gradient descent method for building the decision tree. Mourtada et al. presented in \cite{Mourtada2021} an online \ac{MT} that uses the recursive properties of the Mondrian process to split the multidimensional space into regions hierarchically. The online growth of the tree is carried out with an adaptation of the context tree weighting algorithm.

The methods discussed so far have been designed for a multi-class classification scenario with a single label as output.  Although they could be deployed in \ac{MLC} by applying any \ac{PT} technique as \ac{BR}. For \acp{IDT} natively designed for \ac{MLC}, the main proposal for years has been \ac{MLHT}, proposed by Read et al. \cite{Read2012} as an adaptation of \ac{HT} with multi-label classifiers at leaves: the majority labelset in the base case and a tree transformation method in the variation \ac{MLHTPS}. More recently, Osojnik et al. presented \ac{iSOUPT} \cite{Osojnik2017}, a Hoeffding-based decision tree that transforms the \ac{MLC} problem into a multi-target regression one and places an adaptive perceptron in the leaves to perform the predictions.

\begin{table*}[htb]
\centering
	\caption{\acp{IDT} for \ac{MLC} on data streams}\label{tab:rwork}
	\begin{tblr}{
        colspec={X[0.9,l,m] Q[c,m] Q[0.4,c,m] X[0.6,l,m] X[1.3,l,m] X[0.5,c,m] X[0.6,c,m]},
        colsep=3pt,
        rowsep=1pt,
        row{1-Z}={font=\scriptsize},
        row{Z}={font=\bfseries\scriptsize},
    }
		\toprule
		Algorithm & Acronym & Ref & Applicable to multi-label & Split criteria & Adaptable to concept drift & Adaptable to class imbalance \\
		\midrule
		\acl{HT} & \acs{HT} & \cite{Domingos2000} & Transforming the domain & Information gain assuming classes independence & \ding{53} & \ding{53}\\
		\acl{EFDT} & \acs{EFDT} & \cite{Manapragada2018} & Transforming the domain & Information gain assuming classes independence & \ding{53} & \ding{53}\\
		\acl{HAT} & \acs{HAT} & \cite{Bifet2009} & Transforming the domain & Information gain assuming classes independence & \ding{51} & \ding{53}\\
		\acl{SGT} & \acs{SGT} & \cite{Gouk2019} & Transforming the domain & Loss function minimization between target and prediction & \ding{53} & \ding{53}\\
		\acl{MT} & \acs{MT} & \cite{Mourtada2021} & Transforming the domain & Loss function minimization between target and prediction & \ding{53} & \ding{53}\\
		\acl{MLHT} & \acs{MLHT} & \cite{Read2012} & Natively & Information gain considering co-occurrence between labels & \ding{53} & \ding{53}\\
		\acl{iSOUPT} & \acs{iSOUPT} & \cite{Osojnik2017} & Natively & Reduction of intra-cluster variance & \ding{53} & \ding{53}\\
		\acl{MLHAT}  & \acs{MLHAT} & Our proposal & Natively & Multi-label information gain based on multi-variate Bernoulli process & \ding{51} & \ding{51}\\
		\bottomrule
	\end{tblr}
\end{table*}

Table \ref{tab:rwork} summarizes these previous proposals on \ac{IDT}, together with our \ac{MLHAT}, considering their main limitations regarding achieving high performance in evolving and imbalanced data streams in a multi-label environment. Thus, we analyze if they natively support \ac{MLC} or if a problem transformation is needed and if the splitting criteria considers the multi-label specific problems like the co-occurrence between labels. We also analyze if the proposals are adaptable to concept drift, i.e., if the algorithm is able to modify previously built branches. For this characteristic, \ac{MLHAT} stands out as the only proposal that incorporates a concept drift detector adapted to \ac{MLC}. Finally, we study if the proposals are sensitive to the greater class imbalance that exists in multi-label versus the traditional multi-class scenario. This characteristic is quantified by considering whether the models incorporate some mechanism dependent on the cardinality of the received instances. In this case, \ac{MLHAT} is the only proposal adaptable to the imbalance between labels, by monitoring metrics associated with this problem that determine which multi-label classifier to use in each leaf of the tree.

\section{Multi-Label Hoeffding Adaptive Tree}\label{sec:proposal}

This section presents the complete specification of the proposed \acf{MLHAT}, an \ac{IDT} that attempts to overcome the limitations of previous decision tree-based methods for multi-label data streams. Previous works on adapting \acp{HT} to multi-label data streams have three main drawbacks that have caused a lack of popularity, in contrast to the equivalent in traditional data streams. \ac{MLHAT} evolves from the classical \ac{HAT} \cite{Bifet2009} by adding multiple components to address these limitations as follows:

Firstly, previous \acp{IDT} for \ac{MLC} may not consider the relationship between labels when deciding whether to split a leaf node. These \acp{IDT} rely on the entropy function to calculate information gain between the original leaf and potential splits, which assumes that labels are mutually exclusive, as in multi-class scenarios. However, in \ac{MLC} environments, where labels often co-occur, this method creates additional uncertainty and leads to equalized entropies of the original node and its potential splits, impeding tree growth. The proposed \ac{MLHAT} algorithm uses a multivariate Bernoulli process \cite{Dai2013} to calculate the entropy, which considers groups of labels that appear together with higher probability, leading to more accurate approximations of the information gain. This approach also affects the computation of the Hoeffding bound, which determines the significance of the best-split point.

Secondly, previous \acp{IDT} for \ac{MLC} do not account for the imbalance in observed label sets, which may be very severe due to the large number of potential label combinations. This issue affects the final classification at the leaf nodes, as previous models assume either a naive scenario predicting the majority set or a complex multi-label classifier trained on each leaf. The proposed \ac{MLHAT} deals with imbalanced label sets by incorporating two markers at the leaves, multivariate binary entropy and cardinality, and using four prediction scenarios based on their values and set thresholds. Depending on the difficulty of the scenario, the tree leaf will dynamically alternate between a simpler or a more complex classifier.

\begin{figure*}[t]
	\centering
	\includegraphics[width=\linewidth]{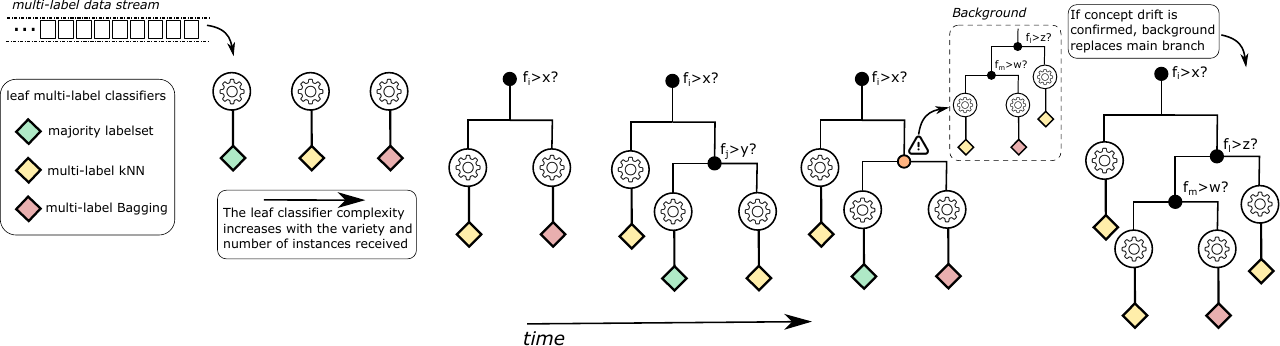}
	\caption{Flowchart of \acl{MLHAT}}\label{fig:flowchart}
\end{figure*}

Finally, existing proposals for concept drift detection in multi-label \ac{IDT}  \cite{Gomes2017, Bifet2007, Liang2022} delegate the mechanism to an ensemble approach, leading to not pruning at branch level, which adds unnecessary complexity to the model. In contrast, \ac{MLHAT} incorporates an \ac{ADWIN} detector at each node, similar to \ac{HAT} \cite{Bifet2009} but with two main differences: it is adapted to monitor multi-label accuracy and it accelerates the reaction to concept drift by triggering an early warning to build a parallel sub-tree in the background of the current node. The background sub-tree replaces the main one if the concept drift is confirmed.

The pseudocode is presented by Algorithm \ref{alg:MLHAT} and Algorithm \ref{alg:leaf}, which defines the function \FleafLearning{} used previously. Figure \ref{fig:flowchart} shows an example of how the \ac{MLHAT} tree evolves over time, starting with a single node that splits as more instances from the stream arrive. The general architecture is detailed in Section \ref{sec:architecture}. The colored squares represent the different levels of the classifier to be used on the leaves, which will depend on the impurity and cardinality of the instances that reach them (see Section \ref{sec:leaves}). When the received instances are sufficient and different enough, the node is split into two sub-branches based on a feature $f_i$ with split point $x$ (see Section \ref{sec:splitting}). It may happen that a split point is no longer valid due to concept drift. In this case, it is replaced by another tree built in parallel and updated to the new data distribution (see Section \ref{sec:drift}). In addition to the learning process described above, \ac{MLHAT} can produce predictions at any time as described in Section \ref{sec:classification}.

\newcommand{\DT}{\textit{MLHAT }}

\begin{algorithm*}[t]
    \caption{\acf{MLHAT}}\label{alg:MLHAT}
    
    \KwIn{\;
        $S \leftarrow \{ (X_0, Y_0), (X_1, Y_1), ..., (X_i, Y_i),... \}$: potentially unbounded multi-label data stream\;
        $\delta_{alt}, \delta_{spl}$: significance levels in the Hoeffding bound for managing alternate trees and splitting nodes respectively\;
        $\kappa_{alt}$: number of instances an alternate tree should see to being considered relevant\;
        $\lambda$: Poisson distribution parameter\;
    }
    \KwSymbols{\;
        $DT(n)$: sub-tree growing from a node $n$\;
        $path$: succession of nodes followed by $X_i$ from an starting point $DT(n_j)$ until a leaf it reached\;
        $\alpha$: multi-label ADWIN concept-drift detector based for each $n$\;
        $W$: instances seen by each $n$, weighted according $Poisson(\lambda)$\;
        \;
    }
    \DT $\leftarrow$ new leaf
    \Comment*[f]{Initialize the tree with a single leaf with empty statistics and fresh classifiers}
    
    \For{$S_i = (X_i, Y_i) \in S$}{
        $w \leftarrow Poisson(\lambda)$ weight for learning\;
        $path \leftarrow$ traverse \DT on $X_i$\;
        $Z_i \leftarrow $ \DT prediction on $X_i$\;
        
        \For{$n \in path$}{
            $\alpha \leftarrow$ ADWIN update based on $Y_i$ and $Z_i$
            \Comment*[f]{Error monitoring and drift update}
            
            \If(\Comment*[f]{Start growing an alternate tree in the node's background}){$\alpha$ detects warning and $\nexists DT'(n)$}{
                $DT'(n) \leftarrow$ new sub-tree in background\;
            }
            \If{$\exists~DT'(n) $ and $W(DT'(n)) > \kappa_{alt}$}{
                $e \leftarrow \alpha$ monitored error in $DT(n)$\;
                $e' \leftarrow \alpha'$ monitored error in $DT'(n)$\;
                $\epsilon_{alt} \leftarrow$ Hoeffding bound associated to $e$ (Eq. \ref{eq:hoeffalt})\;
                \If(\Comment*[f]{Alternate tree is better than main one with confidence $1-\delta_{alt}$}){$(e - e') > \epsilon_{alt}$}{
                    $DT(n) \leftarrow DT'(n)$\;
                    $DT'(n) \leftarrow \varnothing$\; 
                }
                \If(\Comment*[f]{Alternate tree has significantly worsened the main one}){$(e' - e) > \epsilon_{alt}$}{
                    $n \leftarrow$ prune $DT'(n)$\;
                }
            }
            $s \leftarrow$ update statistics of $n$ based on $Y_i$ weighted by $w$\;
            \If(\Comment*[f]{Incremental learning on leaf classifiers and attempt to split in more branches}){$n$ is leaf}{
                \FleafLearning{$n, S_i, w, \delta_{spl}$}\;
            }
            \If(\Comment*[f]{Alternate tree keeps growing in the background}){$\exists~DT'(n)$}{
                $path' \leftarrow $ traverse $DT'(n)$\;
                \For{$n \in path'$}{
                    $\alpha' \leftarrow$ ADWIN update based on $Y_i$ and $Z_i$\;
                    $s \leftarrow$ update statistics of $n$ based on $Y_i$ weighted by $w$\;
                    \If{$n$ is leaf}{
                        \FleafLearning{$n, S_i, w, \delta_{spl}$}\;
                    }
                }
            }
        }
    }
\end{algorithm*}

\begin{algorithm*}[t]
    \caption{Leaf learning in \ac{MLHAT}}\label{alg:leaf}
    \KwIn{\;
        $n$: leaf node belonging to \DT to update\;
        $S_i = (X_i, Y_i)$: multi-label instance\;
        $w$: weight of the instance in learning process\;
        $\delta_{spl}$: significance levels in the Hoeffding bound for splitting nodes\;
        $\kappa_{spl}$: number of instances a leaf should observe between split attempts\;
        $\eta$: number of instances a leaf should see to consider high cardinality\;
    }
    \KwSymbols{\;
        $\gamma_{\downarrow C}$: online classifier for low complexity scenario\;
        $\gamma_{\uparrow C}$: online classifier for high complexity scenario\;
        $\mathcal{L}$: labels' space\;
        $\mathcal{F}$: features' space\;
        $\mathcal{N}_{f}(\mu, \sigma^2)$: Gaussian estimator for conditional probabilities $p(l|x_f)~\forall l \in \mathcal{L}$\;
        \;
    }
    \Fn{\FleafLearning{$n, S_i, w, \delta_{spl}$}}{
        $\mathcal{N}_{fl}(\mu, \sigma^2) \leftarrow $ update node stats on $X_i(f)$, $Y_i$, $w$ $\forall f,l \in \mathcal{F}, \mathcal{L}$\;
        $H_0 \leftarrow$ current entropy in $n$ (Eq. \ref{eq:h})\;
        \If{$H_0 > 0$}{
            \If(\Comment*[f]{Attempt to split}){$\kappa_{spl}$ instances received since last split try}{
                \For{$f \in \mathcal{F}$}{ 
                    $H(f,s) \leftarrow $ entropy at the $s$ that minimizes post-split entropy (Eq. \ref{eq:h1}) among possible splits given by $\mathcal{N}_{f}(\mu, \sigma^2)$
                }
                $\epsilon_{spl} \leftarrow $ Hoeffding bound associated to $L$ (Eq. \ref{eq:hoeffspl})\;
                $G_{f1} \leftarrow$ information gain of the best split candidate $f_i$ at $s_k$ (Eq. \ref{eq:g})\;
                $G_{f2} \leftarrow$ information gain of the second best $f_j$ at $s_l$\;
                \If(\Comment*[f]{There is a split outperforming the rest ones with confidence $1-\delta_{spl}$}){$(G_{f1}-G_{f2}) > \epsilon_{spl}$}{
                    $n \leftarrow$ replace leaf by a split at $(f_1,s_k)$\;
                    $DT(n) \leftarrow$ new branch for $x_f \leq s_k$\;
                    $DT(n) \leftarrow$ new branch for $x_f > s_k$\;
                }
            }
            \eIf(\Comment*[f]{No split, training of leaf classifiers continues}){n is $leaf$}{
                \If(\Comment*[f]{Node cardinality is low}){$W(n) < \eta$}{
                    $\gamma_{\downarrow C} \leftarrow$ incremental learning on $\gamma_{\downarrow C}(S_i, w)$\;
                }
                $\gamma_{\uparrow C} \leftarrow$ incremental learning on $\gamma_{\uparrow C}(S_i, w)$
                \Comment*[f]{$\gamma_{\uparrow C}$ starts learning before $n$ reaches high cardinality}
                
            }(\Comment*[f]{Learning on the new leaves grown from $n$}){
                $leaf \leftarrow $ traverse $DT(n)$\;
                \FleafLearning{$n, S_i, w, \delta_{spl}$}\;
            }
        }
    }
\end{algorithm*}

\subsection{Building the tree}\label{sec:architecture}

Decision tree algorithms work through recursive partitioning of the training set to obtain subsets that are as pure as possible to a given target class, or set of labels in \ac{MLC}. Classical decision trees such as ID3, C4.5 and CART, as well as their multi-label adaptation or transformation approximations \cite{Bogatinovski2022}, assume that all training instances are available at training time, building the tree considering all the characteristics simultaneously. However, \acp{IDT} learn from data sequences, updating the model as the data distribution evolves and more information is available. This paradigm eliminates the need for retraining the whole model when new data arrives, and for keeping all data in memory. It also allows to perform predictions at any moment, even without having received many instances for training. The general workflow of \ac{MLHAT} is similar to previous Hoeffding tree works \cite{Domingos2000, Bifet2009, Read2012}, but with the particularity of being the first proposal, to the best of our knowledge, of incorporating concept drift adaptation in a native multi-label incremental tree.

\ac{MLHAT} is built on two fundamental components in leaves: (i) the Hoeffding bound that determines when a node should split, and (ii) node statistics about the instances that reach it. Initially, \ac{MLHAT} starts with a single node, the root, which will be a leaf node receiving the income instances and updating its statistics about label distributions (details in Section \ref{sec:bernoulli}), as well as multi-label classifiers for imbalanced learning (details in Section \ref{sec:leaves}).
For every $\kappa_{spl}$ instances received, the model tries to split the node using the feature that minimizes the node entropy. For that purpose, a Hoeffding bound $\eta_{spl}$ determines with high probability if the estimated entropy minimization is significant enough given the number of instances observed. How to determine possible node split points and associated Hoeffding bound in the multi-label scenario are described in detail in Section \ref{sec:splitting}.

Once the Hoeffding bound is passed, the root node is divided into two children nodes at the selected split point in the feature space, having now a branch node with two leaf nodes. Each of these children will generate statistics equivalent to those described above from the new instances it receives according to the new partition. Likewise, at any time a node may encounter a feature whose split point exceeds the Hoeffding bound, which will cause a new splitting of that node, causing the tree to grow to a new level of depth. In this way, the tree would expand incrementally as long as the entropy of the leaf nodes can be minimized with new splits.
\ac{MLHAT} implements a mechanism for pruning branches that are no longer needed due to concept drift. Thus, each node incorporates a concept drift detector $\alpha$ that tracks accuracy during training. If the performance starts to decrease at any node $n$, a background sub-tree $DT'(n)$ starts to be built from that node. This tree will grow in parallel to the main one, $DT(n)$. For every $\kappa_{alt}$ instances received, errors of the current and alternate subtrees, $e$ and $e'$, are compared. Again, a Hoeffding bound $\epsilon_{alt}$ is used to determine if the difference in performance is significant enough to make changes in the general structure of \ac{MLHAT}. Section \ref{sec:drift} describes the complete specification of the concept drift adaptation in \ac{MLHAT}.

Finally, it should be noted that the entire training process is conditioned by an online bootstrapping following a Poisson($\lambda$) distribution, in order to apply an extra weight in some instances to perform resampling with replacement from the stream. This approach has been used in multiple previous proposals for data stream classification \cite{Alberghini2022, Bifet2009a, DeBarros2016, Bifet2010, Oza2001}. This extra weight $w$ affects the node statistics, the count of instances seen $W$, as well as the multi-label classifiers used in leaves in general. However, if the classifier in question is based on bagging, it will already have its own modification of the weight of the instances following a Poisson($\lambda$) distribution independent of that of the general \ac{MLHAT} model, since this is the canonical way to simulate bagging in the online paradigm \cite{Oza2001}. In this case, applying both modifiers to the instances used to train the classifier would distort the data too much, so only the Poisson($\lambda$) of the classifier is applied to its learning process.

\subsection{Modeling label co-occurrences with the Multivariate Bernoulli Process}\label{sec:bernoulli}

In several steps of the \ac{MLHAT} building process, as in any decision tree, the entropy of the labels observed at each node plays an important role. To obtain a real measure of entropy in \ac{MLC}, the co-occurrence of labels must be taken into account. In this paper, we use a multivariate Bernoulli distribution \cite{Dai2013} to model this behavior, whose purpose is modeling multiple binary random variables simultaneously. Each binary random variable can take on one of two possible values (0 or 1), which represents the presence or absence of a specific event or condition, just like in the \ac{MLC}. The Bernoulli distribution has interesting properties analogous to the Gaussian distribution that allow us to extend it to high dimensions and construct the so-called multivariate Bernoulli distribution \cite{Dai2013}.

In our setup, each label $L \in \mathcal{L}$ is defined as a Bernoulli random variable with binary outcomes chosen from $l \in \{0,1\}$ and with a probability mass function
\begin{equation}\label{eq:pmf}
    P(L=l) = p^l(1-p)^{1-l}
\end{equation}
Since the possible outcomes are mutually exclusive, $P(L=1) = 1-P(L=0)$, the entropy of $L$ is given by
\begin{equation}
    H(L) = -p \log_2(p) - (1-p) \log_2(1-p)
\end{equation}

Considering all the labels in the problem $(L_1, L_2, ..., L_n)$, we have a $n$-dimensional random vector of possible correlated Bernoulli random variables that can take values $(0, 0, ..., 0)$, $(1, 0, ..., 0)$, ... $(1, 1, ... 1)$. Extending from (\ref{eq:pmf}), the probability mass function in this case is
\begin{equation}
\begin{split}
    &P(L_1=l_1, L_2=l_2, ... L_n=l_n) = p_{0,0,... 0}^{\prod^n_{i=1} (1-l_i)}\\&p_{1,0,... 0}^{l_1\prod^n_{i=2} (1-l_i)} p_{0,1,... 0}^{(1-l_1)l_2\prod^n_{i=3} (1-l_i)} ... p_{1,1,...1}^{\prod^n_{l_i}}
\end{split}
\end{equation}

And the total entropy is defined as the sum of the entropies of all $n$ Bernoulli random variables:
\begin{equation}\label{eq:h}
\begin{split}
    &H(L_1,L_2,...L_n) =\\ &\sum^n_{i=1} -p_{l_i} \log_2(p_{l_i}) - (1-p_{l_i}) \log_2(1-p_{l_i})
\end{split}
\end{equation}

Furthermore, as part of the exponential distribution family \cite{Andersen1970}, the multivariate Bernoulli distribution has other properties applicable to \ac{MLC}, like equivalence of independence and uncorrelatedness, and the fact that both marginal and conditional distributions of a random vector that follows a multivariate Bernoulli distribution are also multivariate Bernoulli. This implies that the conditional probability of any subset of labels $A=(L_i=l_i,.. L_j=l_j)$ given any subset of the rest of them $B=(L_k=l_k,.. L_m=l_m)$ can be computed applying:
\begin{equation}
    P(A|B) = \frac{P(A \cup B)}{P(B)} = \frac{p(l_i, ...l_j, ... l_k,... l_m)}{p(l_k, ... l_m)}
\end{equation}

\subsection{Dynamic multi-label learning at leaves}\label{sec:leaves}

\ac{MLHAT} incorporates multi-label classifiers in the leaves to find the last co-dependencies after the discrimination carried out by the rest of the path in the tree. Due to the complexities of the data flows, and especially in \ac{MLC}, there is a large imbalance between label sets that also affects the number of instances received in each leaf node. Therefore, the learning and prediction strategy should not be uniform for all leaf nodes. There is a previous \ac{MLC} proposal \cite{Law2022} that employs different classifiers depending on the data partition carried out by a decision tree. However, this proposal is designed for a batch learning scenario, so the inference process, the classifiers employed, and the criteria are not applicable to \ac{MLC} in the data streams. \ac{MLHAT} dynamically adapts leaves components as the label distribution of the data stream evolves. For this purpose, each node monitors two markers: (i) the current multi-label entropy at the node $H_0$ as in previously discussed Equation (\ref{eq:h}), and (ii) the cardinality $W$ of the node, which is the number of instances that have reached it at a given time. These markers provide \ac{MLHAT} with four possible learning/prediction scenarios for the multi-label classifiers used in leaves, affecting both the main tree and alternate ones:
\begin{itemize}
    \item $H_0 = 0$: entropy cannot be minimized anymore, i.e., the path to the leaf perfectly separates instances of the same label set. Thus, the model does not need to train additional multi-label classifiers because predicting the majority label set is accurate and computationally efficient.
    
    \item $H_0 > 0~\&~W \leq \eta$: the leaf is going through an intermediate state with entropy starting to increase but cardinality under a given threshold $\eta$. At this point, the leaf incrementally trains a low-cardinality classifier $\gamma_{\downarrow C}$ good at detecting relationships between labels early in data-poor scenarios. In Section \ref{sec:optimization} we study the effect of several multi-label online classifiers to finally select a \ac{LP} transformation of \ac{kNN} because its balance between low complexity and high accuracy with few instances.
    
    \item $\Delta G(H_0) \leq \epsilon_{spl}~\&~W > \eta$: a high number of instances but a low information gain $\Delta G$ between the entropies currently and after the eventual split, imply that it is difficult to separate the label-sets given their feature space. The information gain computation is discussed in Section \ref{sec:splitting}. In this scenario, it is necessary to employ a more data demanding and computationally expensive multi-label classifier, but capable of finding deeper relationships, while keeping computational complexity under control so that \ac{MLHAT} remains competitive in a stream data scenario. Section \ref{sec:optimization} discusses the effect of various multi-label online classifiers based on ensembles and determines \ac{BR} transformation of the ensemble \ac{ABA}+\ac{LR} as the most suitable option due to its superiority in learning from larger data streams.
    
    \item $\Delta G(H_0) > \epsilon_{spl}~\&~W > \eta$: if after passing the high cardinality threshold it is also observed that entropy increases and there are significant differences between possible splits, the node has seen enough and sufficiently diverse instances, so it is ready to attempt to split following the procedure described in Section \ref{sec:splitting}.
\end{itemize}

\subsection{Multi-label splitting into new branches}\label{sec:splitting}

The Hoeffding bound mathematically supports the split decisions in \ac{MLHAT} leaves determining the minimum number of instances needed to decide if the split candidate would be equivalent to the one selected in a batch learning scenario where all instances are available. We define the Hoeffding bound for \ac{MLC} based on \cite{Domingos2000} but considering the number of known label sets $|L|$ instead of the single labels:
\begin{equation}
    \epsilon_{spl} = \sqrt{\frac{\log_2(|L|)^2 \ln(1/\delta_{spl})}{2W}}
    \label{eq:hoeffspl}
\end{equation}
where $W$ is the number of instances received and $\delta_{spl}$ is an error parameter for measuring the confidence in the decision.

The Hoeffding bound is used to determine if the first best-split candidate is significantly better than the second one. To find these two best candidates, each possible candidate in the features space is evaluated using the information gain $G(f, s)$ between the current entropy $H_0$ at the node that may be partitioned, and the estimated posterior entropy $H(f, s)$ if the data received in that node so far were partitioned at feature $f$ taking value $s$:
\begin{equation}
    G(f, s) = H_0 - H(f, s)
    \label{eq:g}
\end{equation}
The information gain difference $\Delta G$ between the attribute $f_i$ with the best split at the value $s_k$, and the second best split given at the attribute $f_j$ with value $s_l$, with $i \neq j$, is computed to pass the Hoeffding test:
\begin{equation}
    \Delta G = (G(f_i, s_k) - G(f_j, s_l)) > \epsilon_{spl}
    \label{eq:deltag}
\end{equation}

If this test is satisfied, the best split point $f_i = s_j$, found after observing $W$ instances in the leaf, would be the same as the one that would be selected in a batch learning scenario, with a confidence of $1-\delta_{spl}$.

To calculate the different entropies implied in the process considering label co-occurrences, we use the multivariate Bernoulli distribution as discussed in Section \ref{sec:bernoulli}. Thus, $H_0$ is directly obtained from Equation (\ref{eq:h}), where label priors $p(l)$ are obtained from the counters maintained by the node:
\begin{equation}\label{eq:pl}
    p(l) = \frac{W_l}{W} ~\forall ~l \in \mathcal{L}
\end{equation}

The entropy after the candidate binary split $H(f,s)$ is also estimated as a succession of binary entropies, aggregating the entropies from the generated branches given by whether the splitting criterion is met, $(x_f \in s)$ or not $(x_f \notin s)$:
\begin{equation}
    H(f,s) = p(s)H(x_f \in s) + p(\neg s)H(x_f \notin s) \label{eq:h1}
\end{equation}
where $p(s)$ and $p(\neg s)$ are the probabilities of occurrence of the splitting criterion, to balance the importance of each partition, and $H(f \in s)$ and $H(f \notin s)$ measure the entropies of each branch by conditioning the labels probabilities to the splitting criterion:
\begin{equation}
    \begin{split}
    H(x_f \in s) &= \sum_{l \in \mathcal{L}} - p(l|s) \log_2 (p(l|s))-\\&- p(\neg l|s) \log_2 (p(\neg l|s))\\
    H(x_f \notin s) &= \sum_{l \in \mathcal{L}} - p(l|\neg s) \log_2 (p(l|\neg s))-\\&- (p(\neg l|\neg s) \log_2 (p(\neg l|\neg s)
    \end{split}
\end{equation}

The computation of the conditional probabilities of equation depends on the nature of the feature $f$. Our \ac{MLHAT} natively handles categorical and numerical features. In categorical cases, the given feature has already well-defined partitions $f \in \{ A, B, ... X \}$. Since both the feature and the label take discrete and mutually exclusive values, conditional probabilities are obtained by counting the occurrences of the given label $W_{l,s}$ among all the observed instances in the node that meet the criterion $W_s$:
\begin{equation}
    \begin{split}
        &p(l|x_f = s) = \frac{W_{l,s}}{W_s}, p(\neg l|x_f = s) = 1 - p(l|x_f = s)\\
        &p(l|x_f \neq s) = \frac{W_{l,\neg s}}{W_{\neg s}}, p(\neg l|x_f \neq s) = 1 - p(l|x_f \neq s)
    \end{split}
\end{equation}

On the other hand, for numerical features that do not have discrete partition points but move in a certain range, $f \in [a,b], ~ a,b \in \mathbb{R}$, we use a common approach in decision trees \cite{Korycki2021} consisting of modeling the feature space with Gaussian estimators $\mathcal{N}(\mu, \sigma^2)$. This provides an efficient way to calculate conditional probabilities given a splitting point $s \in [a, b]$ to create the branches such that:
\begin{equation}
    \begin{split}
        p(l|x_f \leq s)&= \frac{p(x_f \leq s|l)p(l)}{p(x_f \leq s)}, \\
        p(\neg l|x_f \leq s)&= \frac{p(x_f \leq s|\neg l)p(\neg l)}{p(x_f \leq s)}\\
        p(l|x_f > s)&= \frac{(1-p(x_f \leq s|l))p(l)}{p(x_f > s)}, \\
        p(\neg l|x_f > s)&= \frac{(1-p(x_f \leq s|\neg l))p(\neg l)}{p(x_f > s)} 
    \end{split}
\end{equation}
where $p(x_f \leq s)$ and $p(x_f > s)$ are normalizing constants for all labels, and $p(l)$ and $p(\neg l)$ are defined in (\ref{eq:pl}). Finally, $p(x_f \leq s |l)$ and $p(x_f \leq s |\neg l)$ are obtained from the cumulative density functions, $\Phi_{f,l} (f \leq s)$ and $\Phi_{f, \neg l} (f \leq s)$, associated with the respective Gaussian estimators maintained in the node for each pair of feature $f \in \mathcal{F}$ and target in $l \in \mathcal{L}$. Thus, the inverse conditional probability in each case is obtained as:
\begin{equation}
    p(x_f \leq s | l) = \Phi_{f,l} (f \leq s) = 0.5\frac{1+\erf(s-\mu)}{\sigma\sqrt{2}}
\end{equation}

\begin{figure*}[t]
    \centering
    \subfloat[Evaluating a categorical feature]{\includegraphics[width=.9\linewidth]{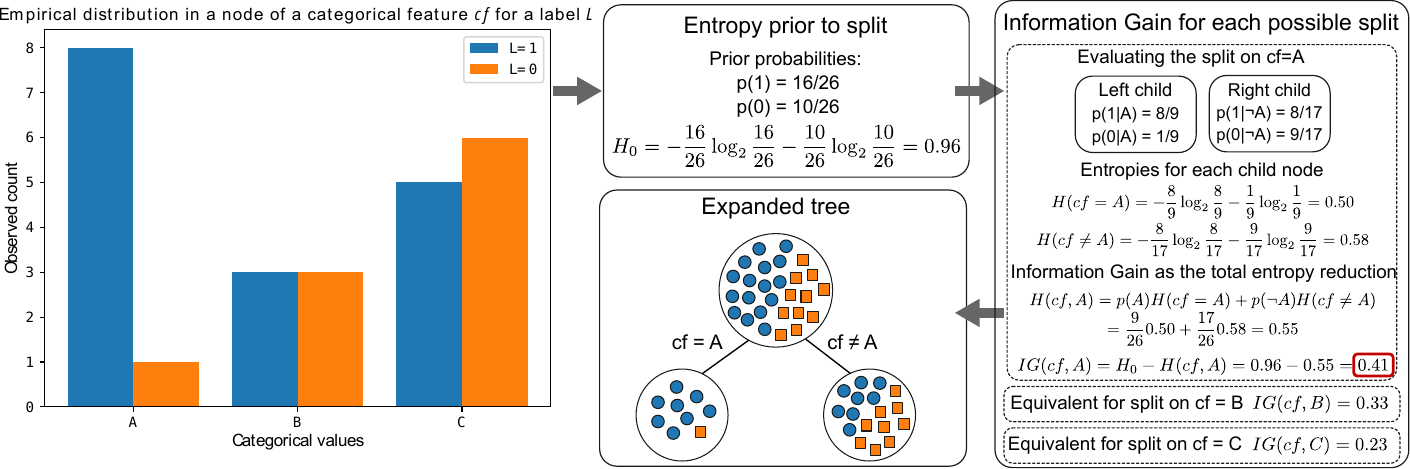}}\\
    \subfloat[Evaluating a numerical feature]{\includegraphics[width=.9\linewidth]{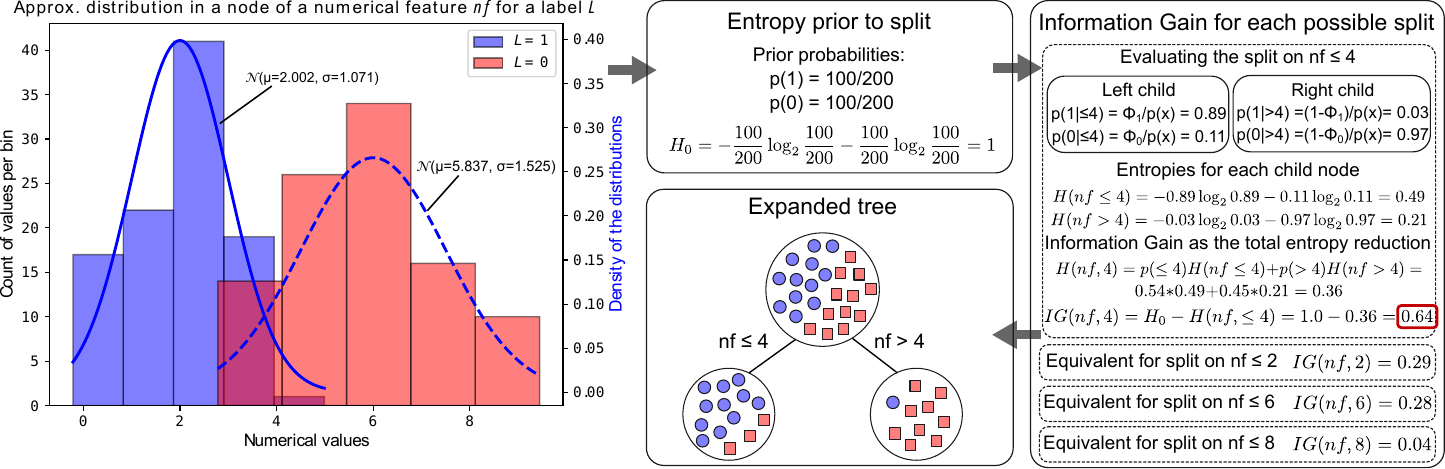}}
    \caption{Example of information gain computation in MLHAT in the categorical and the numerical cases}\label{fig:ig}
\end{figure*}

For the categorical features, the splitting values to evaluate are given by the support of the given feature, while for the numerical features, a binning process from the observed range of the feature is applied to obtain possible splitting points $s$. Figure \ref{fig:ig} shows a very simple example for each case, obtaining the information gain of the possible splits given a problem with only one feature and one label. In the first case, the node has so far received 26 instances, which are distributed in 16 positive and 10 negative for the only label considered, which gives it an initial entropy of $H_0=0.96$. For these instances, the information gain of a hypothetical categorical variable that takes three possible values is evaluated. After applying the process described above, it is obtained that splitting on $s=A$ provides the highest discriminant power so that split value would be the candidate to apply the Hoeffding bound and, if it passes it, expand the node as shown in the figure to create two new branches.
In the second case of Figure \ref{fig:ig}, a node has received 200 instances equally distributed between positive and negative for the considered label, so the initial entropy is maximum. These instances have a numerical variable that has been observed to move between in the range $[0, 9]$ approximately. The figure shows the observed distribution of the variable for each target and the associated Gaussian estimate. In the example, four possible split values are considered, obtained from applying the binning process in the observed range. It is determined that $s=4$ is the one that maximizes the information gain. If this candidate exceeded the Hoeffding limit, it would be used to expand the node into two branches, as shown in the example.

\subsection{Concept drift adaptation}\label{sec:drift}

\ac{ADWIN} \cite{Bifet2007} is an adaptive algorithm that detects changes in data distribution on a set number of instances. It works by comparing the statistical properties of two sub-portions of a window and determining if there's a significant difference in the mean values.  Many methods in the field of stream learning implement \ac{ADWIN} \cite{Bifet2009, Alberghini2022, Gomes2017}. \ac{MLHAT} utilizes \ac{ADWIN} by incorporating a detector $\alpha$ at each node of the decision tree and potential background sub-trees. This allows \ac{ADWIN} to detect changes in the data distribution that could only impact specific features determined by different paths in the tree. Specifically, $\alpha$ monitors the error $e$ in the Hamming loss between the baseline $Y_i$ and the prediction $Z_i$ of the training instances. Upon detection of a warning drift, a background sub-tree starts growing from the node affected, with the same components and expanding upon the new data distribution. More discussion of the effect of the monitored metric on the final performance is presented in Section \ref{sec:optimization}.

The Hoeffding bound is used to determine the moment when the difference between errors in main $e$ and alternate $e'$ sub-trees are significant enough, with a confidence of $1-\delta_{alt}$, to perform structural changes. The Hoeffding bound $\epsilon_{alt}$ for alternating trees is based on \cite{Bifet2009} and defined from monitored errors and instances seen by both main $W$ and alternate $W'$ sub-trees as:
\begin{equation}
    \epsilon_{alt} = \sqrt{\frac{2e (1-e') (W+W') \ln(2/\delta_{alt})}{W \cdot W'}}
    \label{eq:hoeffalt}
\end{equation}
Given this threshold, we have three possible scenarios:

\begin{itemize}
    \item $(e - e') > \delta_{alt}$: the alternate tree is better than the main one so it will be replaced, pruning the previous structure at the node affected by the concept drift and resetting the drift detector in the alternate tree for future concept drifts.

    \item $(e' - e) > \delta_{alt}$: the alternate tree is not improving the main one because the concept drift was reversed. In this case, the main tree remains without changes and the alternate tree is pruned.

    \item $\delta_{alt} \geq |e - e'|$: if no significant differences are found, the two sub-trees continue to receive instances and expand to eventually differentiate their performance.
\end{itemize}

\subsection{Classification at leaves}\label{sec:classification}

In the stream learning paradigm, the instances to be classified can arrive at any time, without there being a differentiated phase between building the model and generating the predictions. In \ac{MLHAT}, the classification process is detailed in Algorithm \ref{alg:classification} and consists of two phases: (i) traversing the tree with the incoming instance using the tree branches that partition the feature space and (ii) assigning a label set to the instance depending on the leaf state reached. Three scenarios are considered in the leaf node, related to the entropy and cardinality thresholds used in learning (see Section \ref{sec:leaves}):
\begin{itemize}
    \item $H_0 = 0$: minimal entropy implies that all the instances seen until the moment by the reached node belong to the same label set, so the target instance is also assigned to that label set. 

    \item $H_0 > 0~\&~W \leq \eta$: the cardinality of the node is still low, so the least data demanding classifier $\gamma_{\downarrow C}$ is used to assign a label set to the target instance.

    \item $H_0 > 0~\&~W > \eta$: the cardinality of the node is high and, because it is still not partitioned, the decision boundaries in the feature space should be difficult to determine. Therefore, the classification of the target instance is performed with the most computationally complex classifier $\gamma_{\uparrow C}$.
\end{itemize}

\begin{algorithm*}[t]
    \caption{Classification in MLHAT}\label{alg:classification}

    \KwIn{\;
        $X_i$: an unknown multi-label instance in the $S$ domain\;
        $\eta$: number of instances a leaf should see to consider high cardinality\;
        $\kappa_{alt}$: number of instances an alternate tree should see to being considered relevant\;
    }
    \KwSymbols{\;
        \DT: Initialized model trained with at least one instance\;
        $DT(n)$: subtree growing from a node $n$\;
        $e$: monitored error by $\alpha$ at a node\;
        $W$: instances seen by each $n$, weighted according $Poisson(\lambda)$\;
        $\gamma_{\downarrow C}$: online classifier for low complexity scenario\;
        $\gamma_{\uparrow C}$: online classifier for high complexity scenario\;
        \;
    }
    $path \leftarrow$ traverse \DT on $X_i$\;
    $leaf \leftarrow$ leaf node reached by $X_i$ through $path$\;
    \eIf(\Comment*[f]{All the instances discriminated by $path$ belong to same labelset}){node entropy is 0}{
        $Z_i \leftarrow $ final prediction as the majority labelset\;
    }{
    \eIf{$W(leaf) < \eta$}{
        $p(Z_i) \leftarrow \gamma_{\downarrow C}$ probabilistic multi-label prediction on $X_i$ pondered by $e$\;
    }{
        $p(Z_i) \leftarrow \gamma_{\uparrow C}$ probabilistic multi-label prediction on $X_i$ pondered by $e$ \;
    }
    \For(\Comment*[f]{$Z_i$ is complemented with the ones provided by significant alternate trees}){$n \in path$}{
        \If{$\exists ~ DT'(n)$}{
            $leaf' \leftarrow $ leaf node reached traversing $DT'(n)$ on $X_i$\;
            \If{$W(path) > \kappa_{alt}$}{
                \eIf{$W(leaf') < \eta$}{
                    $p(Z_i) +\leftarrow \gamma_{\downarrow C}$ probabilistic multi-label prediction on $X_i$ pondered by $e$\;
                }{
                    $p(Z_i) +\leftarrow \gamma_{\uparrow C}$ probabilistic multi-label prediction on $X_i$ pondered by $e$ \;
                }
            }
        }
    }
    $Z_i \leftarrow $ final prediction after accumulated probabilities normalization\;
    }
\end{algorithm*}

As mentioned, if \ac{MLHAT} is in the midst of a concept change, it will have a main tree and one or more alternate sub-trees that are candidates to replace the branches whose performance is declining. In these cases, the instance to be classified could simultaneously reach a leaf node of the main tree and one or more leaves of the alternate sub-trees. In the context of additional complexity of multi-label stream classification, one of the main challenges is to accelerate the speed of reaction to concept drift. Along these lines, in Section \ref{sec:drift} we have proposed the alarm-based drift early adaptation mechanism. And following this line, in this section we propose a system for combining predictions between the main tree and the possible alternate sub-trees that an instance to be predicted can reach. The idea is to maintain a balance between the main structure that has the statistical guarantees of the Hoeffding bound, and to provide a fast response to possible concept drifts. Thus, this weighting will be done under two conditions: (i) only leaves from alternate trees that have received at least $\kappa_{alt}$ instances (the same threshold as for considering the replacement of an alternate tree) will be taken into account, and (ii) the prediction will be weighted based on the error monitored in each of the leaves involved. The section \ref{sec:optimization} discusses how this measure affects model performance in a practical way.

\section{Experimental study}\label{sec:exp}

This section presents the experimental study and comparison with the state-of-the-art works. The experiments are designed to answer the following research questions:
\begin{itemize}
    \item \textbf{RQ1:} Can \ac{MLHAT} demonstrate competitive performance compared to the state of the art incremental decision tree methods for multi-label data streams?
    \item \textbf{RQ2:} Is it more beneficial to use \ac{MLHAT}'s native multi-label split criterion rather than applying \acf{PT} to use a non-adapted split criterion?
    \item \textbf{RQ3:} Is \ac{MLHAT} robust in a multi-label data stream scenario where concept drift may appear?
\end{itemize}

\subsection{Experimental setup}

\subsubsection{Algorithms}

Table \ref{tab:algorithms} presents a taxonomy of the 16 multi-label incremental algorithms used in this experimental study, including \acf{BR} to transform the problem and \acf{AA} methods, as well as whether they are adaptable to concept drift and adaptable to class-imbalance (yes (\ding{51}), no (\ding{53})). As discussed in Section \ref{sec:rwork}, we have considered whether algorithms are adaptive to class imbalance if they include explicit elements such as an ensemble-based construction or our proposal with dynamic leaf classifiers. The same for concept drift, they are considered adaptive if they incorporate explicit elements for this purpose.
All algorithms are implemented in Python and are publicly available in the River library \cite{Montiel2021}. The source code of \acs{MLHAT} is also publicly available in the repository associated with this work\footnote{\url{https://github.com/aestebant/mlhat}} for the sake of reproducibility. In model selection, approaches based on \ac{IDT} have special prevalence. Thus, we consider 5 multi-class \acp{IDT} transformed to \ac{MLC} using \ac{BR}; 3 \acp{IDT} for \ac{MLC}, and 2 forest-based methods, i.e., ensembles designed for specific trees as base models. Moreover, the experimentation includes other algorithms from well-known paradigms like neighbors, rules, Bayes, and bagging and boosting, as well as the recent \acs{MLBELS}, based on neural network and trained in mini-batches \cite{Bakhshi2024}. In this last case, in order to make the experimentation conditions as similar as possible to the other proposals, which work purely online, the smallest batch size studied by the authors, 50 instances, is used.
In general, the hyperparameter setting has followed three principles: (i) not to make individualized adjustments per dataset, (ii) to start from the configurations suggested by the authors or mostly used in the literature, and (iii) to keep equivalent configurations in models of the same family. Thus, all Hoeffding trees, including our \acs{MLHAT}, have been configured with Hoeffding significance $\delta_{spl}=1\mathrm{e}{-5}$ and a grace period of $\kappa_{spl}=200$, and all ensembles have been configured with 10 base models. The complete parameter specification per algorithm can be found in the associated repository.

\begin{table*}[ht]
    \centering
    \caption{Algorithms included in the comparative studies}\label{tab:algorithms}
    \begin{tblr}{
        width=.95\linewidth,
        colspec={Q[l,m] Q[c,m] Q[c,m] Q[l,m] X[l,m] Q[c,m] Q[c,m]},
        colsep=3pt,
        rowsep=0.2pt,
        row{1-Z}={font=\scriptsize}
    }
    \toprule
    Family & Ref & Year & Acr & Algorithm & {Adaptable to\\concept-drift} & {Adaptable to\\class-imbalance}\\
    \midrule
    BR+IDT & \cite{Domingos2000} & 2000 & \acs{HT} & \acl{HT} & \ding{53} & \ding{51}\\
    BR+IDT & \cite{Bifet2009} & 2009 &\acs{HAT} & \acl{HAT} & \ding{51} & \ding{51}\\
    BR+IDT & \cite{Manapragada2018} & 2018 & \acs{EFDT} & \acl{EFDT} & \ding{53} & \ding{51}\\
    BR+IDT & \cite{Gouk2019} & 2019 & \acs{SGT} & \acl{SGT} & \ding{53} & \ding{51}\\
    BR+IDT & \cite{Mourtada2021} & 2021 & \acs{MT} & \acl{MT} & \ding{53} & \ding{51}\\
    
    BR+Bayesian & \cite{Bifet2009a} & 2009 & \acs{NB} & \acl{NB} & \ding{53} & \ding{51}\\
    BR+Distance & \cite{Montiel2021} & 2021 & \acs{kNN} & \acl{kNN} & \ding{53} & \ding{51}\\
    BR+Rules & \cite{Duarte2016} & 2016 & \acs{AMR} & \acl{AMR} & \ding{51} & \ding{51}\\
    
    BR+Ensb & \cite{Bifet2009a} & 2009 & \acs{ABA} & \acl{ABA} + \acl{LR} & \ding{51} & \ding{51}\\
    BR+Ensb & \cite{Bifet2009a} & 2009 & \acs{ABO} & \acl{ABO} + \acl{LR} & \ding{51} & \ding{51}\\
    
    
    BR+Forest & \cite{Gomes2017} & 2017 & \acs{ARF} & \acl{ARF} & \ding{51} & \ding{51}\\
    BR+Forest & \cite{Mourtada2021} & 2021 & \acs{AMF} & \acl{AMF} & \ding{53} & \ding{51}\\
    
    AA+IDT & \cite{Read2012} & 2012 & \acs{MLHT} & \acl{MLHT} & \ding{53} & \ding{53}\\
    AA+IDT & \cite{Read2012} & 2012 & \acs{MLHTPS} & \acl{MLHTPS} & \ding{53} & \ding{53}\\
    AA+IDT & \cite{Osojnik2017} & 2017 & \acs{iSOUPT} & \acl{iSOUPT} & \ding{53} & \ding{53}\\
    AA+NN & \cite{Bakhshi2024} & 2024 & \acs{MLBELS} & \acl{MLBELS} & \ding{53} & \ding{53}\\
    
    AA+Ensb+DT & \cite{Bykakir2018} & 2018 & \acs{GOCC} & \acl{GOCC} & \ding{53} & \ding{51}\\
    AA+Ensb+DT & \cite{Bykakir2018} & 2018 & \acs{GORT} & \acl{GORT} & \ding{53} & \ding{53}\\
    
    AA+IDT & This & 2024 & \acs{MLHAT} & \acl{MLHAT} & \ding{51} & \ding{51}\\
    \bottomrule
    \end{tblr}
\end{table*}

\subsubsection{Datasets}
\label{sec:dataset}

Multi-label datasets used in the experimental study cover a wide range of properties. On the one hand, we evaluate the performance of the algorithms included in the study on 29 real datasets publicly available\footnote{\url{https://www.uco.es/kdis/mllresources/}} of up to 269,648 instances, 31,802 features, and 374 labels. These datasets are used to address RQ1 and RQ2. The complete specification of each dataset is shown in Table \ref{tab:datasets} attending to the number of instances, features, and labels, as well as other statistics like cardinality (average amount of labels per instance), density (cardinality divided by the number of labels), and the mean imbalance ratio per label (average degree of imbalance between labels). Moreover, the last column indicates whether the instances in the dataset are presented in a temporal order: yes (\ding{51}), no (\ding{53}), or this information can not be known from the provided description (-). Since in these datasets there is no explicit information about concept drift, this factor can give an idea about the chances of finding concept drift in them.

\begin{table*}[t]
    \centering
    \caption{Real-world datasets used in the experimental study}\label{tab:datasets}
    \begin{tblr}{
        width=.8\linewidth,
        colspec={X[l,m] X[0.5,l,m] *{6}{Q[r,m]} Q[c,m]},
        colsep=3pt,
        rowsep=0.1pt,
        row{1-Z}={font=\scriptsize}
    }
    \toprule
    Dataset & Abbreviation & Instan. & Features & Labels & Card. & Dens. & MeanIR & {Temporal\\order}\\
    \midrule
    Flags & Flags & 194 & 19 & 7 & 3.39 & 0.48 & 2.255 & \ding{53} \\
    WaterQuality & WQ & 1,060 & 16 & 14 & 5.07 & 0.36 & 1.767 & - \\
    Emotions & Emo & 593 & 72 & 6 & 1.87 & 0.31 & 1.478 & \ding{53} \\
    VirusGO & Virus & 207 & 749 & 6 & 1.22 & 0.20 & 4.041 & \ding{53} \\
    Birds & Birds & 645 & 260 & 19 & 1.01 & 0.05 & 5.407 & - \\
    Yeast & Yeast & 2,417 & 103 & 14 & 4.24 & 0.30 & 7.197 & \ding{53} \\
    Scene & Scene & 2,407 & 294 & 6 & 1.07 & 0.18 & 1.254 & \ding{53} \\
    GnegativePseAAC & Gneg & 1,392 & 440 & 8 & 1.05 & 0.13 & 18.448 & - \\
    CAL500 & CAL500 & 502 & 68 & 174 & 26.04 & 0.15 & 20.578 & \ding{53}  \\
    HumanPseAAC & Human & 3,106 & 440 & 14 & 1.19 & 0.08 & 15.289 & - \\
    Yelp & Yelp & 10,806 & 671 & 5 & 1.64 & 0.33 & 2.876 & \ding{51} \\
    Medical & Med & 978 & 1,449 & 45 & 1.25 & 0.03 & 89.501 & \ding{51} \\
    EukaryotePseAAC & Eukar & 7,766 & 440 & 22 & 1.15 & 0.05 & 45.012 & - \\
    Slashdot & Slashdot & 3,782 & 1,079 & 22 & 1.18 & 0.05 & 19.462 & \ding{51} \\
    Hypercube & HC & 100,000 & 100 & 10 & 1.00 & 0.10 & - & - \\
    Hypersphere & HS & 100,000 & 100 & 10 & 2.31 & 0.23 & - & - \\
    Langlog & Langlog & 1,460 & 1,004 & 75 & 15.94 & 0.21 & 39.267 & \ding{51} \\
    StackexChess & Stackex & 1,675 & 585 & 227 & 2.41 & 0.01 & 85.790 & \ding{51} \\
    ReutersK500 & Reuters & 6,000 & 500 & 103 & 1.462 & 0.01 & 54.081 & - \\
    Tmc2007500 & Tmc & 28,596 & 500 & 22 & 2.22 & 0.10 & 17.134 & \ding{51} \\
    Ohsumed & Ohsum & 13,929 & 1,002 & 23 & 0.81 & 0.04 & 7.869 & \ding{51} \\
    D20ng & D20ng & 19,300 & 1,006 & 20 & 1.42 & 0.07 & 1.007 & \ding{51} \\
    Mediamill & Media & 43,907 & 120 & 101 & 4.38 & 0.04 & 256.405 & - \\
    Corel5k & Corel5k & 5,000 & 499 & 374 & 3.52 & 0.01 & 189.568 & \ding{53} \\
    Corel16k001 & Corel16k & 13,766 & 500 & 153 & 2.86 & 0.02 & 34.155 & \ding{53} \\
    Bibtex & Bibtex & 7,395 & 1,836 & 159 & 2.40 & 0.02 & 12.498 & \ding{53} \\
    NusWideCVLADplus & NWC & 269,648 & 129 & 81 & 1.87 & 0.02 & 95.119 & - \\
    NusWideBoW & NWB & 269,648 & 501 & 81 & 1.87 & 0.02 & 95.119 & - \\
    Imdb & Imdb & 120,919 & 1,001 & 28 & 1.00 & 0.04 & 25.124 & \ding{53}  \\
    YahooSociety & YahooS & 14,512 & 31,802 & 27 & 1.67 & 0.06 & 302.068 & - \\
    EurlexSM & Eurlex & 19,348 & 5,000 & 201 & 2.21 & 0.01 & 536.976 & \ding{51} \\
    \bottomrule
    \end{tblr}
\end{table*}

To the authors' knowledge, there are no public multi-label datasets that expressly contain information about concept-drift \cite{Read2012, Shi2014,Cerri2021}. Therefore, to address RQ3, 12 additional datasets are synthetically generated to study the impact of various types of concept drift. These datasets have been generated with the MOA framework \cite{Bifet2010a} and are public in the repository associated to this paper. We employ three multi-label generators based on Random Tree, \ac{RBF} and Hyper-plane, under two configurations widely used \cite{Read2012}: 30 attributes and 8 labels, and 80 attributes and 25 labels. The drifts are controlled by two parameters: the label cardinality $Z$ and the label dependency $u$, that are altered in different ways to generate the four main types of concept drift in \ac{MLC}: sudden, gradual, incremental and recurrent. In all cases, the drifts take place at 3 times depending on the size $N$ of the stream: at times $N/4$, $2N/4$ and $3N/4$, and each drift changes the underlying concept of the stream. Thus, the controlled parameters in the first stretch of the stream are $Z=1.5$, $u=0.25$; in the second only the label dependency changes, so we have $Z=1.5$, $u=0.15$; in the third stretch only the cardinality changes to have $Z=3.0$, $u=0.15$; and finally, in the fourth, they both change to end up with $Z=1.5$, $u=0.25$ again. In the recurrent case, only the configurations of the 1st and 3rd stretches are alternated twice. Table \ref{tab:datasets-synth} shows the main characteristics of these synthetic datasets regarding features, labels, type of concept drift and its width, i.e., for how many instances the drift change extends. For more details, refer to the paper repository.

\begin{table*}[t]
    \centering
    \caption{Synthetic datasets used in the experimental study}\label{tab:datasets-synth}
    \begin{tblr}{
        width=0.7\linewidth,
        colspec={X[l,m] *{6}{Q[c,m]}},
        colsep=2pt,
        rowsep=0.1pt,
        row{1-Z}={font=\scriptsize}
    }
    \toprule
    Dataset & Instances & Features & Labels & Generator & Drift type & Drift width\\
    \midrule
    SynTreeSud & 50,000 & 20 num.+10 cat. & 8 & Random Tree & sudden & 1\\
    SynRBFSud & 50,000 & 80 numeric & 25 & Random RBF & sudden & 1\\
    SynHPSud & 50,000 & 30 numeric & 8 & Hyper Plane & sudden & 1\\
    SynTreeGrad & 50,000 & 20 num.+10 cat. & 8 & Random Tree & gradual & 500\\
    SynRBFGrad & 50,000 & 80 numeric & 25 & Random RBF & gradual & 500\\
    SynHPGrad & 50,000 & 30 numeric & 8 & Hyper Plane & gradual & 500\\
    SynTreeInc & 50,000 & 20 num.+10 cat. & 8 & Random Tree & incremental & 275\\
    SynRBFInc & 50,000 & 80 numeric & 25 & Random RBF & incremental & 275\\
    SynHPInc & 50,000 & 30 numeric & 8 & Hyper Plane & incremental & 275\\
    SynTreeRec & 50,000 & 20 num.+10 cat. & 8 & Random Tree & recurrent & 1\\
    SynRBFRec & 50,000 & 80 numeric & 25 & Random RBF & recurrent & 1\\
    SynHPRec & 50,000 & 30 numeric & 8 & Hyper Plane & recurrent & 1\\
    \bottomrule
    \end{tblr}
\end{table*}

\subsubsection{Evaluation metrics}\label{sec:evaluation}

Due to the incremental nature of the data streams, the canonical way to evaluate performance working with them follows a test-then-train scheme known as prequential evaluation \cite{Gama2009}. In this work, we apply this methodology to all experiments setting a forgetting factor of $\alpha=0.995$ and 50 steps between evaluations.

For comparing the performance between algorithms, we employ 11 different metrics that evaluate the total and partial correctness of multi-label prediction \cite{Bogatinovski2022}, as well as the total computing time in seconds. The metrics are defined as follows on the labels $L$ and the instances $n$.

Example-based metrics evaluate the difference between the actual and predicted labelsets, averaged over $n$. Given a true labelset $Y_i = \{y_{i1}...y_{i L}\}$ and a predicted one $Z_i = \{ z_{i1}...z_{i L} \}$, the example-based metrics considered are:

\begin{equation}
    \begin{split}
    &\text{\footnotesize Subset accuracy} = \frac{1}{n}\sum^n_{i=0} 1 | Y_i = Z_i \\
    &\text{\footnotesize Hamming loss} = \frac{1}{n L} \sum^n_{i=0} \sum^L_{l=0} 1 | y_{il} \neq z_{il} \\
    &\text{\footnotesize Example-based precision} = \frac{1}{n} \sum^n_{i=0} \frac{|Y_i \cup Z_i|}{|Z_i|} \\
    &\text{\footnotesize Example-based recall} = \frac{1}{n} \sum^n_{i=0} \frac{|Y_i \cup Z_i|}{|Y_i|} \\
    &\text{\footnotesize Example-based F1} = \frac{1}{n} \sum^n_{i=0} \frac{2 |Y_i \cup Z_i|}{|Y_i| + |Z_i|}
    \end{split}
\end{equation}

Label-based metrics measure the performance across the different labels, that can be micro- or macro-averaged depending on if it is used the joint statistics for all labels or the per-label measures are averaged into a single value. Given for each label $l$ its true positives $tp_l = \sum^n_{i=0} 1 | y_{il} = z_{il} = 1$, true negatives $tn_l = \sum^n_{i=0} 1 | y_{il} = z_{il} = 0$, false positives $fp_l = \sum^n_{i=0} 1 | y_{il}=0, z_{il}=1$, and false negatives $fn_l = \sum^n_{i=0} 1 | y_{il}=1, z_{il}=0$, the macro-averaged metrics considered are:
\begin{equation}
\begin{split}
    &\text{\footnotesize Macro-averaged precision} = \frac{1}{L} \sum^L_{l=0} \frac{tp_l}{tp_l + fp_l}\\
    &\text{\footnotesize Macro-averaged recall} = \frac{1}{L} \sum^L_{l=0} \frac{tp_l}{tp_l + fn_l}
\end{split}
\end{equation}
while the considered micro-averaged metrics are defined as:
\begin{equation}
\begin{split}
&\text{\footnotesize Micro-averaged precision} = \frac{\sum^L_{l=0} tp_l}{\sum^L_{l=0} tp_l + \sum^L_{l=0} fp_l}\\
&\text{\footnotesize Micro-averaged recall} = \frac{\sum^L_{l=0} tp_l}{\sum^L_{l=0} tp_l + \sum^L_{l=0} fn_l}
\end{split}
\end{equation}

Additionally, Macro-averaged F1 and Micro-averaged F1 are also considered, defined as the harmonic mean of precision and recall in an equivalent way to the example-based F1 defined above.

\subsection{Analysis of components in MLHAT}\label{sec:optimization}

As discussed in the model description in Section \ref{sec:proposal}, \ac{MLHAT} is highly configurable through a wide variety of hyperparameters ranging from the entropy and cardinality thresholds to the classifiers to be used in the leaves and the metrics to be used to monitor the concept drift. The configuration of these hyperparameters is non-trivial, as it encompasses a multitude of possible combinations that evolve categorical, integer, decimal, and logarithmic parameters. In response to this challenge, we employ a well-known systematic approach for hyperparameter optimization \cite{Bischl2023} that combines the \ac{TPE} algorithm to generate configurations that maximize the objective function, with the Hyperband pruner that accelerates the search by discarding the less promising combinations. This methodology allows us to efficiently traverse the vast hyperparameter space of \ac{MLHAT}, as well as to understand the importance of each hyperparameter and its influence on the final performance. The study has been implemented in Python through the Optuna framework \cite{Akiba2019}.

\begin{table*}[htb]
    \centering
    \caption{Hyperparameter study for MLHAT}\label{tab:hyperparams}
    \begin{tblr}{
        width=.8\linewidth,
        colspec={Q[l,m] X[l,m] Q[c,m]},
        colsep=3pt,
        rowsep=1pt,
        row{1-Z}={font=\scriptsize},
    }
    \toprule
    Parameter & Studied values & Final Value\\
    \midrule
    Significance for split $\delta_{spl}$ & Fixed by state-of-the-art & 1e-7\\
    Leaf grace period $\kappa_{spl}$ & Fixed by state of the art & 200\\
    Significance for tree replacement $\delta_{alt}$ & [0.01, 0.10] & 0.05 \\
    Alternate tree grace period. $\kappa_{alt}$ & [100, 500] & 200\\
    Cardinalty threshold $\eta$ & [100, 1000] & 750\\
    Poisson parameter $\lambda$ & [1, 6] & 1\\
    Drift metric $e$ & Subset Acc., Hamming loss, Micro-F1, Macro-F1 & Hamming loss\\
    $\gamma_{\downarrow C}$ base & \acs{HT}, \acs{kNN}, \acs{NB}, \acs{LR}  & \acs{kNN}\\
    $\gamma_{\downarrow C}$ transformation & \acs{BR}, \acs{LP}, \acs{CC} & \acs{LP}\\
    $\gamma_{\uparrow C}$ base & \acs{NB}, \acs{LR} & \acs{LR}\\
    $\gamma_{\uparrow C}$ ensemble & \acs{BA} \cite{Oza2001}, \acs{BO} \cite{Oza2001}, \acs{BOLE} \cite{DeBarros2016}, \acs{SRP} \cite{Gomes2019} & \acs{BA}\\
    $\gamma_{\uparrow C}$ transformation & \acs{BR}, \acs{LP}, \acs{CC} & \acs{BR}\\
    \bottomrule
    \end{tblr}
\end{table*}

The conditions of the study are primarily determined by the optimization function and the range of each hyperparameter. The optimization function is defined as the Subset accuracy, measured through prequential evaluation (see Section \ref{sec:evaluation}). To make the iterative search feasible, a minimum subset of datasets was selected from the full experimental set. These datasets were chosen based on lower complexity and maximizing differences in terms of cardinality, density, and concept drift.
The selected datasets, representing 19\% of the total set used in the experiment, are: \textit{Flags}, \textit{Emotions}, \textit{VirusGO}, \textit{Birds}, \textit{Yeast}, \textit{Scene}, \textit{SynHPSud}, and \textit{SynHPGrad}. Details on these datasets can be found in Tables \ref{tab:datasets} and \ref{tab:datasets-synth}.

The search ranges for each hyperparameter are shown in Table \ref{tab:hyperparams}. For numerical attributes, typical ranges from Hoeffding tree literature have been used.
To monitor concept drift, any \ac{MLC} metric can be employed. However, it's preferable to use a metric that covers overall performance and is sensitive to class imbalance, as it will operate at each tree node level. In this study, two example metrics and two label metrics are tested.
For the leaf nodes, any pair of online multi-label classifiers can be used, provided they meet two criteria. First, they must not incorporate drift detection mechanisms, as this is done at the tree level. Second, they should balance short and long-term performance to cover various scenarios a leaf may encounter, as discussed in Section \ref{sec:leaves}.
The study examines different classifier complexities. For low-complexity classifiers, it represents four main paradigms in online classification: decision trees, distance-based models, Naïve Bayes, and linear models. For high-complexity classifiers, it compares different ensembles, using low-complexity methods as base models to control overall model complexity. All ensembles use 10 base models, as is standard in online learning paradigms \cite{Alberghini2022}.
Both low and high-complexity classifiers explore various transformations from multi-label to single-label space.

\begin{figure*}[htb]
    \centering
    \subfloat[Optimization history]{\includegraphics[width=.4\linewidth]{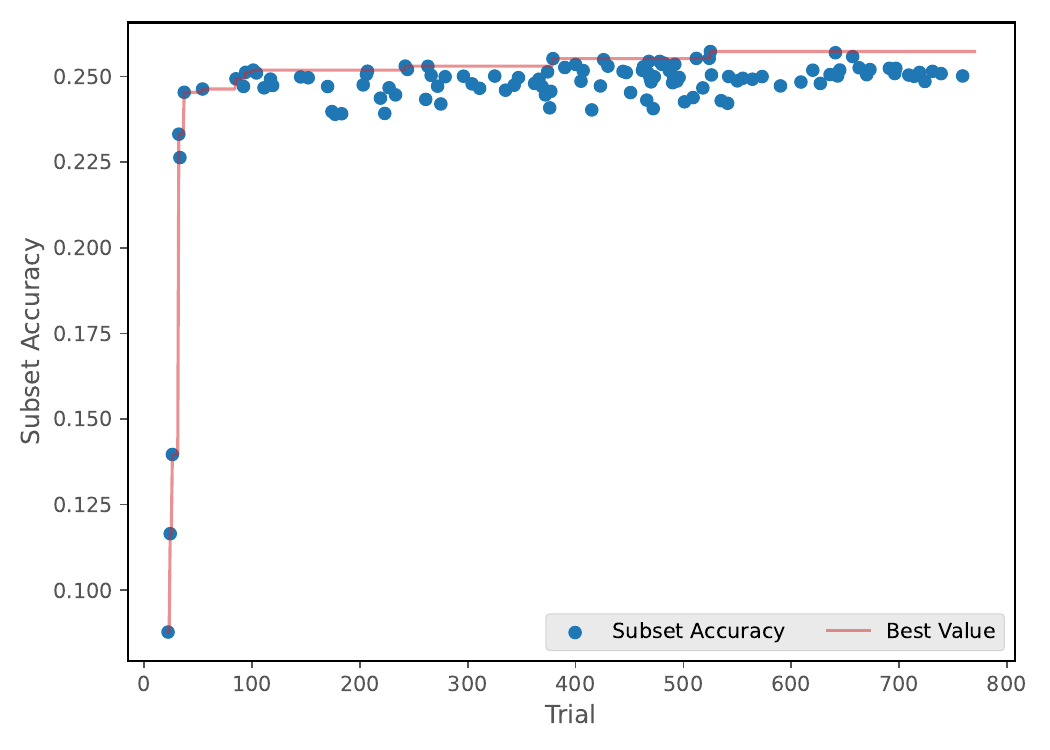}}\quad
    \subfloat[Hyperparameter importance according functional ANOVA]{\includegraphics[width=.4\linewidth]{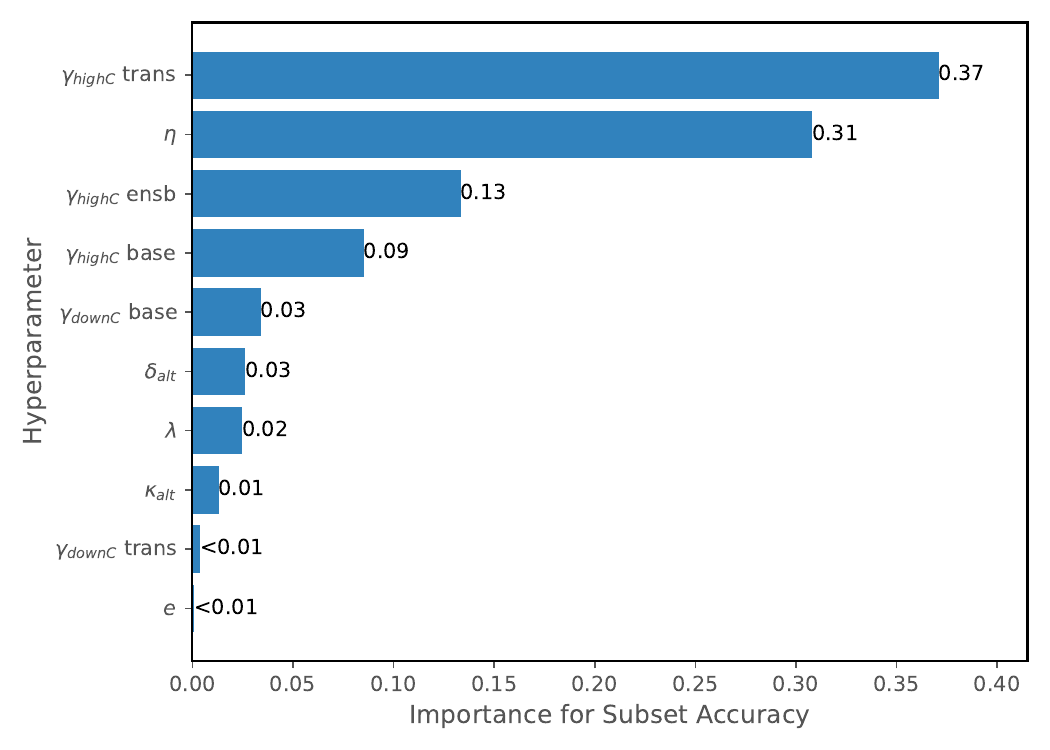}}\\
    \subfloat[Parallel coordinate plot with the most influential parameters. Some parameters have been omitted for visual clarity as they do not have a sufficiently meaningful range of variability.]{\includegraphics[width=.8\linewidth]{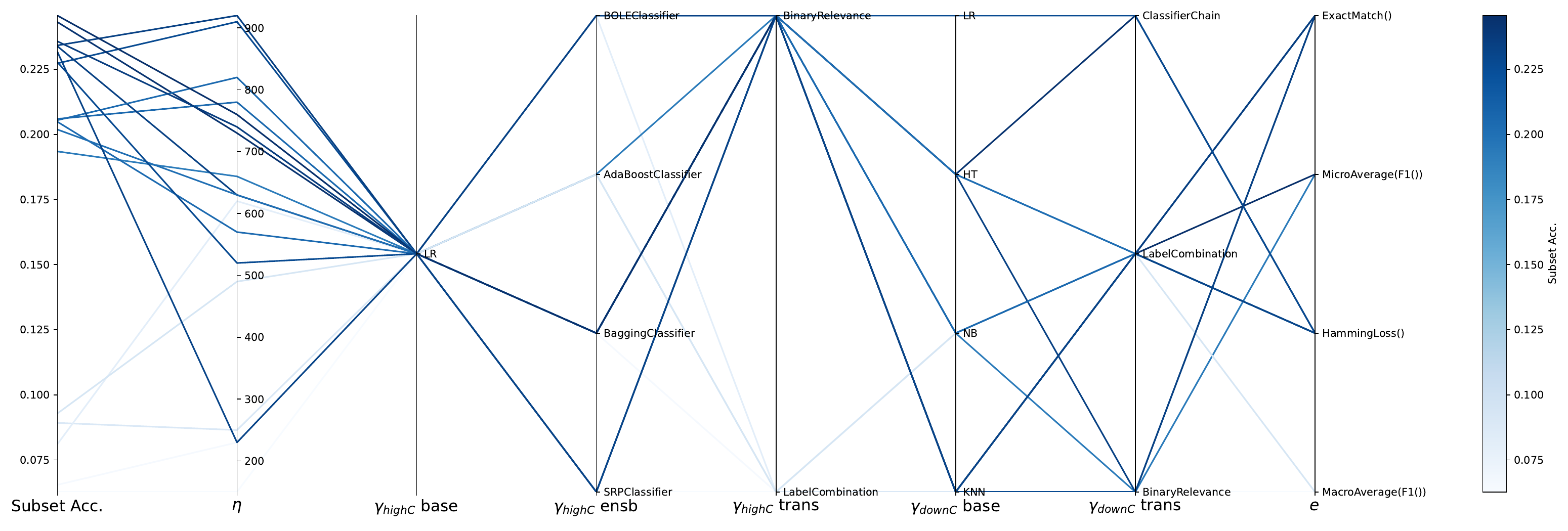}}
    \caption{Automatic optimization of MLHAT}
    \label{fig:hyperparams-study}
\end{figure*}

Figure \ref{fig:hyperparams-study} shows the results of the hyperparameters search, including the optimization history obtained with the framework described above, and the importance of each hyperparameter in the final performance according to the  ANOVA functional framework \cite{Hutter2014}, that quantifies the influence of both single hyperparameters and the interactions between them from the marginal predicted performance. Finally, the parallel coordinate plot shows the most promising hyperparameter combination based on  the average performance in the subset of datasets selected for this phase. Please note that not all combinations are present because of the guided search method employed. The best hyperparameter configuration found by the optimization method is shown in the last column of Table \ref{tab:hyperparams}. This configuration will be used in the rest of the experimental study over the whole set of datasets.

The study indicates that the high-complexity classifier is the most relevant parameter for \ac{MLHAT} performance, specifically the label transformation method, where \ac{BR} performs much better than \ac{LP} and \ac{CC}, probably because it deals more efficiently with label imbalance. In contrast, the transformation used in the low-complexity classifier does not have very important differences in the final result: when working with little data, the multi-label space is small and the three studied methods provide equivalent results. The second most important parameter is the threshold for establishing a high-cardinality scenario. Although the results will vary depending on the pair of classifiers used, it appears that around 600 instances received at a leaf node is the optimal point to move to the most complex classifier. When dealing with the highly complex scenario, it becomes evident that bagging-based ensembles, represented by \ac{BA} and \ac{SRP}, offer more advantages compared to boosting-based ensembles, namely \ac{BO} and \ac{BOLE}. This observation arises from the fact that boosting primarily concentrates on enhancing the performance of misclassified instances. However, in the context of \ac{MLHAT}, where the feature space has already undergone discrimination through the tree structure, this boosting emphasis may result in overfitting. As a base classifier, \ac{LR} far outperforms \ac{NB}, which, having not completed any runs because they were pruned earlier, does not appear in the graph. In the low complexity case, working with a single model and not an ensemble, it is required a more sophisticated learner to obtain good results. Thus, \ac{HT} and \ac{kNN} are the most promising. Finally, attending to the metric for monitoring concept drift, although not as influential as other parameters, there is a clear loser: macro-average is not a good approach, as it does not take into account label imbalance. The rest of the metrics have a similar potential to detect concept drift in general, although probably in studies where only a specific type of concept drift is affected, one metric would respond better than another to accelerate detection. The final configuration for \ac{MLHAT}, showed in Table \ref{tab:hyperparams}, follows these lines: for the low complexity leaf classifier a multi-label \ac{kNN} obtains better performance than the closest alternative of \ac{HT}. The \ac{LP} transformation is preferred due to the possibility to capture better the label correlation than \ac{BR}. The high complexity classifier is conformed as a bagging of \acp{LR}, that combines good performance and fast execution in data-intense scenarios. Hamming loss is the metric selected for monitoring the concept drift. 
 This metric that maintains a balance between the proportion of correct and incorrect labels inside the labelset, rather than relying on subset accuracy, which is too strict to be informative. With this configuration, 620 instances are the optimal threshold a leaf should see to pass from low to high cardinality scenarios, and 450 are the minimum instances an alternate tree should see to try to replace the main tree. Finally, the $Poisson(\lambda=1)$ indicates that is desirable to model the instances' weight as events that occur relatively infrequently but with a consistent average rate.

\begin{figure*}[htb]
    \centering
    \includegraphics[width=.8\linewidth]{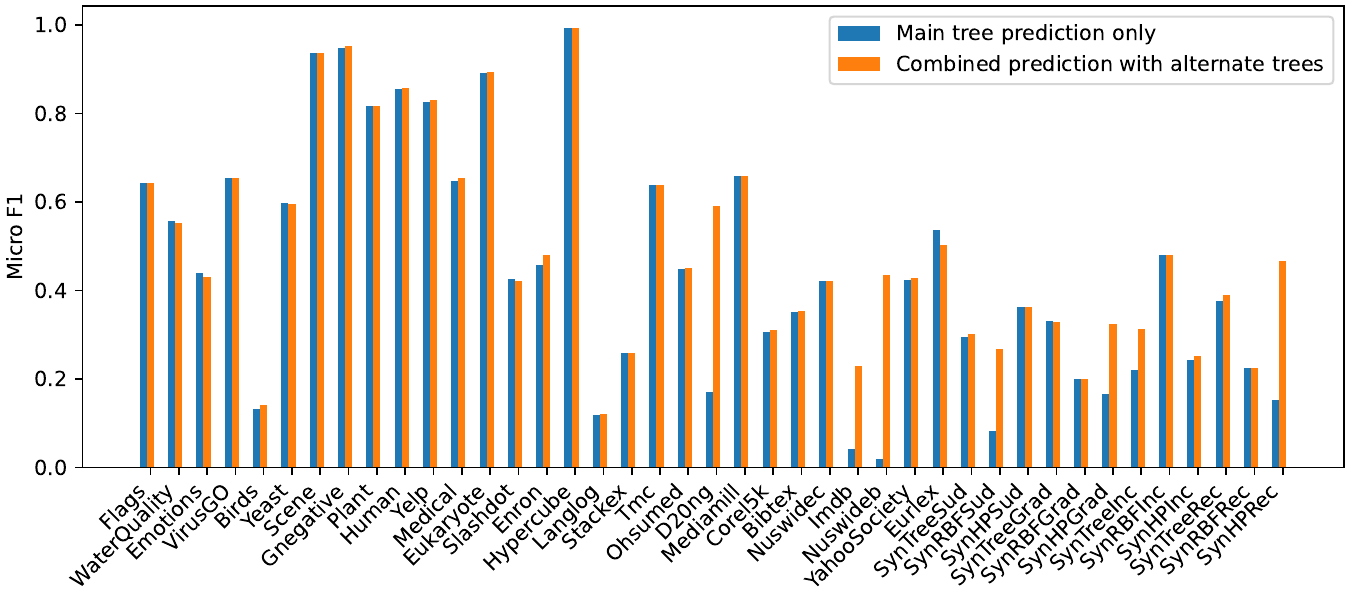}
    \caption{Effect on the performance of combining predictions from the main and the alternate trees in \ac{MLHAT}}
    \label{fig:combpred}
\end{figure*}

Once the \ac{MLHAT} hyperparameters have been established, the last study examines the effect of combining the predictions of the alternate trees that are constructed during the occurrence of concept-changes explained in Section \ref{sec:classification}. This study focuses on the micro-average F1 score obtained with the fixed parameters of Table \ref{tab:hyperparams} and altering only the way of obtaining the predictions during the prequential evaluation. Figure \ref{fig:combpred} shows the differences between results per dataset. Combining the predictions gives an average F1 score of 0.4916, while using the main tree alone gives 0.4475. In other words, the method proposed in Section \ref{sec:proposal} improves the results on average by 5\%. The results per dataset will depend on whether \ac{MLHAT} detects the concept drift alarm and whether it is finally confirmed or not. In most datasets, this measure has little effect because either the concept drift does not occur or it is quickly confirmed and the alternate structure becomes part of the main tree. In any case, since the grace period $\kappa_{alt}$ must be exceeded to consider the predictions of the alternate trees, and the predictions are weighted with the branch error in real time, the inclusion of the possible alternate trees is done in a fairly conservative manner that prevents it from having significant adverse effects. Moreover, we see datasets such as \textit{D20ng}, \textit{Nuwideb} or synthetic datasets with explicit concept drift information where combining the predictions has a very positive effect on the average result, with differences of up to 30 times in F1. This is due to the fact that in these cases there are more gradual concept drifts in which it is positive to consider at the same time the main tree and the alternatively emerging branches. In conclusion, this measure is considered positive for the overall performance of MLHAT and will be used in the remainder of the comparative study.

\subsection{Analysis of MLHAT compared to previous proposals}\label{sec:overallperf}

\begin{figure*}[htb]
    \centering
    \subfloat[Subset Accuracy]{\includegraphics[width=.49\linewidth]{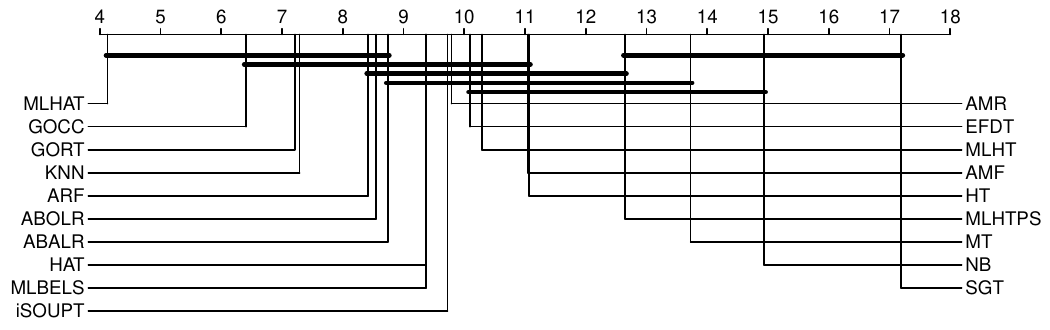}}
    \subfloat[Hamming loss]{\includegraphics[width=.49\linewidth]{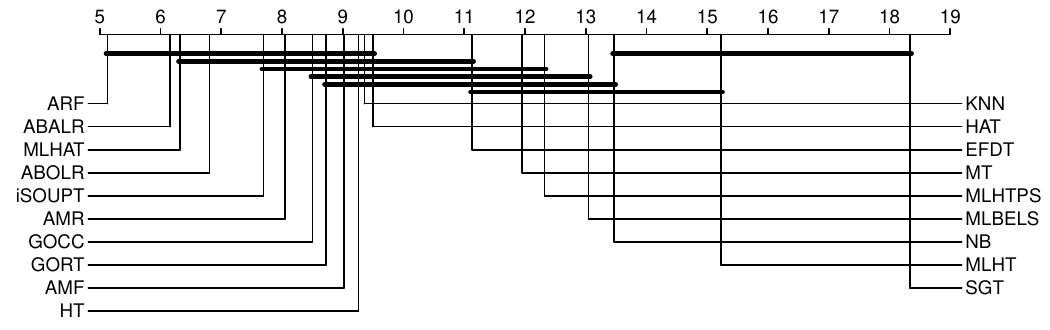}}\\
    \subfloat[Example-based F1]{\includegraphics[width=.49\linewidth]{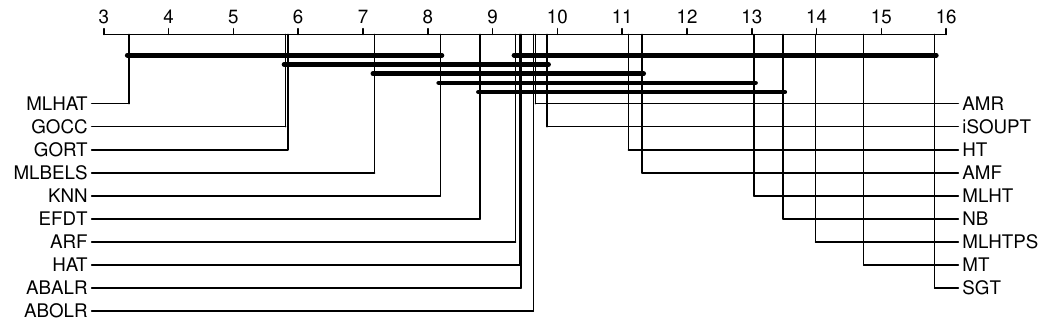}}
    \subfloat[Micro-averaged F1]{\includegraphics[width=.49\linewidth]{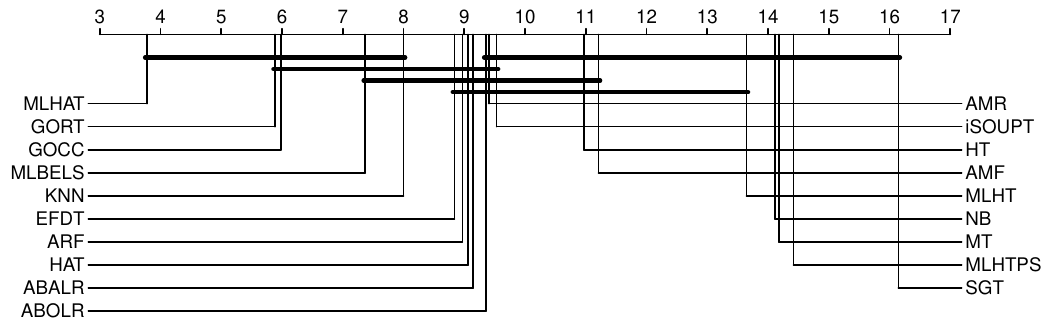}}\\
    \subfloat[Macro-averaged F1]{\includegraphics[width=.49\linewidth]{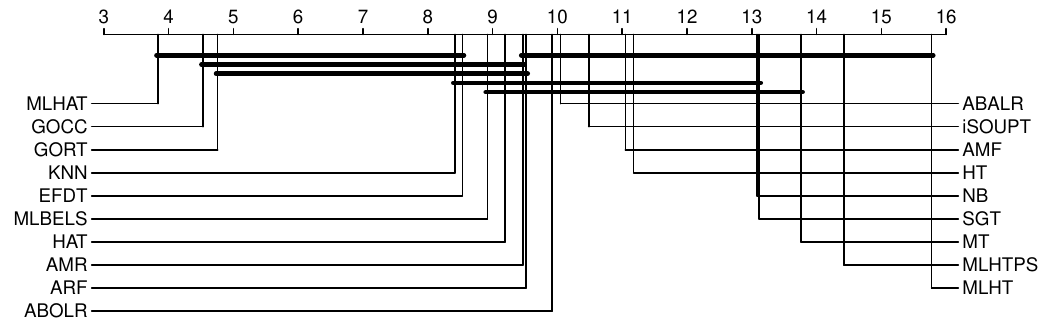}}
    \subfloat[Time (s)]{\includegraphics[width=.49\linewidth]{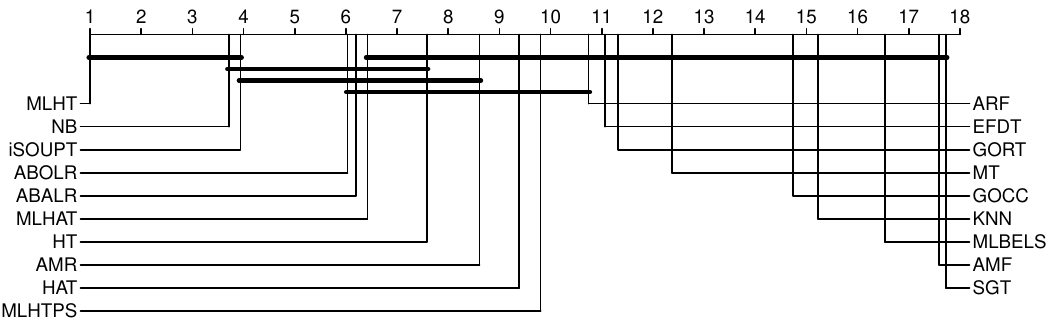}}
    
    \caption{Critical distance of Bonferroni-Dunn’s post-hoc tests ($\alpha$ = 99\%)}\label{fig:post-hoc}
\end{figure*}

This section evaluates and compares the overall performance of the 19 online algorithms in the 31 real world datasets for the 11 \ac{MLC} metrics, plus time of execution, to address RQ1. The analysis is organized as follow.
Table \ref{tab:datasetf1} analyzes in detail the results broken down by model and dataset, showing the detail of one of the \ac{MLC} metrics of the study: the example-based F1. We have chosen to show this metric in detail due to its relevance in the \ac{MLC} field as it adequately summarizes the balance between sensitivity and specificity of the different labels that make up the dataset. Due to space limitations and the large number of methods and datasets included in the experimentation, the details of the rest of the metrics included in the study are available in the repository associated with the work\footnote{\url{https://github.com/aestebant/mlhat}}. The last two rows of Table \ref{tab:datasetf1} show the summary of the metric under study in two useful ways to get an idea of the overall performance of each method: On the one hand, it is indicated the average that each model has obtained taking into account all the datasets that make up the body of the table. On the other hand, the ranking of each model is shown as the position it obtains with respect to the other methods in the comparison, averaged over all the datasets that make up the body of the table. These two statistics are in turn obtained for the rest of the \ac{MLC} metrics that make up the study, and are shown in Tables \ref{tab:singleresults} and \ref{tab:friedmansingle} respectively. Thus, Table \ref{tab:datasetf1} serves to make a detailed study of the particularities of each model by type of dataset, while Table \ref{tab:singleresults} and \ref{tab:friedmansingle} show the general trends by approach and family of algorithms. 
In addition, our experiments are supported with statistical tests that validate the significant differences between methods. Specifically, the Friedman test \cite{Dietterich1998} is applied for each metric under study with the rankings in Table \ref{tab:friedmansingle}. Friedman's test shows that there are significant differences with high confidence ($p\mathrm{-value} \rightarrow 0$) for all metrics. Therefore, the post-hoc Bonferroni-Dunn procedure \cite{Dietterich1998} is applied to a selection of the most representative metrics to find between which groups of algorithms these differences occur. Figure \ref{fig:post-hoc} shows the results at confidence level $\alpha=99\%$ for subset accuracy, hamming loss, example-based F1, micro-averaged F1, macro-averaged F1, and time.

\begin{table*}[!t]
    \caption{Results for Example-based F1 on each dataset}\label{tab:datasetf1}
    \begin{tblr}{
        width=\linewidth,
        colspec={Q[l,m] *{19}{X[c,m]}},
        rowsep=-1pt,
        colsep=0pt,
        row{1-Z}={font=\scriptsize},
    }
\toprule
\SetCell[r=2]{} Dataset & MLHAT & &  NB &  & HT &  &  EFDT &  &  MT &  & AMF &  &  ABOLR & & GORT & & MLHTPS & &  MLBELS \\
 &  & KNN & & AMR & & HAT &  &  SGT &  & ARF &  & ABALR &  & GOCC & & MLHT &  & iSOUPT \\
\midrule
Flags & 0.605 & 0.619 & 0.635 & 0.554 & \textbf{0.696} & 0.677 & \textbf{0.696} & 0.634 & 0.613 & 0.694 & 0.617 & 0.561 & 0.557 & 0.553 & 0.557 & 0.615 & 0.628 & 0.554 & 0.450 \\
WQ & \textbf{0.533} & 0.524 & 0.525 & 0.528 & 0.394 & 0.406 & 0.430 & 0.259 & 0.330 & 0.503 & 0.480 & 0.524 & 0.528 & 0.530 & 0.517 & 0.239 & 0.361 & 0.522 & 0.428 \\
Emo & 0.446 & 0.445 & \textbf{0.601} & 0.291 & 0.544 & 0.542 & 0.492 & 0.318 & 0.019 & 0.461 & 0.338 & 0.273 & 0.283 & 0.294 & 0.288 & 0.333 & 0.477 & 0.291 & 0.514 \\
Virus & 0.730 & 0.702 & 0.095 & 0.561 & 0.196 & 0.439 & 0.196 & 0.355 & 0.428 & 0.373 & \textbf{0.841} & 0.539 & 0.561 & 0.577 & 0.537 & 0.337 & 0.164 & 0.547 & 0.520 \\
Birds & \textbf{0.482} & 0.062 & 0.000 & 0.089 & 0.000 & 0.000 & 0.038 & 0.026 & 0.023 & 0.039 & 0.041 & 0.072 & 0.070 & 0.377 & 0.380 & 0.000 & 0.000 & 0.089 & 0.134 \\
Yeast & 0.588 & 0.562 & 0.533 & 0.464 & 0.582 & 0.575 & 0.570 & 0.422 & 0.454 & 0.548 & 0.461 & 0.463 & 0.465 & 0.563 & 0.503 & 0.519 & 0.551 & 0.476 & \textbf{0.591} \\
Scene & \textbf{0.948} & 0.811 & 0.611 & 0.916 & 0.419 & 0.533 & 0.646 & 0.193 & 0.538 & 0.803 & 0.812 & 0.913 & 0.912 & 0.933 & 0.933 & 0.184 & 0.622 & 0.876 & 0.786 \\
Gneg & 0.913 & 0.724 & 0.682 & 0.948 & 0.623 & 0.632 & 0.621 & 0.205 & 0.678 & 0.728 & 0.865 & 0.947 & 0.946 & \textbf{0.953} & \textbf{0.953} & 0.392 & 0.628 & 0.950 & 0.696 \\
CAL500 & \textbf{0.351} & 0.349 & 0.348 & 0.327 & 0.314 & 0.305 & 0.314 & 0.280 & 0.305 & 0.317 & 0.325 & 0.330 & 0.325 & 0.327 & 0.322 & 0.338 & 0.297 & 0.327 & 0.318 \\
Human & 0.885 & 0.588 & 0.348 & 0.896 & 0.358 & 0.478 & 0.410 & 0.095 & 0.209 & 0.672 & 0.467 & 0.890 & 0.894 & 0.904 & \textbf{0.904} & 0.295 & 0.234 & 0.862 & 0.678 \\
Yelp & \textbf{0.836} & 0.631 & 0.373 & 0.705 & 0.552 & 0.617 & 0.650 & 0.388 & 0.453 & 0.758 & 0.616 & 0.741 & 0.743 & 0.770 & 0.771 & 0.438 & 0.510 & 0.723 & 0.666 \\
Med & 0.539 & 0.419 & 0.002 & 0.199 & 0.375 & 0.395 & 0.525 & 0.129 & 0.196 & 0.135 & 0.453 & 0.130 & 0.121 & 0.212 & 0.166 & 0.231 & 0.002 & 0.217 & \textbf{0.667} \\
Eukar & 0.912 & 0.726 & 0.317 & 0.903 & 0.509 & 0.682 & 0.617 & 0.038 & 0.403 & 0.786 & 0.579 & 0.912 & 0.911 & 0.922 & \textbf{0.922} & 0.260 & 0.207 & 0.882 & 0.790 \\
Slashdot & 0.361 & 0.149 & 0.000 & 0.018 & 0.060 & 0.108 & 0.159 & 0.049 & 0.007 & 0.085 & 0.175 & 0.007 & 0.006 & 0.058 & 0.077 & 0.150 & 0.006 & 0.045 & \textbf{0.421} \\
HS & 0.855 & 0.846 & 0.401 & 0.839 & 0.726 & 0.873 & 0.820 & 0.324 & 0.587 & 0.867 & 0.614 & 0.853 & 0.853 & \textbf{0.892} & 0.879 & 0.501 & 0.562 & 0.689 & - \\
HC & 0.993 & 0.994 & 0.966 & 0.981 & 0.906 & 0.963 & 0.915 & 0.015 & 0.994 & 0.993 & \textbf{0.995} & 0.987 & 0.987 & 0.993 & 0.993 & 0.786 & 0.806 & 0.909 & 0.987 \\
Langlog & \textbf{0.221} & 0.047 & 0.000 & 0.030 & 0.000 & 0.000 & 0.054 & 0.013 & 0.006 & 0.002 & 0.006 & 0.018 & 0.019 & 0.198 & 0.199 & 0.001 & 0.029 & 0.033 & 0.125 \\
Stackex & 0.222 & 0.098 & 0.001 & 0.082 & 0.009 & 0.018 & 0.120 & 0.030 & 0.014 & 0.028 & 0.033 & 0.097 & 0.096 & 0.094 & 0.100 & 0.144 & 0.000 & 0.103 & \textbf{0.309} \\
Reuters & 0.034 & 0.165 & 0.000 & 0.001 & 0.113 & 0.129 & 0.171 & 0.012 & 0.047 & 0.066 & 0.092 & 0.000 & 0.000 & 0.035 & 0.027 & 0.069 & 0.002 & 0.000 & \textbf{0.460} \\
Tmc & 0.628 & 0.451 & 0.294 & 0.590 & 0.489 & 0.489 & 0.425 & 0.288 & 0.389 & 0.482 & 0.480 & 0.600 & 0.590 & 0.609 & 0.619 & 0.215 & 0.446 & 0.580 & \textbf{0.633} \\
Ohsum & \textbf{0.398} & 0.042 & 0.233 & 0.144 & 0.286 & 0.309 & 0.257 & 0.142 & 0.017 & 0.104 & 0.004 & 0.188 & 0.180 & 0.283 & 0.260 & 0.142 & 0.294 & 0.174 & 0.378 \\
D20ng & \textbf{0.600} & 0.090 & 0.000 & 0.079 & 0.329 & 0.307 & 0.375 & 0.124 & 0.067 & 0.108 & 0.066 & 0.208 & 0.185 & 0.367 & 0.327 & 0.083 & 0.259 & 0.162 & 0.530 \\
Media & \textbf{0.633} & 0.564 & 0.170 & 0.528 & 0.491 & 0.498 & 0.492 & 0.294 & 0.450 & 0.553 & 0.504 & 0.532 & 0.531 & 0.624 & 0.623 & 0.426 & 0.357 & 0.500 & 0.548 \\
Corel5k & 0.197 & 0.111 & 0.012 & 0.016 & 0.059 & 0.114 & 0.121 & 0.012 & 0.108 & 0.259 & 0.163 & 0.013 & 0.014 & \textbf{0.467} & \textbf{0.467} & 0.043 & 0.012 & 0.014 & 0.204 \\
Corel16k & 0.226 & 0.165 & 0.010 & 0.029 & 0.054 & 0.132 & 0.141 & 0.016 & 0.059 & 0.285 & 0.099 & 0.012 & 0.011 & 0.510 & \textbf{0.511} & 0.059 & 0.082 & 0.015 & 0.225 \\
Bibtex & \textbf{0.310} & 0.019 & 0.000 & 0.159 & 0.171 & 0.183 & 0.190 & 0.145 & 0.095 & 0.120 & 0.111 & 0.144 & 0.145 & 0.124 & 0.129 & 0.067 & 0.088 & 0.159 & 0.289 \\
NWC & \textbf{0.404} & 0.258 & 0.219 & 0.049 & 0.125 & 0.172 & 0.160 & - & - & 0.162 & - & 0.041 & 0.040 & 0.394 & 0.403 & 0.008 & 0.138 & 0.050 & - \\
Imdb & 0.192 & 0.141 & 0.175 & 0.017 & 0.029 & 0.035 & 0.060 & - & - & 0.023 & - & 0.039 & 0.042 & 0.198 & 0.196 & 0.252 & 0.006 & 0.027 & \textbf{0.347} \\
NWB & \textbf{0.431} & 0.247 & 0.106 & 0.060 & 0.085 & 0.174 & 0.177 & - & - & 0.173 & - & 0.058 & 0.057 & - & 0.387 & 0.006 & 0.050 & 0.048 & - \\
YahooS & 0.450 & 0.205 & 0.367 & 0.387 & 0.202 & 0.257 & 0.214 & - & - & 0.216 & - & 0.443 & 0.430 & 0.410 & 0.403 & 0.401 & 0.019 & 0.413 & \textbf{0.483} \\
Eurlex & 0.527 & 0.205 & 0.091 & 0.478 & 0.334 & 0.348 & 0.234 & - & 0.085 & 0.271 & - & \textbf{0.588} & 0.572 & 0.535 & 0.536 & 0.090 & 0.018 & 0.553 & - \\
\hline
Average & \textbf{0.531} & 0.386 & 0.262 & 0.383 & 0.324 & 0.367 & 0.364 & 0.185 & 0.280 & 0.375 & 0.394 & 0.391 & 0.390 & 0.493 & 0.484 & 0.246 & 0.253 & 0.380 & 0.488\\
Ranking & \textbf{3.387} & 8.196 & 13.484 & 9.661 & 11.097 & 9.419 & 8.806 & 15.823 & 14.726 & 9.355 & 11.306 & 9.435 & 9.629 & 5.806 & 5.839 & 13.032 & 13.984 & 9.839 & 7.177\\
\bottomrule
\SetCell[c=20]{l,m,\linewidth} Best results are in bold. \\
\SetCell[c=20]{l,m,\linewidth} Non finished experiments due to scalability problems marked with –. The physical limitations for all the experiments are set to 200GB of RAM and 360 hours of executions.\\
    \end{tblr}
\end{table*}

\begin{table*}[htb]
    \centering
    \caption{Average results for each evaluation metric considering all datasets}\label{tab:singleresults}
    \begin{tblr}{
        width=.95\linewidth,
        colspec={X[l,m] *{12}{Q[r,m]}},
        colsep=3pt,
        rowsep=0pt,
        row{1-Z}={font=\scriptsize}
    }
\toprule
Algorithm & Su. Acc & H. Loss & Ex. F1 & Ex. Pre & Ex. Rec & Mi. F1 & Mi. Pre & Mi. Rec & Ma. F1 & Ma. Pre & Ma. Rec & Time (s)\\
\midrule
MLHAT &      \textbf{0.3323} &   \textbf{0.0745} &      \textbf{0.5306} &        \textbf{0.7494} &       \textbf{0.5397} &    \textbf{0.5342} &      0.6296 &     \textbf{0.4910} &    \textbf{0.4064} &      \textbf{0.5899} &     \textbf{0.3833} &  17,355 \\
KNN &      0.2575 &   0.0850 &      0.3857 &        0.4238 &       0.3808 &    0.4198 &      0.5405 &     0.3746 &    0.2802 &      0.3953 &     0.2629 & 137,827 \\
NB &      0.1216 &   0.1499 &      0.2618 &        0.2597 &       0.3550 &    0.2725 &      0.3346 &     0.3620 &    0.1925 &      0.1920 &     0.2877 &  10,122 \\
AMR &      0.2534 &   0.0861 &      0.3828 &        0.4212 &       0.3826 &    0.4076 &      0.6589 &     0.3718 &    0.2911 &      0.3823 &     0.2764 &  39,651 \\
HT &      0.1902 &   0.0891 &      0.3236 &        0.3694 &       0.3205 &    0.3682 &      0.5291 &     0.3142 &    0.2309 &      0.3045 &     0.2190 &  56,964 \\
HAT &      0.2262 &   0.0842 &      0.3674 &        0.4135 &       0.3650 &    0.4095 &      0.5436 &     0.3558 &    0.2618 &      0.3326 &     0.2480 &  58,059 \\
EFDT &      0.2101 &   0.0857 &      0.3642 &        0.4066 &       0.3661 &    0.4095 &      0.5223 &     0.3599 &    0.2689 &      0.3283 &     0.2555 &  49,423 \\
SGT* &      0.0474 &   0.2863 &      0.1848 &        0.2096 &       0.3119 &    0.2059 &      0.2057 &     0.3162 &    0.1510 &      0.1566 &     0.3010 &  66,433 \\
MT* &      0.1593 &   0.1010 &      0.2804 &        0.3443 &       0.2652 &    0.3185 &      0.5954 &     0.2601 &    0.2053 &      0.3156 &     0.1925 &  61,381 \\
ARF &      0.2489 &   0.0750 &      0.3746 &        0.4307 &       0.3596 &    0.4164 &      \textbf{0.7160} &     0.3498 &    0.2714 &      0.4287 &     0.2490 &  36,633 \\
AMF* &      0.2458 &   0.0874 &      0.3938 &        0.4584 &       0.3786 &    0.4351 &      0.6870 &     0.3716 &    0.3026 &      0.4280 &     0.2801 & 119,653 \\
ABALR &      0.2702 &   0.0826 &      0.3911 &        0.4336 &       0.3856 &    0.4176 &      0.6859 &     0.3734 &    0.2986 &      0.3988 &     0.2774 &  16,920 \\
ABOLR &      0.2669 &   0.0837 &      0.3896 &        0.4305 &       0.3858 &    0.4144 &      0.6817 &     0.3734 &    0.2982 &      0.3969 &     0.2787 &  18,540 \\
GOCC* &      0.2980 &   0.0853 &      0.4935 &        0.6797 &       0.5103 &    0.5022 &      0.5801 &     0.4689 &    0.3928 &      0.5472 &     0.3687 &  74,899 \\
GORT &      0.2887 &   0.0844 &      0.4836 &        0.6848 &       0.5018 &    0.4920 &      0.5810 &     0.4565 &    0.3802 &      0.5407 &     0.3560 &  35,117 \\
MLHT &      0.1667 &   0.1247 &      0.2459 &        0.2816 &       0.2366 &    0.2476 &      0.3072 &     0.2251 &    0.1155 &      0.1370 &     0.1356 &   \textbf{1,934} \\
MLHTPS &      0.1588 &   0.1012 &      0.2534 &        0.3033 &       0.2465 &    0.2727 &      0.4522 &     0.2438 &    0.1749 &      0.2211 &     0.1772 &  39,472 \\
iSOUPT &      0.2510 &   0.0880 &      0.3803 &        0.4216 &       0.3779 &    0.4103 &      0.6762 &     0.3677 &    0.2748 &      0.3624 &     0.2596 &   8,577 \\
MLBELS* &      0.2779 &   0.1027 &      0.4880 &        0.5595 &       0.5251 &    0.4978 &      0.5234 &     0.4805 &    0.2977 &      0.5280 &     0.2973 & 110,060 \\
\bottomrule
\SetCell[c=13]{l} Best results are in bold.\\
\SetCell[c=13]{l,m,.9\linewidth} * Results computed without considering all datasets due to scalability limitations. The physical limitations for all the experiments are set to 200GB of RAM and 360 hours of executions.\\
    \end{tblr}
\end{table*}

\begin{table*}[htb]
    \centering
    \caption{Friedman's test and average ranks for each evaluation metric}\label{tab:friedmansingle} 
    \begin{tblr}{
        width=.9\linewidth,
        colspec={X[l,m] *{12}{Q[r,m]}},
        colsep=2pt,
        rowsep=0pt,
        row{1-Z}={font=\scriptsize},
    }
\toprule
Algorithm & Su.Acc & H.Loss & Ex.F1 & Ex.Pre & Ex.Rec & Mi.F1 & Mi.Prec & Mi.Rec & Ma.F1 & Ma.Prec & Ma.Rec & Time (s) \\
\midrule
MLHAT & \textbf{4.129} & 6.323 & \textbf{3.387} & \textbf{3.548} & \textbf{4.290} & \textbf{3.774} & 7.710 & \textbf{4.613} & \textbf{3.839} & \textbf{3.935} & \textbf{4.161} & 6.419 \\ 
KNN & 7.290 & 9.355 & 8.194 & 9.290 & 9.032 & 8.000 & 10.226 & 9.065 & 8.419 & 7.290 & 9.129 & 15.226 \\ 
NB & 14.935 & 13.468 & 13.484 & 15.194 & 11.710 & 14.113 & 15.016 & 11.387 & 13.081 & 14.629 & 11.677 & 3.710 \\ 
AMR & 9.790 & 8.048 & 9.661 & 9.855 & 10.081 & 9.403 & 7.242 & 9.919 & 9.468 & 8.823 & 9.726 & 8.613 \\ 
HT & 11.065 & 9.258 & 11.097 & 11.129 & 11.403 & 10.968 & 10.645 & 11.339 & 11.177 & 11.855 & 11.339 & 7.581 \\ 
HAT & 9.371 & 9.500 & 9.419 & 9.903 & 9.677 & 9.065 & 10.516 & 9.645 & 9.194 & 10.355 & 9.548 & 9.387 \\ 
EFDT & 10.097 & 11.129 & 8.806 & 9.226 & 8.984 & 8.839 & 11.129 & 8.887 & 8.532 & 10.435 & 8.887 & 11.065 \\ 
SGT & 17.194 & 18.339 & 15.823 & 16.177 & 11.919 & 16.145 & 17.758 & 11.435 & 13.113 & 16.016 & 10.435 & 17.726 \\ 
MT & 13.726 & 11.952 & 14.726 & 13.113 & 14.984 & 14.177 & 10.113 & 15.113 & 13.758 & 12.919 & 14.177 & 12.371 \\ 
ARF & 8.419 & \textbf{5.129} & 9.355 & 8.871 & 9.871 & 8.968 & 5.161 & 9.871 & 9.516 & 6.677 & 10.226 & 10.742 \\ 
AMF & 11.048 & 9.016 & 11.306 & 10.403 & 12.081 & 11.210 & 7.855 & 12.145 & 11.048 & 9.597 & 11.565 & 17.597 \\ 
ABALR & 8.742 & 6.161 & 9.435 & 9.371 & 10.177 & 9.145 & \textbf{5.210} & 10.177 & 10.048 & 8.371 & 10.403 & 6.194 \\ 
ABOLR & 8.548 & 6.806 & 9.629 & 9.468 & 10.048 & 9.355 & 5.581 & 10.177 & 9.919 & 8.661 & 10.048 & 6.032 \\ 
GOCC & 6.403 & 8.500 & 5.806 & 4.919 & 5.581 & 5.984 & 9.468 & 5.774 & 4.532 & 5.048 & 5.129 & 14.742 \\ 
GORT & 7.210 & 8.726 & 5.839 & 4.339 & 5.903 & 5.887 & 8.952 & 5.935 & 4.758 & 4.919 & 5.323 & 11.323 \\ 
MLHT & 10.290 & 15.226 & 13.032 & 13.871 & 13.194 & 13.645 & 15.806 & 13.355 & 15.774 & 16.871 & 14.677 & \textbf{1.000} \\ 
MLHTPS & 12.645 & 12.323 & 13.984 & 13.339 & 14.113 & 14.419 & 12.516 & 14.242 & 14.419 & 15.000 & 14.210 & 9.806 \\ 
iSOUPT & 9.726 & 7.694 & 9.839 & 10.161 & 10.194 & 9.532 & 6.629 & 10.000 & 10.484 & 10.419 & 10.548 & 3.935 \\ 
MLBELS & 9.371 & 13.048 & 7.177 & 7.823 & 6.758 & 7.371 & 12.468 & 6.919 & 8.919 & 8.177 & 8.790 & 16.532 \\  
\hline
Friedman's $\chi^{2}$ & 165.39 & 201.67 & 185.01 & 205.32 & 143.37 & 190.97 & 224.76 & 138.02 & 187.20 & 251.68 & 145.12 & 418.51\\
$p$-value & 2.2e-16 & 2.2e-16 & 2.2e-16 & 2.2e-16 & 2.2e-16 & 2.2e-16 & 2.2e-16 & 2.2e-16 & 2.2e-16 & 2.2e-16 & 2.2e-16 & 2.2e-16 \\
\bottomrule
\SetCell[c=13]{l} Best results are in bold.\\
    \end{tblr}
\end{table*}

The results show that MLHAT outperforms on average the other methods in accurate prediction of the labelset of each instance with subset accuracy of 33.23\%, as well as the best balance between false positives and false negatives at the instance level (example-based F1 of 53.06\%), and at the label level both absolute (macro F1 of 40.64\%) and prevalence-weighted (micro F1 of 53.42\%). These differences are supported by statistical tests, which indicate that for these metrics MLHAT outperforms the baseline of the Bayesian and rules approaches, NB and AMR, as well as non-Hoeffding \acp{IDT} SGT and MT. This indicates that most single-model proposals do not deal well with the additional complexity of \ac{MLC} transformed with \ac{BR} in terms of predicting the exact labelset, the average match by instances, nor the average match by labels. In this line, MLBELS and iSOUPT obtain better results than these proposals, showing the potential to adapt specific components of the decision tree to the multi-target paradigm. In the same way, we see that GOCC and GORT, that also incorporate specific \ac{MLC} mechanisms, are also very competitive. However, these methods obtain worse results than MLHAT with differences per metric. In the case of MLBELS, it is observed that it has problems matching exactly the predicted labelset, although it maintains a good average of hits per label. This makes it significantly worse than MLHAT in subset accuracy, hamming loss, and macro F1, while in Micro F1 and example-based F1, it is at the same level. This may be due to the method used for keeping the co-relation between labels, which is modeled with a neural network. For iSOUPT, the results are worse because it does not control the concept drift as \ac{MLHAT} does. This factor also makes \ac{MLHAT} outperform previous Hoeffding trees, although they will be discussed in detail in Section \ref{sec:treescomp}. GOCC and GORT are also close to \ac{MLHAT} in subset accuracy and F1 score, as they are potent ensembles that combine several \ac{IDT} geometrically averaged. However, combining 10 base trees still performs worse than MLHAT because they do not incorporate their specific tree-level adaptations. kNN works quite well without the need to explicitly consider the concept drift because it already works with a window of the \textit{n} most recent instances. Finally, ensembles in general also perform better because they combine several classifiers and incorporate concept drift detectors usually. Thus, statistical tests indicate that \ac{MLHAT} only may not differ from kNN and ensemble-based methods like MLBELS, GOCC, ARF, or ABA in different metrics, which are significantly more complex as the differences in time indicate. Although when this occurs, there is no algorithm that excels in more than one metric or that achieves better results on average, so we can state that there is no clear alternative to \ac{MLHAT} in terms of overall accuracy.

On the other hand, the results indicate a weakness of MLHAT in terms of the accuracy of the positive labels, that is, it tends to predict an accurate subset of the real labelset, but without including all the active labels. This affects the results in Hamming loss, making MLHAT not statistically superior to the most potent previous methods based on ensembles. Although it is not statistically inferior either. This also affects the micro-averaged precision, where MLHAT does not obtain the best average results either.
Thus, it can be seen that obtaining a good accuracy across instances, or in other words, minimizing Hamming loss, is a complicated problem, where the performance of the main proposals is blurred. Thus, in this metric, the ensemble-based methods of \ac{ABA} and \ac{ARF} obtain the best results, although MLHAT obtains good enough recall results to be superior overall through the F1 score.
Other models like \ac{iSOUPT} and \ac{AMR} are also at the same level of performance in this metric according to statistical tests. However, we can see that these algorithms obtain these results at the cost of worse recall rates, while MLHAT remains competitive in this metric in all scenarios, making the overall performance in the end superior.

Regarding the real-time factor, the results show that MLHAT is in the middle part of the ranking, close to the \ac{BR}+\ac{HAT} transformation. It is slower than the rest of the multi-label \acp{IDT} and other less sophisticated proposals, such as NB or \ac{LR}'s ensembles. On the other hand, it is faster than non-Hoeffding \acp{IDT} SGT and MT, which have scalability problems for larger datasets. This is accentuated in the forest approach AMF.
MLHAT is also faster than the most similar performing methods: kNN, ARF, and MLBELS.
On the one hand, kNN is directly dependent on the number of features and instances. Although \ac{MLHAT} also maintains \acp{kNN} in the leaves, the initial partition of the problem using the tree structure considerably accelerates the process. ARF, GOCC and GORT are slower because they assembles \acp{IDT} in a more computationally demanding system with concept drift detectors at tree level, while \ac{MLHAT} drift detectors operate at branch level, maintaining smaller alternate models. Finally, MLBELS has scalability issues that cause larger datasets to fail. This is due to the combination of an ensemble that maintains as many models as labels, and a mechanism that maintains a pool of discarded models that increases up to 100 per label.
Section \ref{sec:complexcomp} will discuss this factor in more detail in conjunction with the computational complexity and memory consumption evolution of the model.

The results per dataset for a specific metric from Table \ref{tab:datasetf1} confirm that \ac{MLHAT} is not only better on average but consistently outperforms for datasets with varying characteristics, being the best in 12 of 31 data sets and ranking in the first quartile in 23 of 31 data sets. Beyond comparison, these results help identify the problems where \ac{MLHAT} performs best and where it may be more appropriate to use other algorithms. Thus, we can see that for extremely short flows, less than 1000 instances, \ac{MLHAT} works well in high-dimensionality cases such as \textit{Medical} and \textit{VirusGO} where there are more features than instances. If there are fewer features, other simpler models perform better. As other authors discussed \cite{Read2012}, this is due to Hoeffding trees being conservative models that need a large number of instances to grow.  For longer streams, \ac{MLHAT} generally performs very well, as the tree can grow properly. However, if the number of features is low (less than 500), ensemble-based methods like \ac{ABA} or \ac{ARF} can obtain better results, as seen in the cases of \textit{EukaryotePseACC}, and \textit{Corel5k}. \ac{MLHAT} performs better than the other algorithms in high dimensional spaces, probably because of the partitioning of the truly multi-label feature space. By considering label co-occurrence in the decision tree partitioning process, it better handles the additional complexity of this scenario compared to transforming the problem or using other multi-label decision trees that do not incorporate Bernoulli at this stage. Additionally, keeping two types of classifiers in the leaves allows \ac{MLHAT} to remain competitive across streams of all lengths: in shorter cases, \ac{MLHAT}'s behavior more closely mimics a kNN, while in longer ones, it approximates an \ac{ABA}. 
Finally, we can also highlight the good performance of \ac{MLHAT} on unbalanced datasets (given by a high MeanIR) like \textit{Mediamill}, \textit{YahooSociety}, \textit{EurlexSM} or \textit{NuwWideCVLADplus} among others. This is also a consequence of adapting the leaves classifier to the cardinality of the labels seen so far, which prevents overfitting of the majority labels and, on the other hand, moves to a more advanced classifier when the complexity of the dataset requires it.

\subsection{Analysis of MLHAT compared to previous Hoeffding trees}\label{sec:treescomp}

After comparing \ac{MLHAT} with all other online \ac{MLC} paradigms, the second study addresses RQ2 by focusing the comparison on previous Hoeffding \acp{IDT} and specifically \ac{HAT}, an algorithm with several common elements with \ac{MLHAT} but ignoring the additional complexity of multi-label learning. For this analysis, Tables \ref{tab:singleresults}, \ref{tab:friedmansingle}, and \ref{tab:datasetf1}, and Figure \ref{fig:post-hoc} previously discussed are analyzed putting the focus on the Hoeffding trees.

As has been shown in the previous section, \acs{MLHAT} is 10.61\% better at predicting the exact labelset for an instance than the transformation \ac{BR}+\ac{HAT}, and around 15\% better in the different versions of the F1 score, indicating better accuracy per label. These differences are confirmed by Bonferroni-Dunn, that indicates the superiority of \ac{MLHAT} in all the metrics except for Hamming loss and the time of execution, in which they are equivalent.
This implies that introducing a natively multi-label split criterion that takes into account label co-occurrence through a multivariate Bernoulli distribution is a more effective approach than trying to classify each label separately with \ac{BR}+\ac{HAT}. In addition, considering this occurrence allows our model to adapt to the label imbalance that affects the number of labels that each leaf separates, which also contributes to increase the benefits of \ac{MLHAT} over HAT. Datasets with explicit temporal information generally show that  \ac{MLHAT} is more responsive to possible concept drifts than HAT. This means that, although both use \ac{ADWIN}-based branch-level detectors, \ac{MLHAT}'s approach of considering all labels together obtains better results than considering the accuracy of each label individually. However, as the \ac{RBF} generator-based datasets indicate, the HAT drift detector may perform better with data following this distribution. Finally, in terms of execution time, \ac{MLHAT} is on average 2.7 times faster than \ac{HAT}. This is because in \ac{BR}, a separated tree must be maintained for each label, which is slower than maintaining a single tree, even though it has an extra computational load due to the leaf classifiers.

HT and EFDT are Hoeffding trees that have the same problems as HAT due to not adapting the algorithm to the multi-label context. In addition,  they slightly worsen the results by not incorporating concept drift detection. In particular, the way of applying the Hoeffding bound \'with less guarantees" of EFDT versus \ac{HT} seems to benefit it more, since it compensates for the absence of drift adaptation with a faster tree growth method. \ac{MLHAT} applies the Hoeffding bound more similarly to \ac{HT} and \ac{HAT}, although the other modifications and components specific to the multi-label problem have more effect on the performance of the algorithm, so it obtains better results than EFDT, and also than HT, in all  criteria.

Attending to previous Hoeffding trees adaptations to \ac{MLC}, MLHT and MLHTPS, the results show that \ac{MLHAT} outperforms them in all performance metrics. In these cases, we observe how the results penalize a partitioning criterion that does not consider the co-occurrence of labels, as well as the absence of concept drift adaptation. Thus, we see that only a multi-label classifier on the leaves is not enough to obtain competitive results. In this line, it can be extracted from the results in Table \ref{tab:datasetf1} that, in general, the more complex MLHTPS leaf classifier is more beneficial.
On the other hand, the differences in the execution times of MLHT and MLHTPS allow us to quantify the influence of the leaf classifier on the real-time factor in these models. Thus, we can see that, while MLHT is the fastest model in the experiment, the change from the base classifier to an expensive model such as the Prune Set ensemble of \acp{HT} increases the execution time by about 20 times on average. This gives us an idea of the importance of the classifiers in MLHAT for the complexity of the model. Thus, the \ac{MLHAT} time metrics show that in this sense it is more beneficial to maintain two less computationally expensive classifiers such as kNN and ABA+\ac{LR} for each leaf.

\subsection{Analysis of concept drift adaptation}\label{sec:driftcomp}

This section addresses RQ3 by making a detailed analysis of the evolution of \ac{MLHAT} performance along the data stream and how it responds to different concept drift. 
In this study, we focus on a comprehensive analysis of the evolution of the performance across the data stream. Thus, we include a subset of the real-world datasets with a temporal component that allows us to study complex concept drifts that occur in real scenarios. To conduct a categorized study, and because there are no available multi-label datasets that include explicit concept drift information, synthetic datasets have been generated for this study, as mentioned in Section \ref{sec:dataset}, in order to evaluate the performance of MLHAT under concept drifts of different types. 
Finally, to facilitate the presentation of results, this section focuses the comparative study on the most competitive algorithms according to the results of Section \ref{sec:overallperf} and tries to include only one representative in the case of families with more than one algorithm. Thus, HAT, kNN and AMR are included as the most competitive \ac{BR} transformations of single models; ABA and ARF represent the \ac{BR} transformations of ensembles; and MLBELS, iSOUPT and GORT cover the previous \ac{AA}.

\begin{figure*}[!tb]
    \centering
    \subfloat[Yelp]{\includegraphics[width=.248\linewidth]{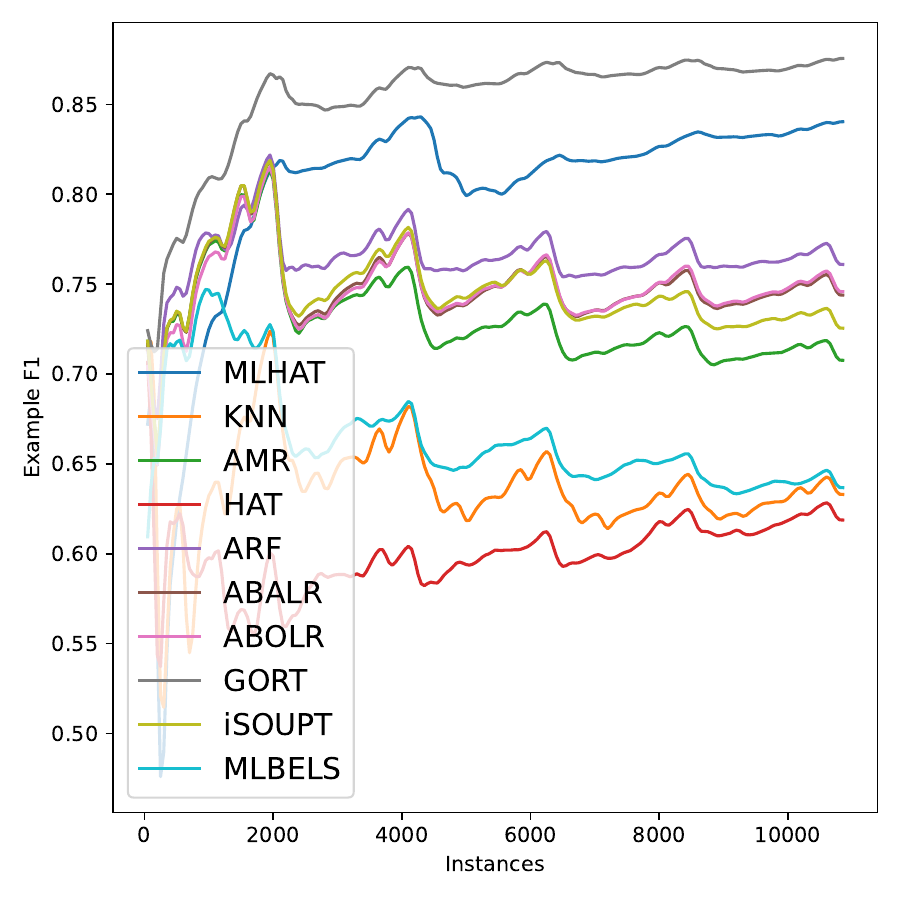}}
    \subfloat[Ohsumed]{\includegraphics[width=.248\linewidth]{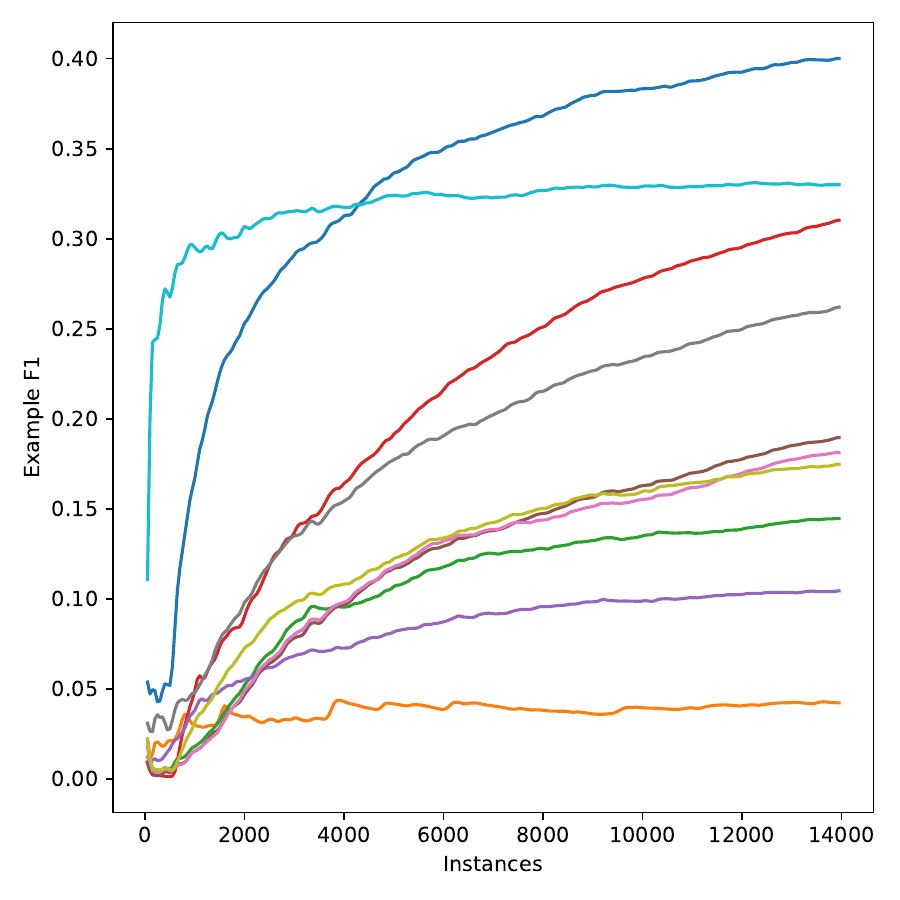}}
    \subfloat[D20ng]{\includegraphics[width=.248\linewidth]{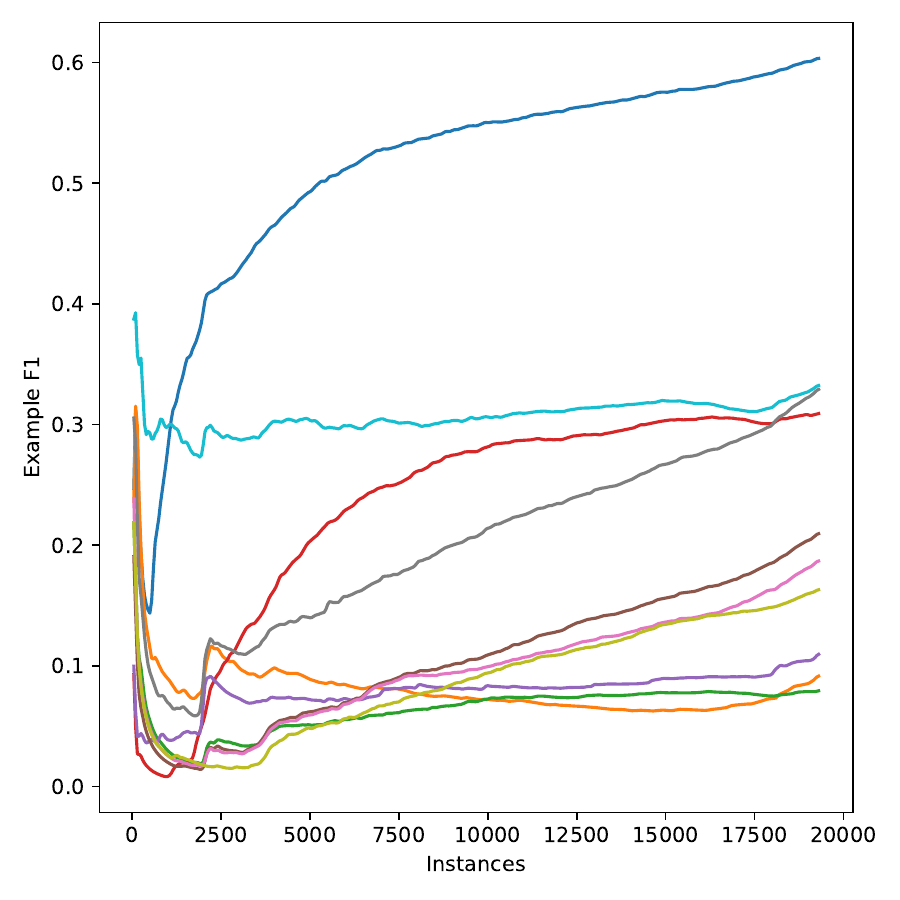}}
    \subfloat[NusWideBow]{\includegraphics[width=.248\linewidth]{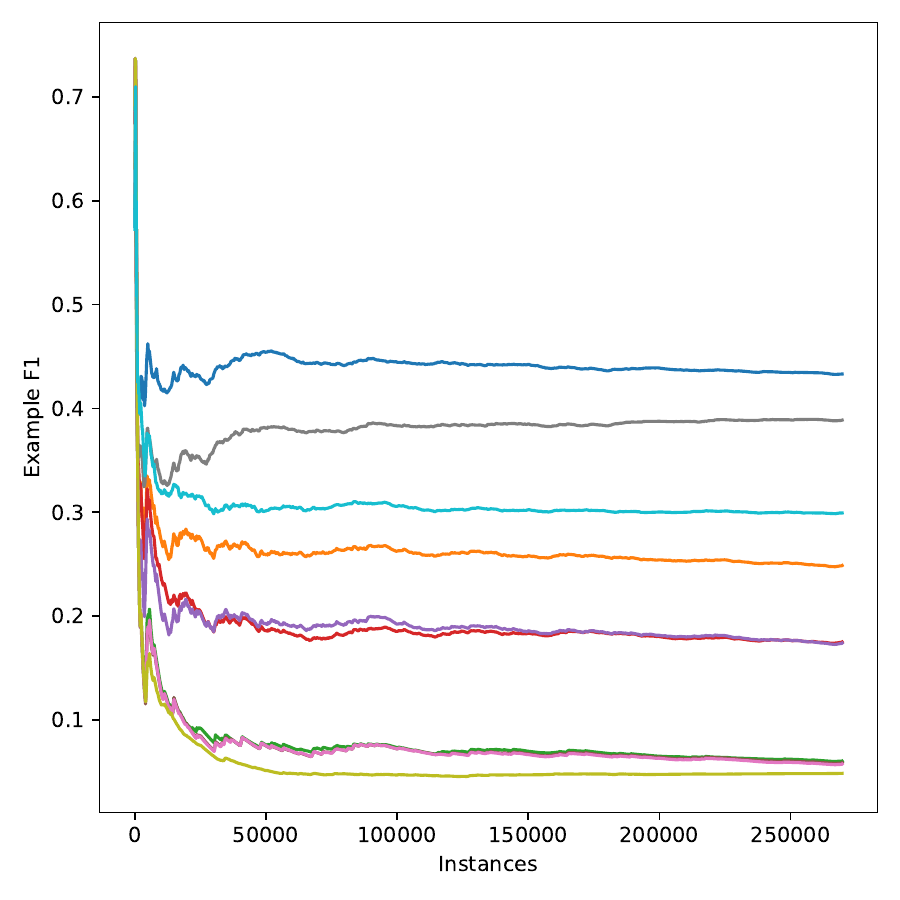}}\\
    
    \subfloat[SynHPSud]{\includegraphics[width=.248\linewidth]{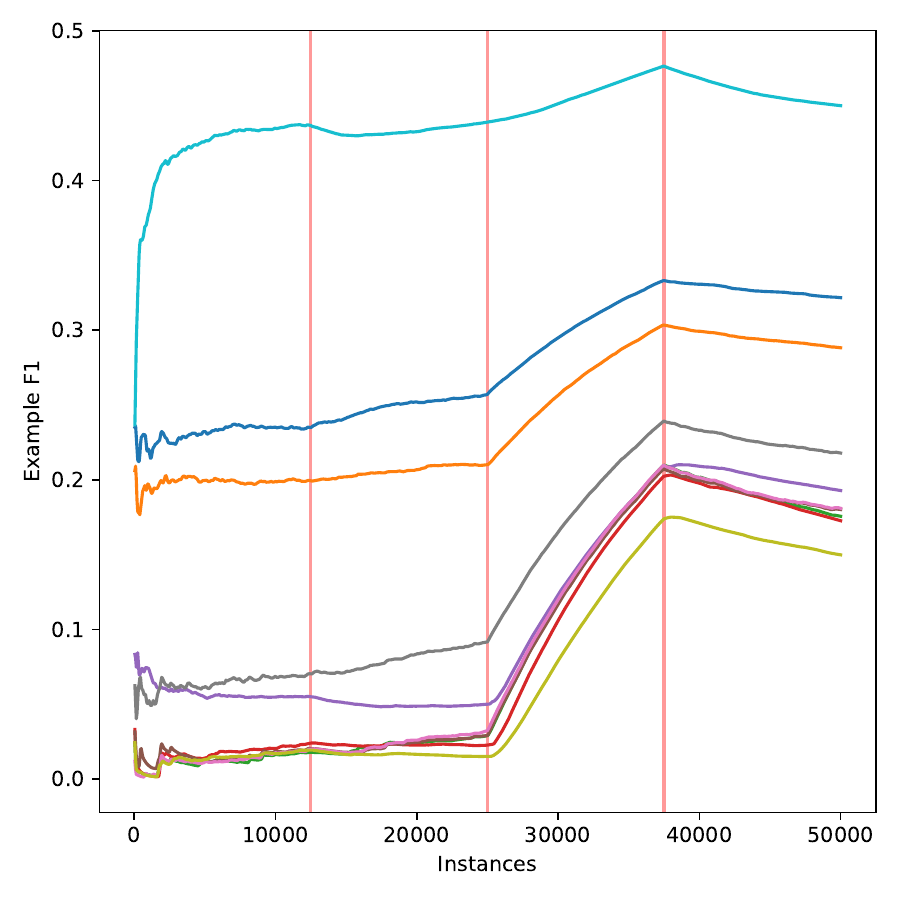}}
    \subfloat[SynHPGrad]{\includegraphics[width=.248\linewidth]{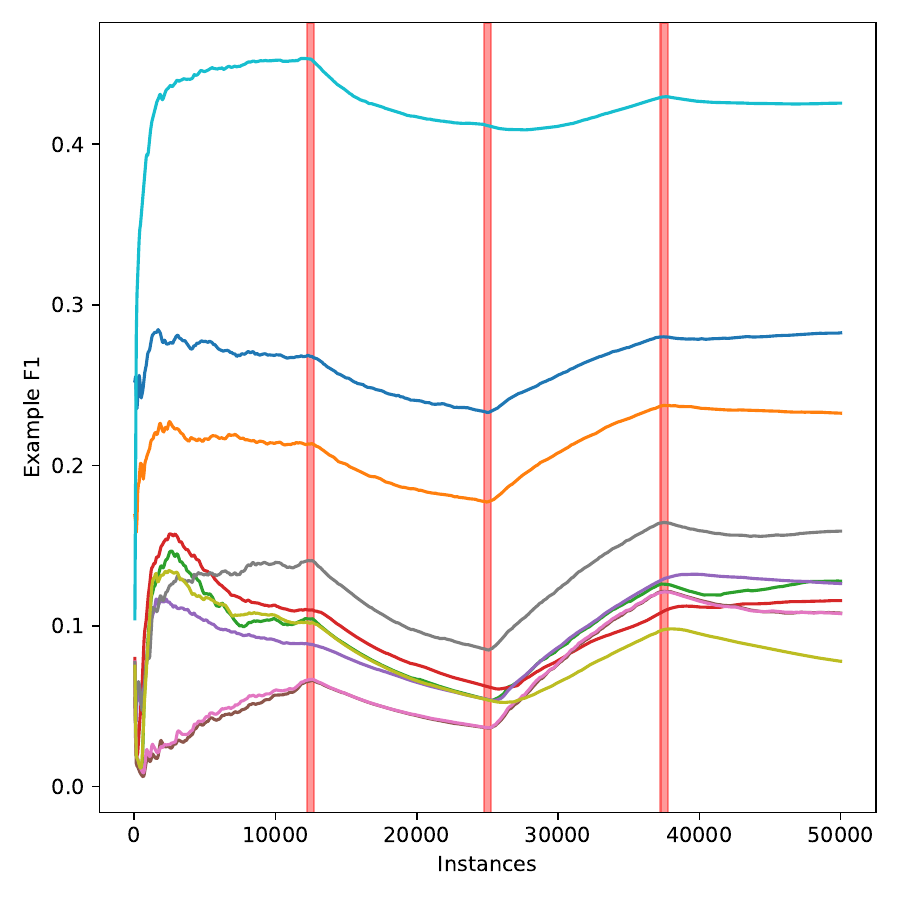}}
    \subfloat[SynHPInc]{\includegraphics[width=.248\linewidth]{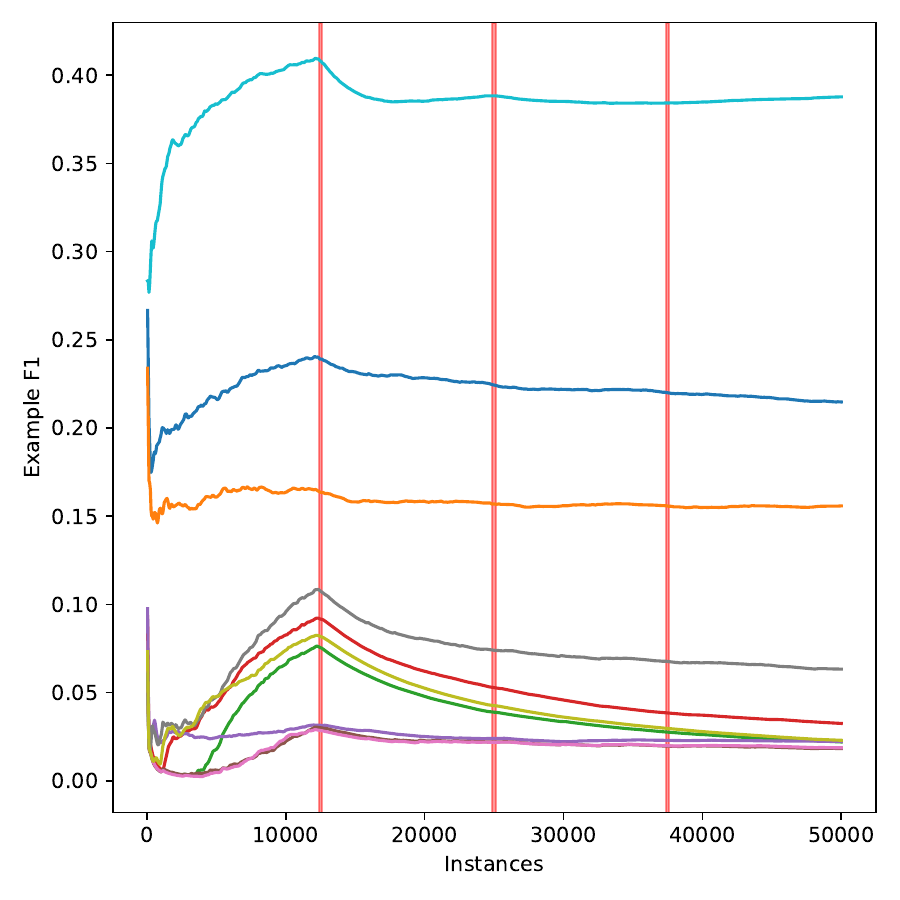}}
    \subfloat[SynHPRec]{\includegraphics[width=.248\linewidth]{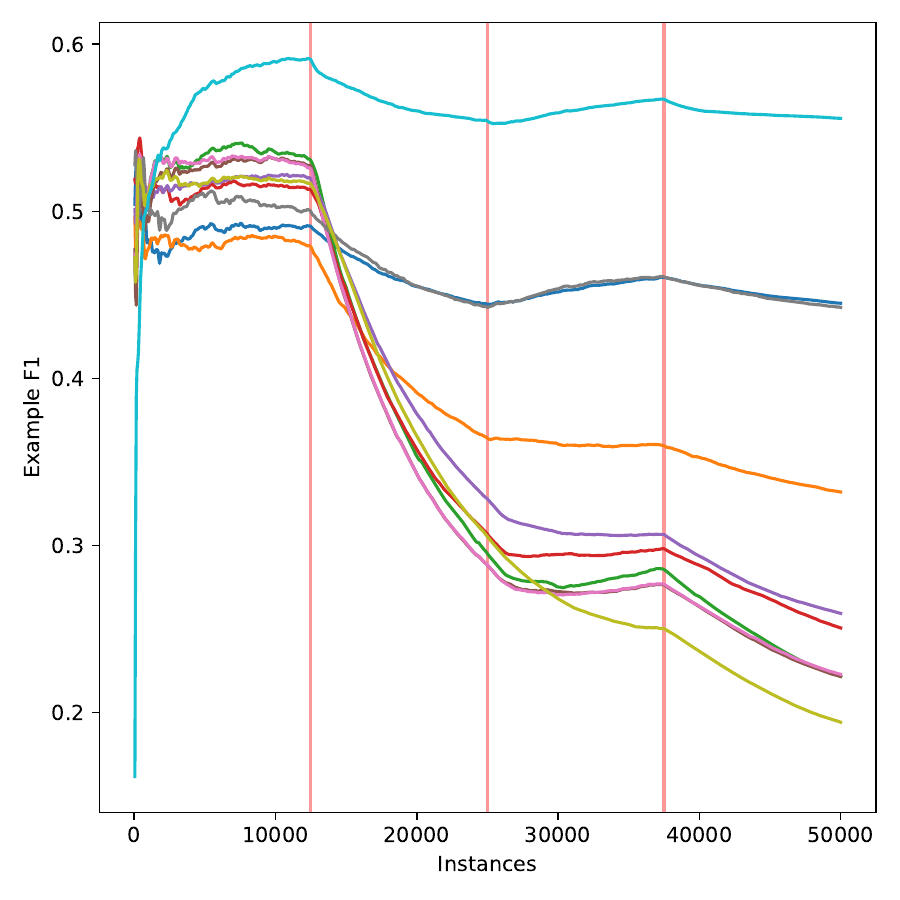}}\\
    
    \subfloat[SynRBFSud]{\includegraphics[width=.248\linewidth]{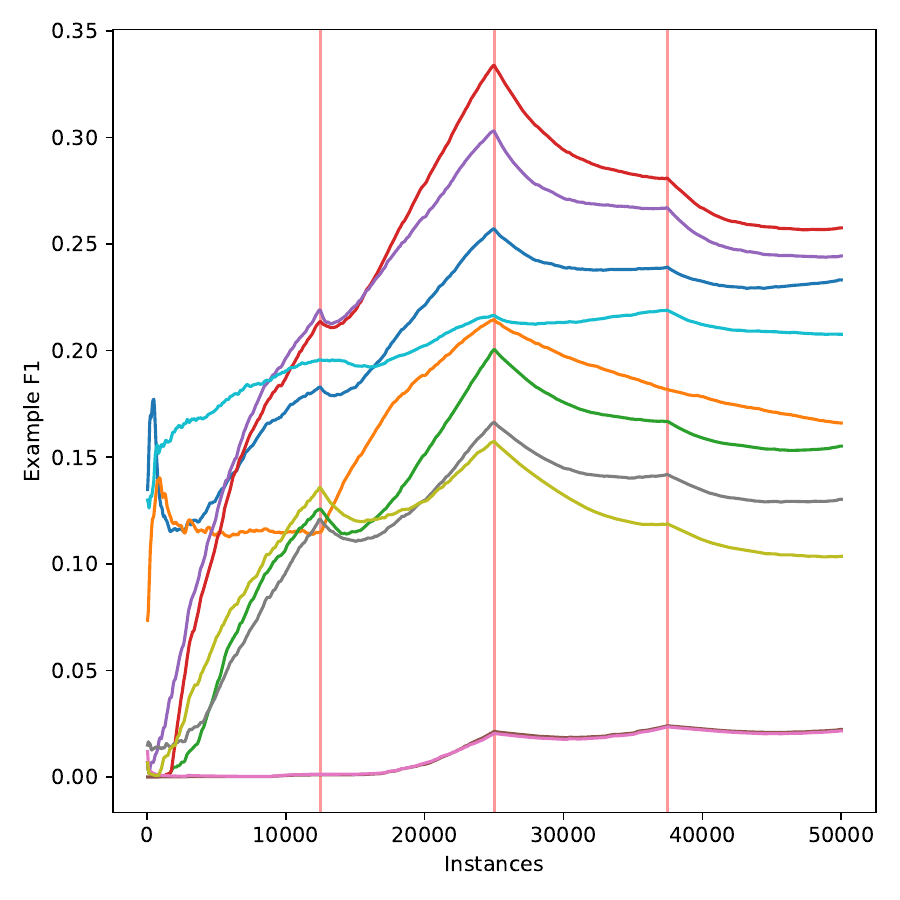}}
    \subfloat[SynRBFGrad]{\includegraphics[width=.248\linewidth]{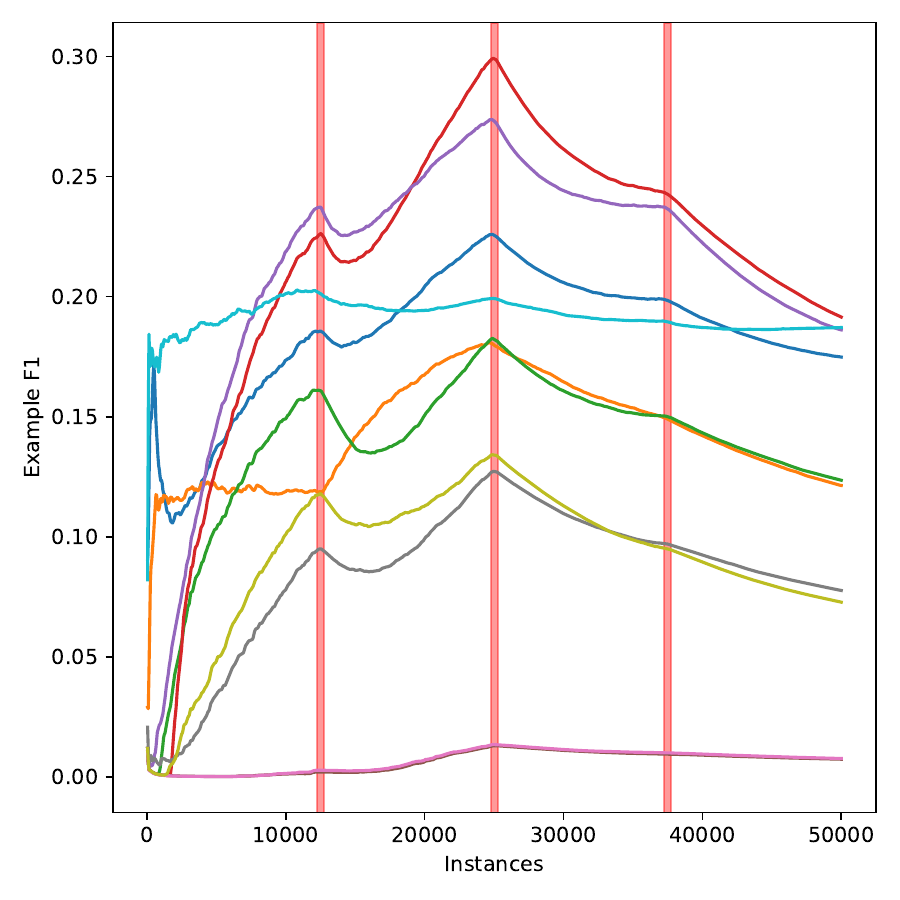}}
    \subfloat[SynRBFInc]{\includegraphics[width=.248\linewidth]{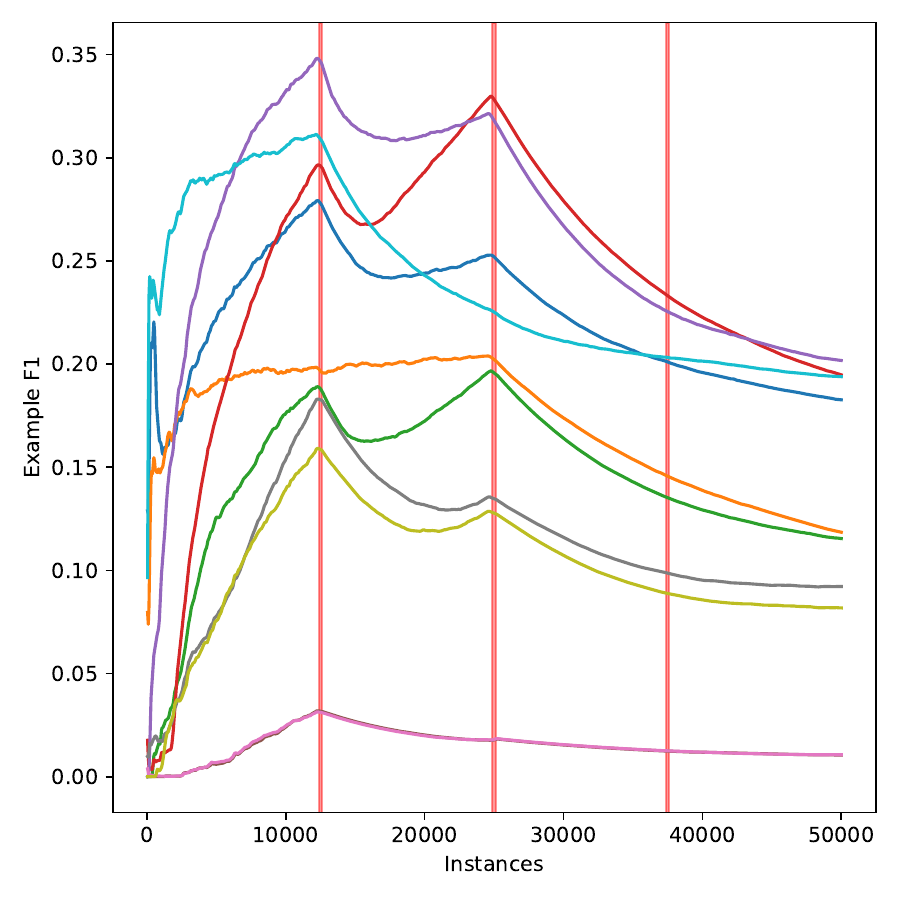}}
    \subfloat[SynRBFRec]{\includegraphics[width=.248\linewidth]{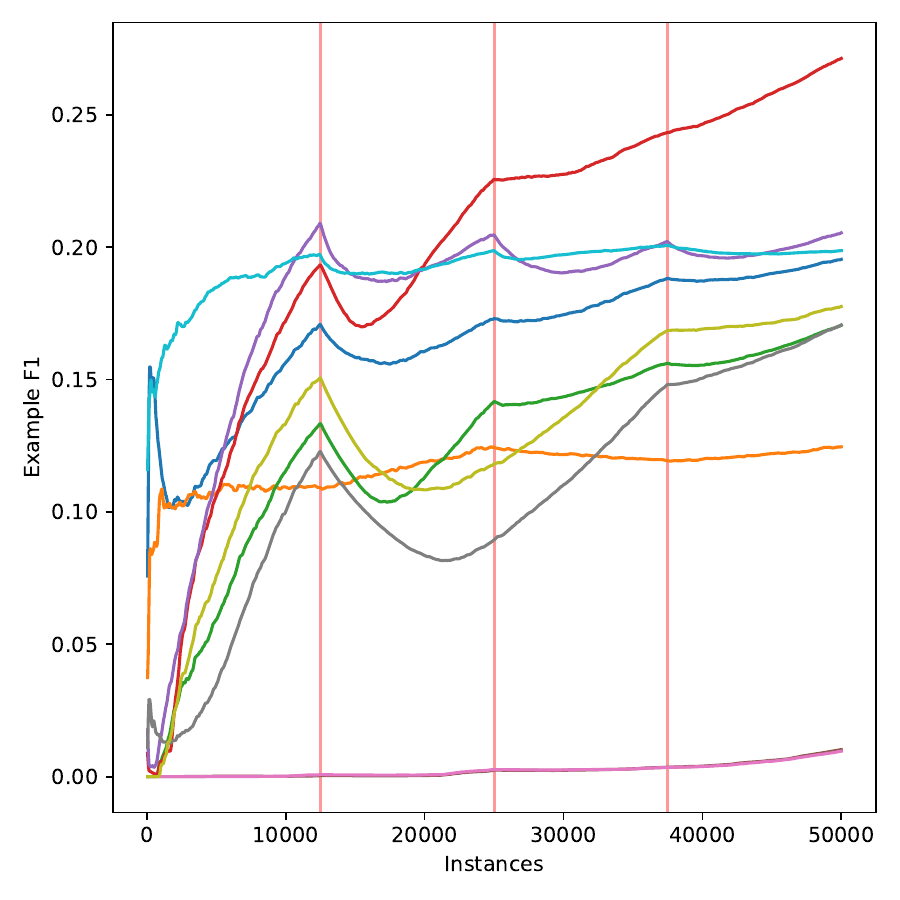}}\\

    \subfloat[SynTreeSud]{\includegraphics[width=.248\linewidth]{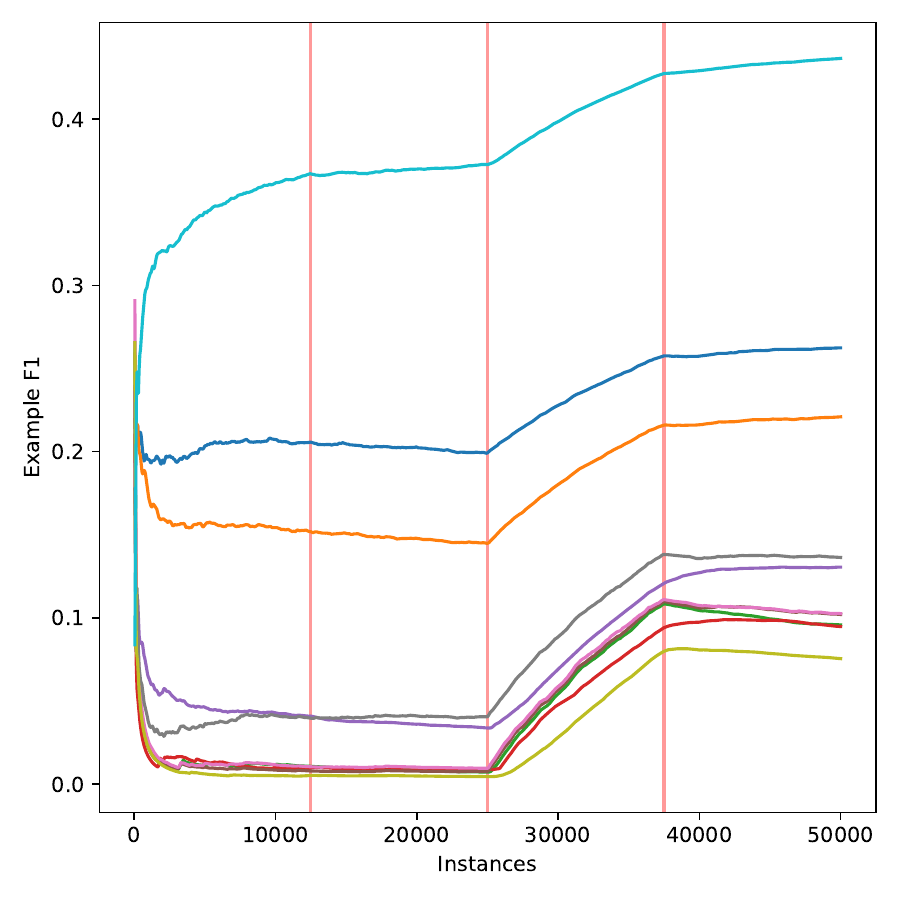}}
    \subfloat[SynTreeGrad]{\includegraphics[width=.248\linewidth]{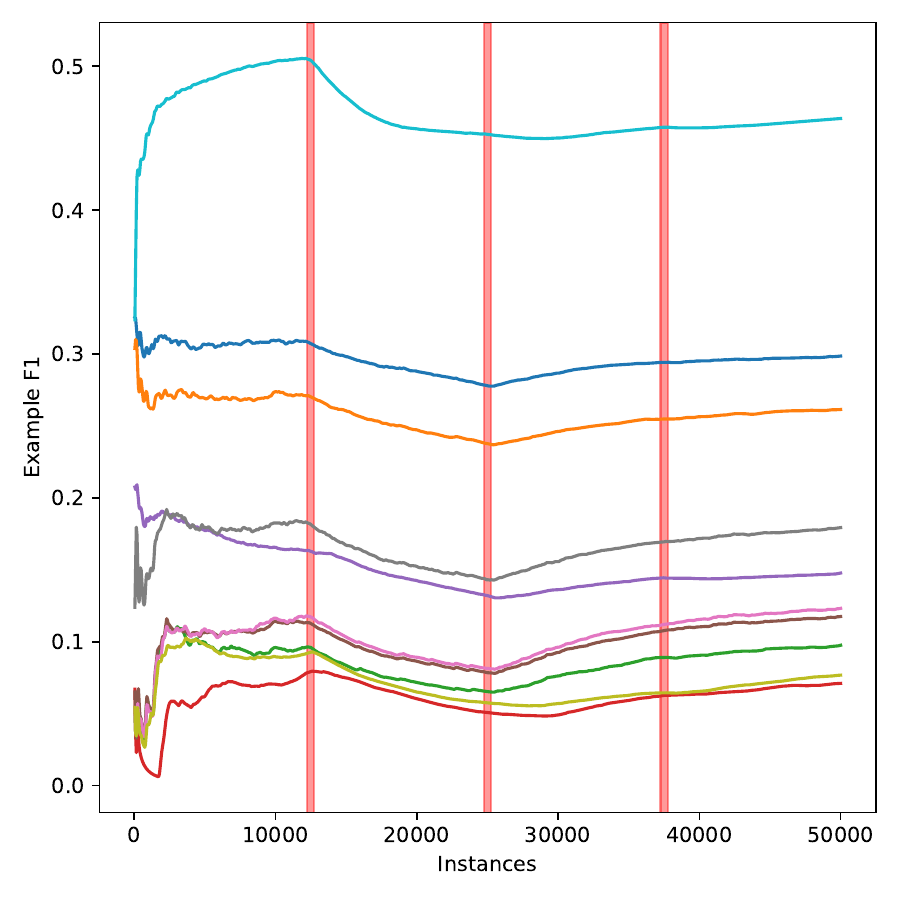}}
    \subfloat[SynTreeInc]{\includegraphics[width=.248\linewidth]{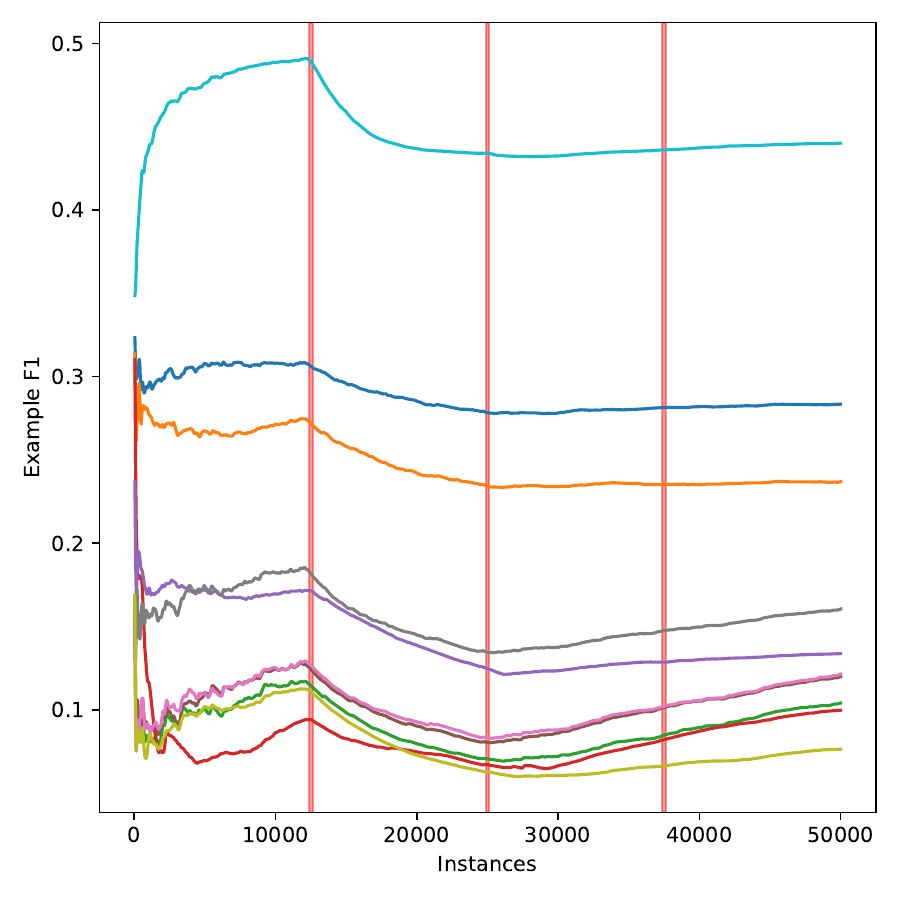}}
    \subfloat[SynTreeRec]{\includegraphics[width=.248\linewidth]{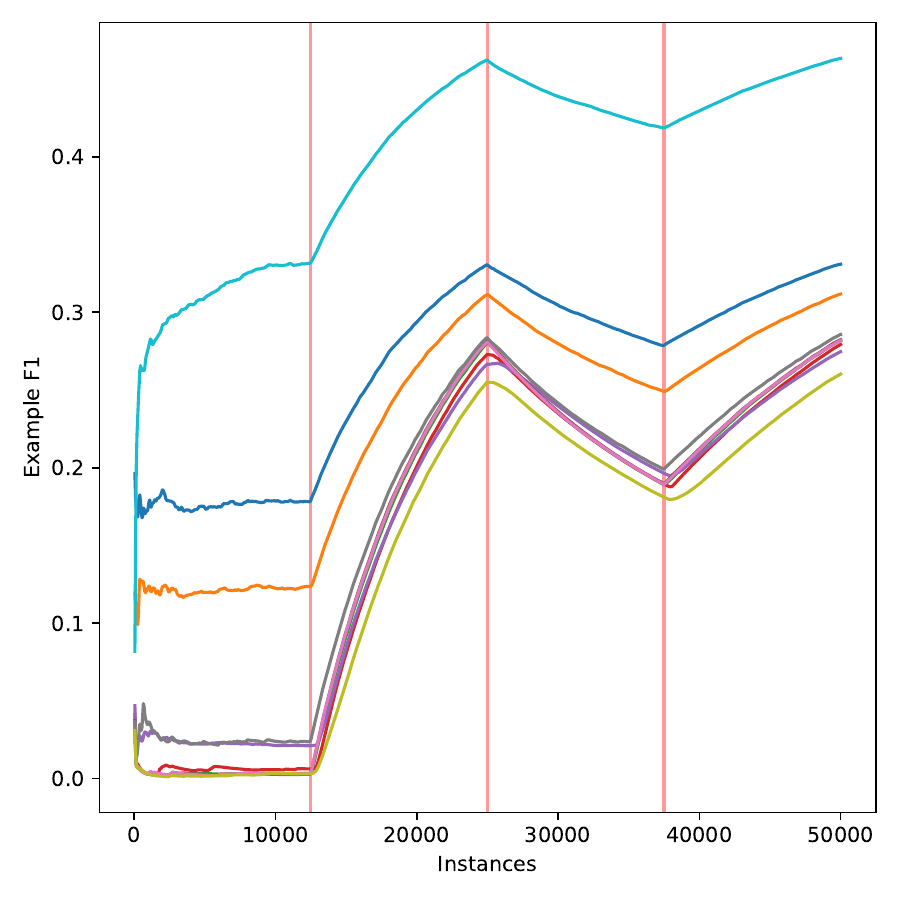}}\\
    
    \caption{Prequential evaluation of example-based F1 on a selection of datasets with concept drift (expressed as vertical lines). The type of concept drift is given by the name of the synthetic dataset; more information in Table \ref{tab:datasets-synth}.}
    \label{fig:prequential}
\end{figure*}

Figure \ref{fig:prequential} shows the evolution of the example-based F1 score, every 50 instances, on the subset of the real-world datasets that have a temporal order, and all the synthetic datasets. 
These results show great variability and that there is no one algorithm that fits all data distributions without exception. The main conclusion is that there is a relationship between the distribution of the input data and the algorithms that perform better: streams generated with random tree benefit more from tree-based algorithms, while those generated with \ac{RBF} will be better solved with NB or kNN, although when working with high dimensionality, ensemble-based methods obtain better results. \ac{MLHAT} in general is more consistent than other proposals, performing acceptably across all distributions. Specifically, MLHAT excels in real-world datasets. However, with more complex datasets, ensemble-based methods such as MLBELS or ARF obtain better results. The reasons for this are detailed below.

It can be extracted that kNN is characterized by having acceptable performance in the early stages of the stream and being resistant to class imbalance and concepts changes. However, the high dimensionality and impossibility of recalling past features make it stagnate compared to other models like \ac{HAT} or ARF, as can be seen in \textit{D20ng} or \textit{Yelp}. Models that in different ways are based on the cooperation of linear models, AMR, ABA, and iSOUPT, follow similar patterns: they perform better after having seen a large number of instances and gradually concept changes as in \textit{Yelp} or \textit{SynHPGrad}. \ac{MLHAT} in a way evolves on these two approaches thanks to the synergy of its two base classifiers: at the beginning of the stream it improves rapidly thanks to the kNN it uses as low complexity leaf classifiers. Later, if complexity increases, the \ac{LR} baggings, improved by the built tree structure, take part in this scenario. This is seen in \textit{D20ng} or \textit{SynHPRec}. 
On the other hand, ARF and HAT, the \ac{BR} transformations of Hoeffding trees, may perform better than \textit{MLHAT} on data generated with \ac{RBF}. However, they are not perfect in these scenarios because they do not perform uniformly for all distributions through which the stream passes. This may be due to the greater complexity of these data and the larger number of labels, making the \ac{BR} ensemble necessary, especially when it is a forest ensemble, as in the case of ARF.
In the case of GORT, also an ensemble of \acp{IDT} in this case applying the stacking technique, it is observed that it may have potential in real-world datasets such as \textit{Yelp}. It is also observed that it improves in all cases to individual iSOUPT, the model used as a basis for this proposal. However, in most cases it is significantly worse than \ac{MLHAT} despite assembling 10 models versus the single \ac{MLHAT} tree. This indicates the power of \ac{MLHAT} design elements versus the assembly of multiple \acp{IDT}.
Finally, MLBELS' performance depends on the type of stream.  In general, it responds well if the labels are linearly separable, as with synthetic generators based on Hyperplane and RandomTree. However, with real data and \ac{RBF}, the nature of its neural networks does not allow it to excel with more complex patterns.

By concept drift type, \ac{MLHAT} offers uniform performance in the different types. Although some models may achieve better results in specific scenarios, \ac{MLHAT} handles concept changes better in sudden, gradual and especially recurrent cases, as seen in \textit{SynHPRec} and \textit{SynTreeRec}. This indicates that using a multi-label metric to monitor concept drift is highly beneficial, as it detects changes more quickly than alternatives that monitor each label separately. This flexibility in handling different types of concept drift makes \ac{MLHAT} particularly effective with real datasets, which often exhibit more mixed concept drift patterns.
The incremental concept drift is the most challenging for all algorithms and \ac{MLHAT} is not an exception. We see how in \textit{SynHPInc} it is outperformed by MLHT and MLBELS, and in \textit{SynRBFInc} by ARF HAT and kNN. In these cases, it might be worth exploring an ensemble option. Hamming loss is the metric used in the experimentation for detecting the concept drift in MLHAT, for serving well in general for all the types of concept drifts, especially those found in real datasets (see Section \ref{sec:optimization}). However, to maximize adaptability to abrupt changes, it might be more beneficial to employ another metric, such as subset accuracy.

We can summarize the contributions of \ac{MLHAT} in terms of concept drift in the following key points:
\begin{itemize}
    \item \ac{MLHAT} shows consistent performance across various data distributions, more so than other proposals.
    \item \ac{MLHAT} combines strengths of different approaches: On the one hand, it used kNN as low-complexity leaf classifiers for rapid improvement at stream start. On the other hand, it employs \ac{LR} baggings, enhanced by tree structure, for handling increased complexity.
    \item \ac{MLHAT} offers uniform performance across different types of concept drift: Handles concept changes better in sudden, gradual, and especially recurrent cases. This flexibility makes \ac{MLHAT} particularly effective with real datasets, which often exhibit mixed-concept drift patterns.
    \item \ac{MLHAT}'s use of a multi-label metric (Hamming loss) to monitor concept drift is highly beneficial, as it detects changes more quickly than alternatives that monitor each label separately. However, other metrics might be more beneficial for maximizing adaptability to abrupt changes.
    \item Attending to the performance in specific scenarios, we can observe that \ac{MLHAT} excels particularly in real-world datasets; that it may be outperformed by ensemble methods (e.g., MLBELS, ARF) on more complex datasets; and that can struggle with incremental concept drift, like all tested algorithms.
    \item While not always the top performer, MLHAT generally provides more consistent results across various data distributions and drift types compared to other algorithms tested.
\end{itemize}

\subsection{Analysis of efficiency}\label{sec:complexcomp}

In the context of data stream classification, the efficiency and speed of processing instances are of great importance, as they are expected to arrive at high speed. Thus, the model must not only be accurate but also scalable, keeping its growth controlled over time. As shown in Section \ref{sec:overallperf}, \ac{MLHAT} is a time-competitive algorithm. In this section, we analyze the details of the time and memory consumption of \ac{MLHAT}.

The space and time complexity of \ac{MLHAT} is upper-bounded by \ac{HT} and lower-bounded by the \ac{ADWIN} method to prune nodes. Following the reasoning of \cite{Garcia-Martin2021}, we can divide the operations in the tree into three main components: updating the statistics in the leaf, evaluating possible splits, and learning in the corresponding leaf classifier. Updating the statistics requires traversing the tree to the leaf and updating the label counts at each node along the path. This has a complexity of $O(hl)$ where $h$ is the height of the tree and $l$ is the number of labels in the multi-label problem. To look for possible splits, $f$ splits will be evaluated, where $f$ is the number of features or attributes. For each $f$ split, $v$ computations of information gains are required. Finally, each information gain requires $c$ operations. Thus, split evaluations require $O(fvc)$ operations. Lastly, the complexity of learning in leaves is in the upper bound of the high-complexity classifier, a bagging method that aggregates $k=10$ logistic regressions, whose complexity depends mainly on the number of features $f$. Thus, this step has a complexity of $O(kf)$. In conclusion, the total complexity of \ac{MLHAT} in the worst-case scenario will be $O(hlf^2vck)$. Empirical results are shown below to observe the real resource consumption of \ac{MLHAT} and how the concept drift affects it.

\begin{figure*}[!tb]
    \centering
    \subfloat[Yelp time]{\includegraphics[width=.248\linewidth]{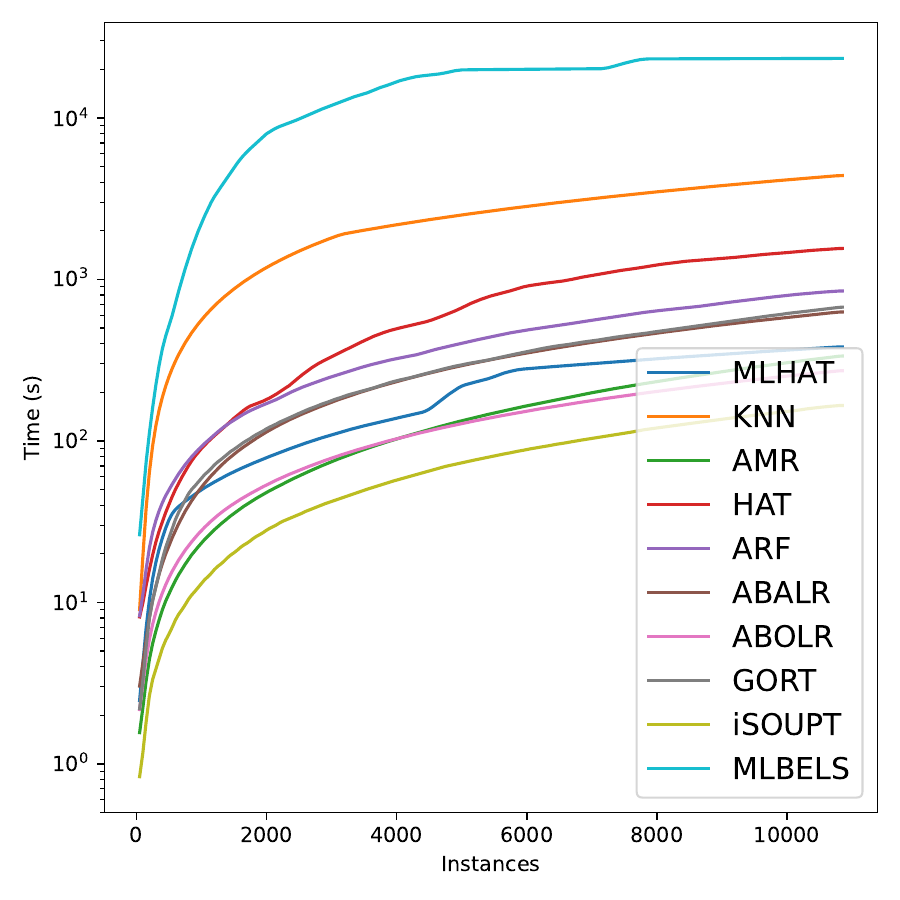}}
    \subfloat[D20ng time]{\includegraphics[width=.248\linewidth]{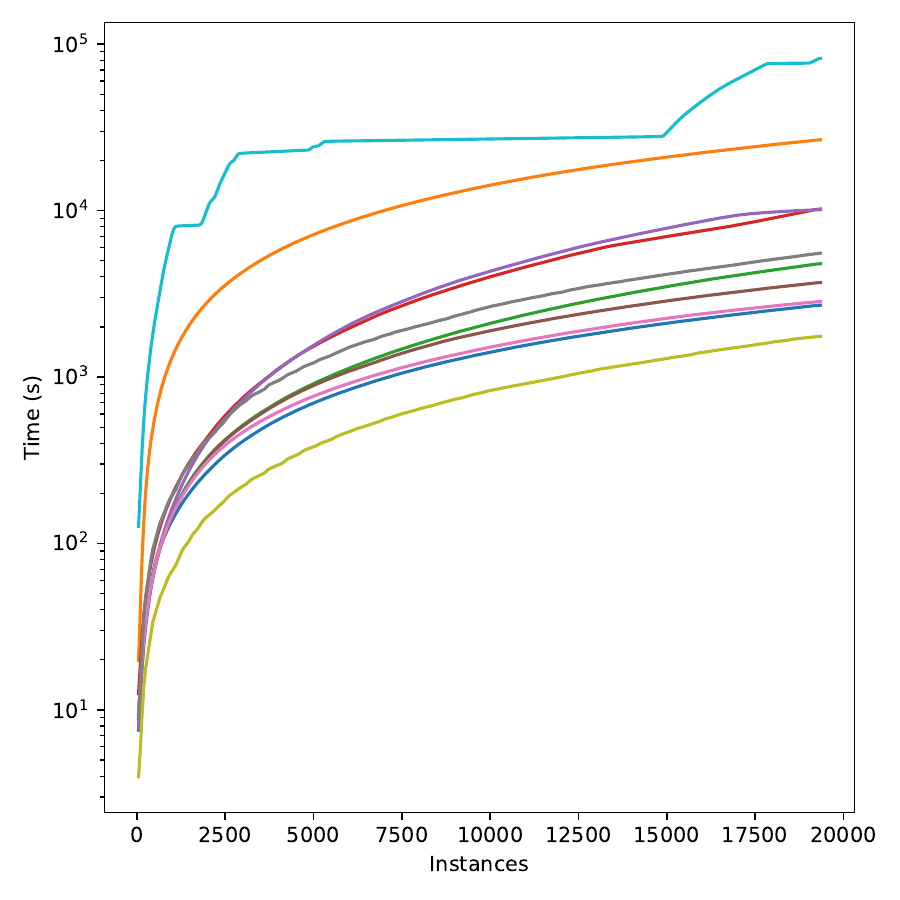}}
    \subfloat[SynRBFSud time]{\includegraphics[width=.248\linewidth]{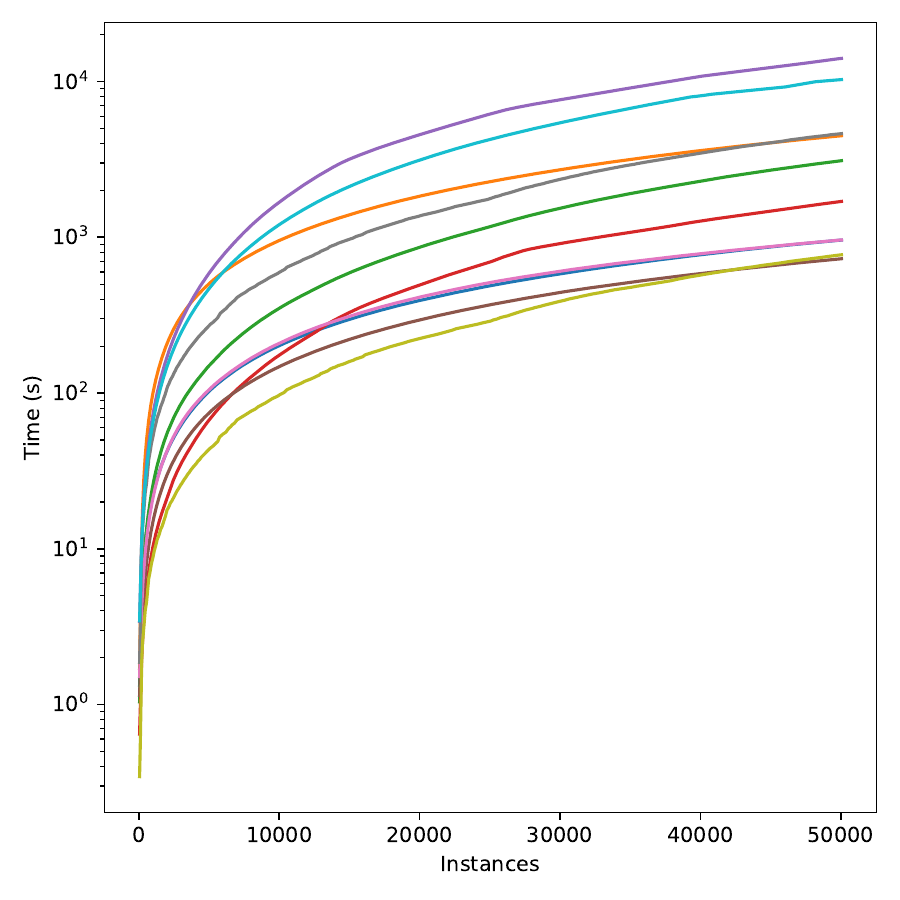}}
    \subfloat[SynHPGrad time]{\includegraphics[width=.248\linewidth]{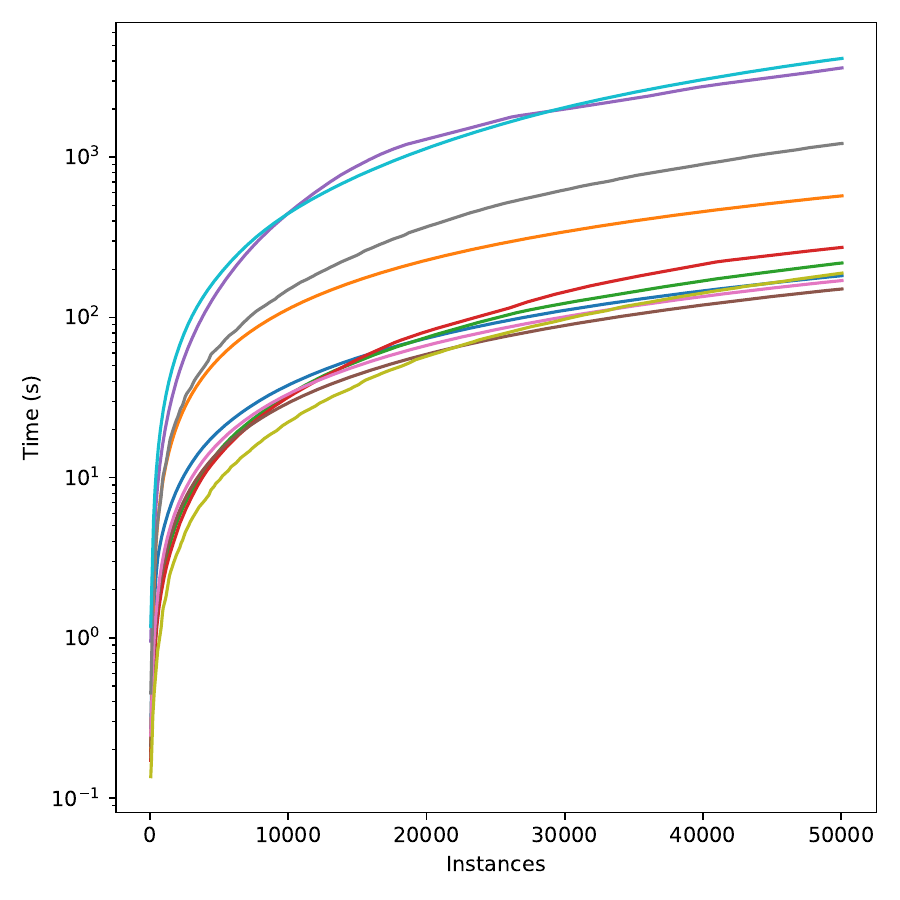}}\\
    \subfloat[Yelp memory]{\includegraphics[width=.248\linewidth]{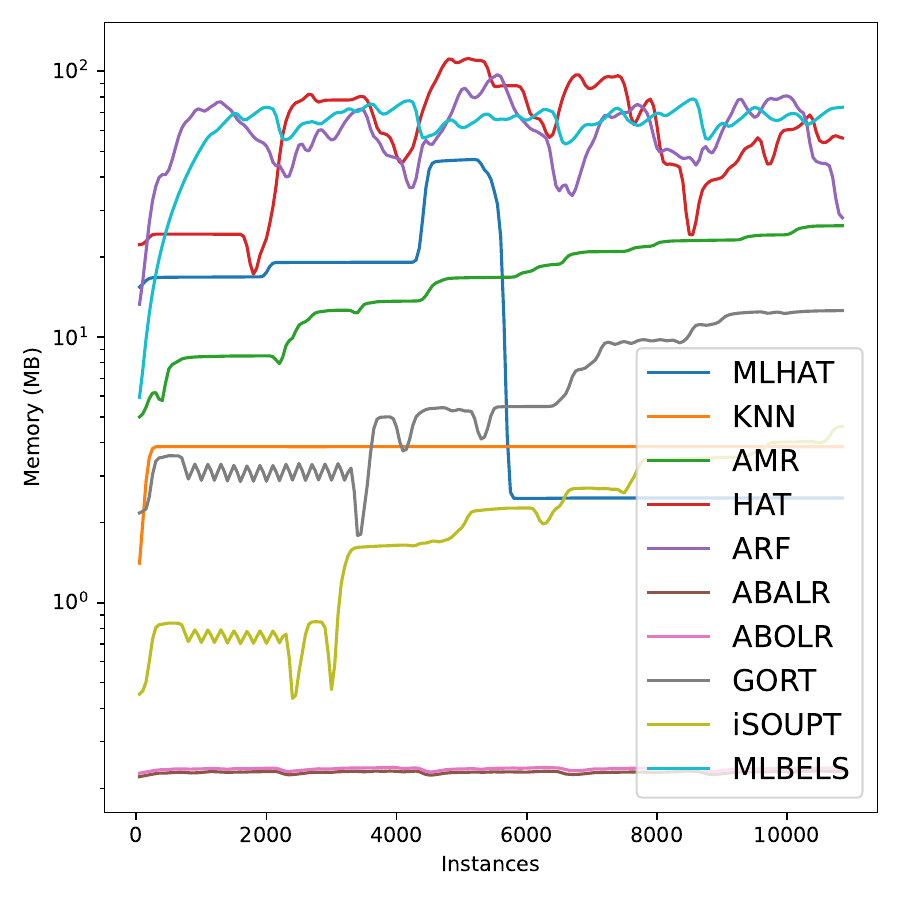}}
    \subfloat[D20ng memory]{\includegraphics[width=.248\linewidth]{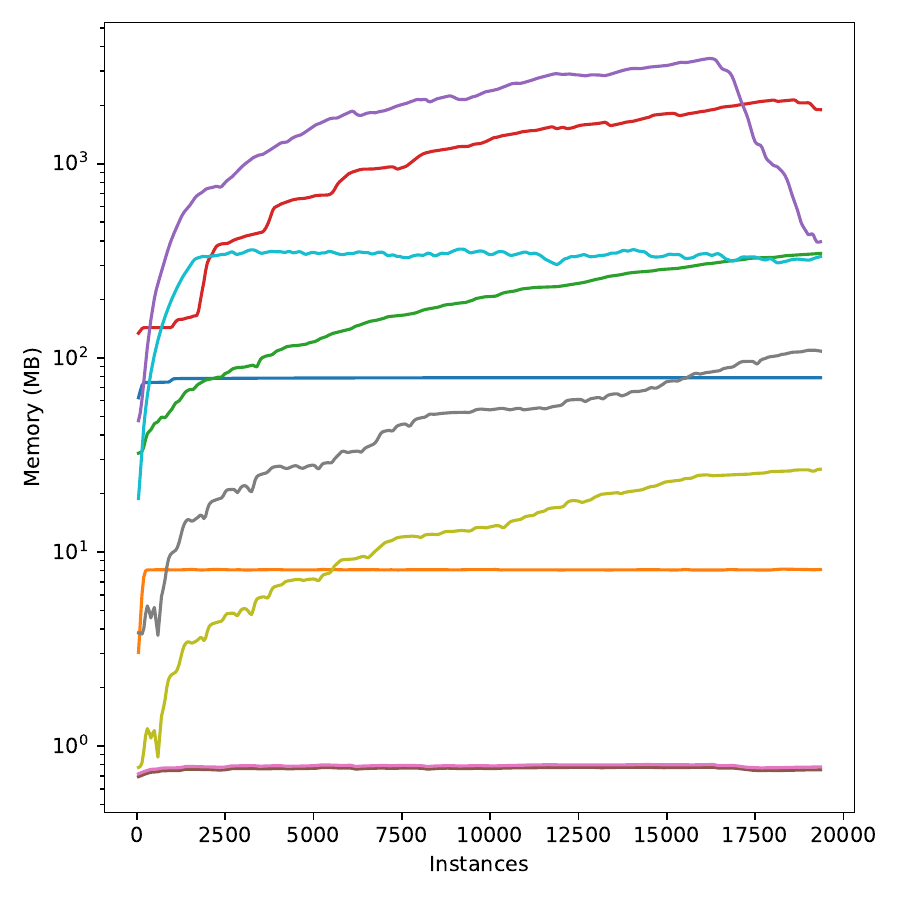}}
    \subfloat[SynRBFSud memory]{\includegraphics[width=.248\linewidth]{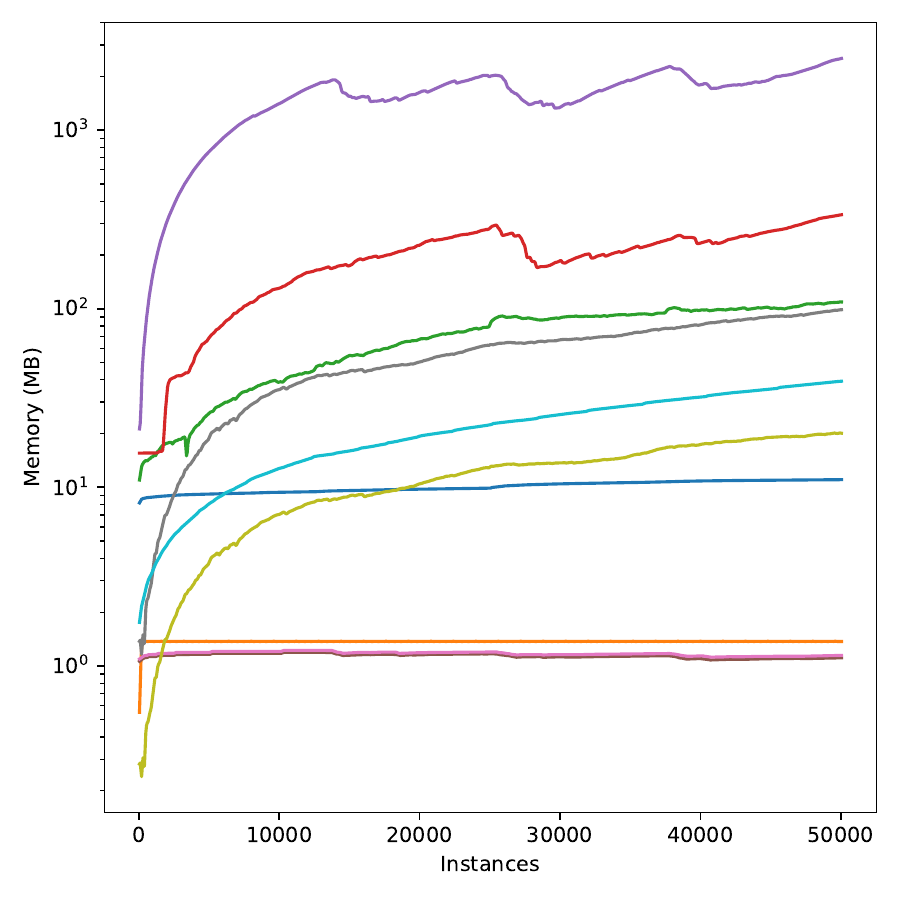}}
    \subfloat[SynHPGrad memory]{\includegraphics[width=.248\linewidth]{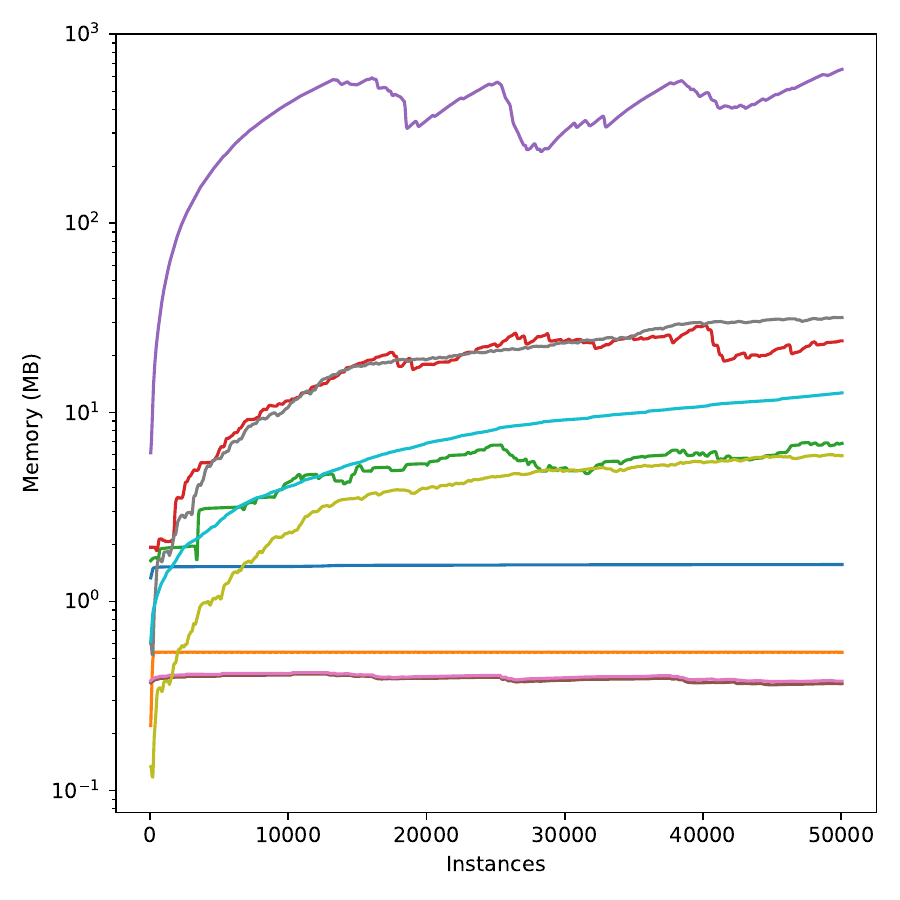}}
    
    \caption{Resources consumption on a selection of the biggest datasets using logarithmic scale}
    \label{fig:resources}
\end{figure*}

Figure \ref{fig:resources} shows the evolution in time and memory consumption of the different paradigms for a selection of datasets that combine the largest number of instances and a different number of attributes. To facilitate the visualization of results, only the most representative methods selected in the previous section are plotted here. The results show that \ac{MLHAT} is faster and more efficient than most models, especially models of similar performance like kNN and MLBELS. This is largely due to the use of a single decision tree and its adaptation to concept drift. Thus, when concept drift is detected and pruning of obsolescent parts of the tree occurs, the memory occupied by the model decreases, which also affects the processing speed of the instances. In contrast, other \ac{IDT}-based models such as MLHTPS, iSOUPT, SGT or AMF grow without limits, which can be problematic for longer streams. This is accentuated in ensemble versions of these ones like GORT. This is evident in the case for \ac{SGT} and AMF in \textit{NusWideBow}, for example. Their results have not been able to finish due to requiring hundreds of GB of RAM and/or more than 240 hours of execution. Other models based on HT and with concept drift, such as ARF and HAT also scale poorly, especially with datasets that have many labels. This is due to the multi-label transformation of \ac{BR} requires training one model for each label. In the case of ARF, this implies a memory consumption that can be 100 times higher than that of \ac{MLHAT} and takes 10 times longer to process the same amount of information. Other proposals such as kNN or ABA imply minimal or no build-up of the model, which makes them very efficient in memory and also in time. \ac{MLHAT} builds its leaf classifiers on these models, so their presence hardly affects the total computation time and memory used by our proposal, keeping it competitive with other tree-based models like MLHTPS or iSOUPT.

\section{Conclusions and future work}\label{sec:conclusions}

This paper introduced \acf{MLHAT}, an \ac{IDT} for multi-label data streams. \ac{MLHAT} incorporates a multi-probabilistic split criterion to natively consider co-occurrences in multi-label data and monitors the labelsets' entropy and cardinality at the leaves to adapt the classification process complexity.  We also proposed to monitor the concept drift on the intermediate nodes of the tree, allowing the review of split decisions if performance decreases, and replacing affected branches with new ones adapted to the new data distribution. These combined mechanisms allow \ac{MLHAT} to overcome various multi-label stream difficulties, such as concept drift and class imbalance. We presented an exhaustive experimental study to assess the competitiveness of \ac{MLHAT} against 16 different multi-label online classifiers, including the main previous \acp{IDT} and representations of other online paradigms. The experimentation covered 41 datasets and 12 multi-label metrics, for which \ac{MLHAT} achieved the top average result in 10 metrics.
We also analyzed the \ac{MLHAT} evolution across the stream through prequential evaluation, which allowed us to observe that it can perform very well from an early stage of the data stream and that it also adapts quickly to concept drift.

Future work on \ac{MLHAT} can focus on several aspects. Firstly, while our work has shown that \ac{MLHAT} is a competitive method for \ac{MLC} in data streams, it may depend on a non-trivial parameter setting process. Therefore, it would be interesting to study modifications to reduce the number of parameters to be adjusted, for example, with hyperparameterization or dynamic adjustments. Secondly, our experiments have shown that there are experiments with a high label density in which \ac{MLHAT} performance can be improved by ensembles. Thus, it can be interesting to explore certain techniques, such as bagging, boosting, or random subspaces. Finally, it would be valuable to explore the application of \ac{MLHAT} to real-world scenarios and domains where multi-label stream learning has great potential, such as predictive maintenance, pose estimation, or autonomous driving.

\section{Acknowledges}

This research was supported in part by grant PID2020-115832GB-I00 funded by MICIN / AEI / 10.13039 / 501100011033, by the ProyExcel-0069 project of University, Research and Innovation Department of the Andalusian Board, and by a FPU predoctoral grant FPU19/03924 from the Spanish Ministry of Universities.
High Performance Computing resources provided by the HPRC core facility at Virginia Commonwealth University (https://hprc.vcu.edu) were used for conducting the research reported in this work.

\bibliography{refs}
\end{document}